%% file: main.tex
\RequirePackage{fix-cm}

\documentclass[smallcondensed]{supplementary/Style/svjour3}     % onecolumn (ditto)
\smartqed

\usepackage{supplementary/Style/my_settings}
\input{supplementary/Style/math_commands}

\begin{document}

\title{A Brain-inspired Algorithm for Training Highly Sparse Neural Networks
\thanks{This project is partially financed by the Dutch Research Council (NWO).}
}

\author{Zahra Atashgahi \and  Joost Pieterse \and Shiwei Liu \and Decebal Constantin Mocanu \and Raymond Veldhuis \and ‪Mykola Pechenizkiy‬}

\institute{  Z. Atashgahi  \and D.C. Mocanu \and 
            R.N.J. Veldhuis\at
            Faculty of Electrical Engineering, Mathematics and Computer Science, University of Twente, 7500AE Enschede, the Netherlands \\
              Tel.: +31534896141\\
              \email{z.atashgahi@utwente.nl}        
           \and
            J. Pieterse \and S. Liu\and D.C. Mocanu \and M.  Pechenizkiy \at
              Faculty of Mathematics and Computer Science, Eindhoven University of Technology, 5600 MB Eindhoven, the Netherlands   
            \and
             M.  Pechenizkiy \at
            Faculty of Information Technology, University of Jyväskylä, 40014 Jyväskylä, Finland
}

\date{Received: date / Accepted: date}

\maketitle
%%%%%%%%%%%%%%%%%%%%%%%%%%%%%%%%%%%%%%%%%%%%%%%%%%%%%%%%%%%%%%%%%%%
%%%%%%%%%%%%                    Content                  %%%%%%%%%%
%%%%%%%%%%%%%%%%%%%%%%%%%%%%%%%%%%%%%%%%%%%%%%%%%%%%%%%%%%%%%%%%%%%
\input{sections/0abstract}
\input{sections/1introduction}

\input{sections/2related_work}

\input{sections/3methodology}

\input{sections/4Experiments}

\input{sections/Discussion}

\section{Conclusion and Broader Impacts} \label{ssec:coclusion}
\looseness=-1
In this research, we introduced a new biologically plausible sparse training algorithm named CTRE. CTRE exploits both the similarity of neurons as an importance measure of the connections and random search, sequentially (CTRE\textsubscript{seq}) or simultaneously (CTRE\textsubscript{sim}), to explore a performant sparse topology. The findings of this study indicate that the cosine similarity between neurons' activations can help to evolve a sparse network in a purely sparse manner even in highly sparse scenarios, while most state-of-the-art methods may fail in these cases. In our view, by using the neurons' similarity to evolve the topology, our proposed approach can be an excellent initial step toward explainable sparse neural networks. Overall, due to the ability of CTRE to extract highly sparse neural networks, it can be a viable alternative for saving energy in both low-resource devices and data centers and pave the way to achieving environmentally friendly AI systems. Nevertheless, the trade-off between accuracy and sparsity, with CTRE deployed on real-world applications, should be considered carefully; particularly, if any loss of accuracy may pose safety risks to the user, the sparsity level of the network needs to be analyzed with greater care.

An interesting future direction of this research is to extend CTRE to CNNs; driven by the decent performance of CTRE on image datasets, we believe that it has the potential to be extended to CNN architectures. However, in-depth theoretical analysis and systematic experiments are required to adapt this similarity metric to CNN architectures. This is due to the fact that CNNs require weight sharing, which does not exist in real neurons, and consequently, it is not straightforward to apply Hebbian learning directly \cite{pogodin2021towards}. There have been some efforts to make CNNs more biologically plausible \cite{pogodin2021towards, bartunov2018assessing}. Therefore, applying CTRE to CNNs should be done with great care and theoretical analysis that we believe is in the scope of future works.

\section*{Declarations}

\begin{itemize}
    \item \textbf{Funding.} This project is partially funded by the NWO EDIC project.
    \item \textbf{Conflicts of interest/Competing interests.} Not Applicable.
    \item \textbf{Ethics approval.} Not Applicable.
    \item \textbf{Consent to participate.} Not Applicable.
    \item \textbf{Consent for publication.} Not Applicable.
    \item \textbf{Availability of data and material.} All the datasets used in this research are currently public. The references for all datasets have been cited in Section \ref{sssec:Datasets}. 
    \item \textbf{Code availability.}  The implementation code is available on Github at \url{https://github.com/zahraatashgahi/CTRE}.
    \item \textbf{Authors' contributions.} J. Pieterse and D.C. Mocanu designed and developed a preliminary idea and proof of concept experiments. Z. Atashgahi designed and developed the CTRE algorithm, designed and carried out the empirical validation, and performed the analysis and interpretation of the results. S. Liu helped in performing the experiments and interpreting the results. D.C. Mocanu, R. Veldhuis and M. Pechenizkiy supervised the research. All authors provided critical feedback and helped shaped the research. Z. Atashgahi wrote the manuscript with input from all authors.
\end{itemize}

\bibliographystyle{plainnat}
\bibliography{sections/resources} 
\input{sections/appendix}

\end{document}

%% file: supplementary/Style/math_commands.tex
\def\eqref#1{equation~\ref{#1}}
\def\1{\bm{1}}

\def\vtheta{{\bm{\theta}}}

\def\vx{{\bm{x}}}
\def\vy{{\bm{y}}}

\def\mA{{\bm{A}}}

\def\mSim{{\bm{Sim}}}
\DeclareMathAlphabet{\mathsfit}{\encodingdefault}{\sfdefault}{m}{sl}
\SetMathAlphabet{\mathsfit}{bold}{\encodingdefault}{\sfdefault}{bx}{n}
\def\sR{{\mathbb{R}}}

\def\sX{{\mathbb{X}}}

\def\emSim{{Sim}}
\newcommand{\R}{\mathbb{R}}

\DeclareMathOperator*{\argmin}{arg\,min}

%% file: sections/0abstract.tex
% Abstract
\begin{abstract}
\looseness=-1
Sparse neural networks attract increasing interest as they exhibit comparable performance to their dense counterparts while being computationally efficient. Pruning the dense neural networks is among the most widely used methods to obtain a sparse neural network. Driven by the high training cost of such methods that can be unaffordable for a low-resource device, training sparse neural networks sparsely from scratch has recently gained attention. However, existing sparse training algorithms suffer from various issues, including poor performance in high sparsity scenarios, computing dense gradient information during training, or pure random topology search. In this paper, inspired by the evolution of the biological brain and the Hebbian learning theory, we present a new sparse training approach that evolves sparse neural networks according to the behavior of neurons in the network. Concretely, by exploiting the cosine similarity metric to measure the importance of the connections, our proposed method, \enquote{Cosine similarity-based and Random Topology Exploration (CTRE)}, evolves the topology of sparse neural networks by adding the most important connections to the network without calculating dense gradient in the backward. We carried out different experiments on eight datasets, including tabular, image, and text datasets, and demonstrate that our proposed method outperforms several state-of-the-art sparse training algorithms in extremely sparse neural networks by a large gap. The implementation code is available on Github\footnote{\url{https://github.com/zahraatashgahi/CTRE}}.
\end{abstract}

%% file: sections/1introduction.tex
% Introduction
\section{Introduction}
\looseness=-1
Dense artificial neural networks are a commonly used machine-learning technique that has a wide range of application domains, such as speech recognition \cite{graves2013speech}, image processing \cite{liang2015recurrent, masi2018deep}, and natural language processing (NLP) \cite{NEURIPS2020_1457c0d6}. It has been shown in \cite{hestness2017deep} that the performance of deep neural networks scales with model size and dataset size, and generalization benefits from over-parameterization \cite{neyshabur2018the}. However, the ever-increasing size of deep neural networks has given rise to major challenges, including high computational cost both during training and inference and high memory requirement \cite{zhang2020deep}. Such an increase in the number of computations can lead to a critical rise in the energy consumption in data centers and, consequently, a deteriorative effect on the environment \cite{yang2018ai}. However, a trustworthy AI system should function in the most environmentally friendly way possible during development, and deployment \cite{AIethic2019}. In addition, such gigantic computational costs will lead to a situation where on-device training and inference of neural network models on low-resource devices, e.g., an edge device with limited computational resources and battery life, might not be economically viable \cite{zhang2020deep}.

%%%%%%%%%%%%%%%%%%   Sparse Neural Networks
\looseness=-1
Sparse neural networks have been considered as an effective solution to address these challenges \cite{hoefler2021sparsity, mocanu2021sparse}. By using sparsely connected layers instead of fully-connected ones, sparse neural networks have reached a competitive performance to their dense equivalent networks in various applications \cite{frankle2018lottery, atashgahi2020quick}, while having much fewer parameters. It has been shown that biological brains, especially the human brain, enjoy sparse connections among neurons \cite{friston2008hierarchical}. Most existing solutions to obtain sparse neural networks focus on inference efficiency in order to reduce the storage requirement of deploying the network and prediction time of test instances. This class of methods, named \textit{dense-to-sparse} training, starts by training a dense neural network followed by a pruning phase that aims to remove unimportant weight from the network. As categorized in \cite{mocanu2021sparse}, in dense-to-sparse training, the pruning phase can be done after training \cite{lecun1990optimal, frankle2018lottery, han2015learning}, simultaneous to training \cite{louizos2018learning}, or one-shot prior to training \cite{lee2018snip}. However, starting from a dense network leads to a memory requirement of fitting a dense network on the device and the computational resources for at least a few iterations of training the dense model. Therefore, training sparse neural networks using dense-to-sparse methods might be infeasible on low-resource devices due to the energy and computational resource constraints.

\begin{figure}[!t]
        \begin{center}
        \centerline{\includegraphics[width=1.1\textwidth]{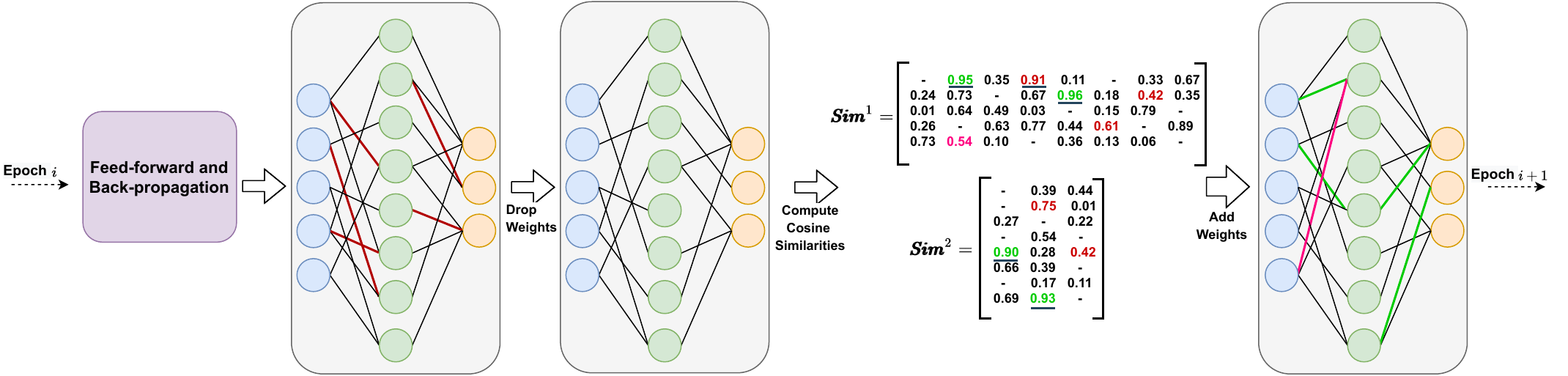}}
        \caption{\looseness=-1
        Schematic of the proposed approach (CTRE\textsubscript{sim}). At each epoch, after feed-forward and back-propagation, a fraction $\zeta$ of the weights with the smallest magnitude is dropped (red connections). Then, similarity matrices  $\mSim^{1}$ and  $\mSim^{2}$ are computed using Equation \ref{eq:cosine_similarity} to find the most important connections to add to the network; however, we do not consider the similarity of the existing connections (empty entries). Finally, the weights corresponding to the highest similarity values in the similarity matrices (underlined values) that have not been dropped in the weight removal step are added to the network (underlined green values), the same amount as removed previously. If a connection with high similarity has been dropped in the weight removal step (underlined red value), a random connection will be inserted instead (pink connection). }
        \label{fig:diagram}
        \end{center}
    \end{figure}

%%%%%%%%%%%%%%%%%%   Sparse Training
With the emergence of the \textit{sparse training} concept in \cite{mocanu2016topological}, there has been a growing interest in training sparse neural networks which are sparse from scratch. This sparse connectivity might be fixed during training (known as static sparse connectivity \cite{mocanu2016topological, mocanu2021sparse, kepner2019radix}), or might dynamically change, by removing and re-adding weights (known as dynamic sparse connectivity \cite{mocanu2018scalable, bellec2018deep}). By optimizing the topology along with the weights during the training, dynamic sparse training algorithms outperform the static ones \cite{mocanu2018scalable}. As discussed in \cite{mocanu2018scalable}, the weight removal in dynamic sparse training algorithms is similar to the synapses shrinkage in the human brain during sleep, where the weak synapses shrink and the strong ones remain unchanged. While most dynamic sparse training methods use magnitude as a pruning criterion, weight regrowing approaches are of different types, including random \cite{mocanu2018scalable, pmlr-v97-mostafa19a} and gradient-based regrowth \cite{evci2020rigging, jayakumar2020top}. As shown in \cite{liu2021we}, random addition of weights might lead to a low training speed, and the performance of sparse training is highly correlated with the total number of parameters explored during training. To speed up the convergence, gradient information of non-existing connections can be used to add the most important connections to the network \cite{dettmers2019sparse}. However, computing the gradient of all non-existing connections in a sparse neural network can be computationally demanding. Furthermore, increasing the network size might escalate the high computational cost into a bottleneck in the sparse training of networks on low-resource devices. Besides, in Section \ref{ssec:performance_evaluation}, we demonstrate that some gradient-based sparse training algorithms might fail in a highly sparse neural network.

In this paper, to address some of these challenges, we introduce a more biologically plausible algorithm for obtaining a sparse neural network. By taking inspiration from the Hebbian learning theory, which states \enquote{neurons that fire together, wire together} \cite{hebb2005organization}, we introduce a new weight addition policy in the context of sparse training algorithms. Our proposed method, \enquote{Cosine similarity-based and Random Topology Exploration (CTRE)}, exploits both the similarity of neurons as an importance measure of the connections and random search simultaneously (CTRE\textsubscript{sim}, Figure \ref{fig:diagram}) or sequentially (CTRE\textsubscript{seq}) to find a performant sub-network. In short, our contributions are as follows:

\begin{itemize}
    \item[$\bullet$] We propose a novel and biologically plausible algorithm for training sparse neural networks, which has a limited number of parameters during training. Our proposed algorithm, CTRE, exploits both similarity of neurons and random search to find a performant sparse topology.
    \item[$\bullet$] We introduce the Hebbian learning theory in the training of the sparse neural networks. Using the cosine similarity of each pair of neurons in two consecutive layers, we determine the most important connections at each epoch during sparse training of the network; we discuss in detail why this approach is an extension to the Hebbian learning theory in Section \ref{ssec:cosine_in_nn}.
    \item[$\bullet$] Our proposed algorithms outperform state-of-the-art sparse training algorithms in highly sparse neural networks.
\end{itemize}

\looseness=-1
While deep learning models have shown great success in vision and NLP tasks, these models have not been fully explored in the domain of tabular data \cite{popov2019neural}. However, designing deep models that are capable of processing tabular data is of great interest for researchers as it paves the way to building multi-modal pipelines for problems \cite{gorishniy2021revisiting}. This paper mainly focuses on Multi-Layer Perceptrons (MLPs), which are commonly used for tabular and biological data. Despite the simple structure of MLPs and having only a few hyperparameters to tune, they have shown good performance in classification tasks \cite{galke2021forget, tolstikhin2021mlp}. In addition, in \cite{jouppi2017datacenter}, authors investigated that despite the massive attention on CNN architectures, they utilize only $5\%$ of the neural network workload of TPUs in Google data centers, while MLPs constitutes $61\%$ of the total workload. Therefore, it is crucial to develop an efficient algorithm that can accelerate MLPs and are resource-efficient during training and inference. To pursue this goal, in this research, we aim to design sparse MLPs with a limited number of parameters during training and inference. To demonstrate the validity of our proposed algorithm, in addition to evaluating the methods on tabular and text datasets, we compare the methods also on the image datasets such as MNIST, Fashion-MNIST, and CIFAR10/100 datasets which are commonly used as benchmarks in previous studies.

%% file: sections/2related_work.tex
% Background
\section{Background}
%%%%%%%%%%%%%%%%%%%%%%%%%%%%%%%%%%%%%%%%%%%%%%%%%%%%%%%%%%%%%%%%%%%%%%%%%%%%%
\subsection{Sparse Neural Networks}
Methods to obtain and train sparse neural networks can be stratified into two major categories: dense-to-sparse and sparse-to-sparse. In the following, we shed light on each of these two approaches.

\looseness=-1
%--------------------------------------------------------------%
\textbf{Dense-to-sparse.}
%--------------------------------------------------------------%
Dense-to-sparse methods to obtain sparse neural networks start training from a dense model and then prune the unimportant connections. They can be divided into three major subcategories:  
$(1)$ \textit{Pruning after training}: Most existing dense-to-sparse methods start with a trained dense network and iteratively (one or several iterations) prune and retrain the network to reach desired sparsity level. Seminal works were performed in the 1990s in \cite{lecun1990optimal, hassibi1993second}, where authors use hessian matrix information to prune a trained dense network. More recently, in \cite{han2015learning, frankle2018lottery}, authors use magnitude to remove unimportant connections. Other metrics, such as gradient \cite{LIU201984}, Taylor expansion \cite{molchanov2016pruning, Molchanov_2019_CVPR}, and low-rank decomposition \cite{wang2019eigendamage, li2020group}, have been also employed to prune the network. While being effective techniques in terms of the performance of the obtained sparse network, these methods suffer from high computational costs during training. 
$(2)$ \textit{Pruning during training}: To decrease the computational cost, this group of methods perform pruning during training \cite{gale2019state, junjie2019dynamic, pmlr-v119-kusupati20a}. Various criteria can be used for pruning, such as magnitude \cite{guo2016dynamic, zhu2017prune}, L$_0$ regularization \cite{louizos2018learning, NEURIPS2020_83004190}, group Lasso regularization \cite{10.5555/3157096.3157329}, and variational dropout \cite{molchanov2017variational}.
$(3)$ \textit{Pruning before training}: The first study to apply pruning prior to training was done by \citeauthor{lee2018snip} in \cite{lee2018snip}, that used connection sensitivity to remove weights. Later works have followed the same approach by pruning the network before training using different approaches, such as gradient norm after pruning \cite{wang2019picking}, connection sensitivity after pruning \cite{de2020progressive}, and Synaptic Flow \cite{tanaka2020pruning}.

\looseness=-1
\textbf{Sparse-to-sparse.}
To lower the computational cost of dense-to-sparse methods, sparse-to-sparse training algorithms (also known as sparse training) use a sparse network from scratch with a sparse connectivity, which might be static (static sparse training \cite{mocanu2016topological, kepner2019radix}) or dynamic (dynamic sparse training (DST) \cite{mocanu2018scalable, bellec2018deep}). By allowing the topology to be optimized along with the weights, sparse neural networks trained with DST have reached a comparable performance to the equivalent dense networks or even outperform them.

DST methods can be divided into two main categories based on the weight addition policy: $(1)$ \textit{Random regrowth}: Sparse Evolutionary Training (SET) \cite{mocanu2018scalable} is one of the earliest works that starts with a sparse neural network and perform magnitude pruning and random weight regrowing at each epoch to update the topology. In \cite{pmlr-v97-mostafa19a}, the authors proposed the idea of parameter reallocation automatically across layers during sparse training in CNNS. Many works have further studied sparse training concept recently \cite{ gordon2018morphnet, atashgahi2020quick,liu2021sparse,  liu2020topological, liu2021selfish, liu2021we}. 
$(2)$ \textit{Gradient information}: A group of works have tried to exploit gradient information to speed up the training process in DST \cite{raihan2020sparse}. \citeauthor{dettmers2019sparse} \cite{dettmers2019sparse} used the momentum of the non-existing connections as a criterion to grow weights instead of random addition in the SET algorithm; While being effective in terms of the accuracy, this method requires computing gradients and updating the momentum for all non-existing parameters. The Rigged Lottery (RigL) \cite{evci2020rigging} addressed the high computational cost by using infrequent gradient information. However, it still requires the computational cost for computing the periodic dense gradients. \cite{jayakumar2020top} tried to further improve RigL by using the gradient for only a subset of non-existing weights. In \cite{dai2019nest}, authors exploit gradient information in the search for a performant sub-network and discuss that gradient-based weight addition is biologically plausible.

%%%%%%%%%%%%%%%%%%%%%%%%%%%%%%%%%%%%%%%%%%%%%%%%%%%%%%%%%%%%%%%%%%%%%%%%%%%%%

\subsection{Hebbian Learning Theory}\label{ssec:background_hebb}
\looseness=-1
The Hebbian learning rule was proposed in 1949 by \citeauthor{hebb2005organization} as the learning rule for neurons \cite{hebb2005organization} inspired by biological systems. It describes how the neurons' activations influence the connections among them. The classical Hebb's rule indicates \enquote{neurons that fire together, wire together}. This can be formulated as $\Delta w_{ij}=\eta p_iq_j$, where $\Delta w_{ij}$ is the change in synaptic weight $w_{ij}$ between two neurons $p_i$ (presynaptic) and $q_j$ (postsynaptic) in two consecutive layers, and $\eta$ is the learning rate. While some previous works have adapted Hebb's rule to some machine learning tasks, \cite{scellier2016towards, liu2017modeling}, it has not been vastly investigated in many others, particularly in the sparse neural networks. By adapting Hebb's rule to artificial neural networks, we can obtain powerful models that might be close to the function of structures found in neural systems of various species \cite{kuriscak2015biological}. In \cite{arora2014provable}, authors have incorporated the Hebbian learning theory to train a newly introduced neural network. In \cite{sun2016sparsifying}, the Hebbian learning concept has been used to sparsify the neural networks for face recognition; they drop the connections between the weakly correlated neurons. In \cite{dai2019nest}, authors proposed a gradient-based algorithm for obtaining a sparse neural network; they discuss the gradient-based connection growth policy is mathematically close to the Hebbian learning theory. In this work, by taking inspiration from the Hebbian learning theory, we aim to introduce a new sparse training algorithm for obtaining sparse neural networks.

%%%%%%%%%%%%%%%%%%%%%%%%%%%%%%%%%%%%%%%%%%%%%%%%%%%%%%%%%%%%%%%%%%%%%%%%%%%%%
\subsection{Cosine Similarity}
\looseness=-1
In most machine learning problems, the Euclidean distance is a common tool to measure the distance due to its simplicity. However, the Euclidean distance is highly sensitive to the vectors' magnitude \cite{XIA201539}. Cosine similarity is another metric that addresses this issue; it measures the similarity of the shapes of two vectors as the cosine of the angle between them. In other words, it determines whether the two vectors are pointing in the same direction or not \cite{han2012getting}. Due to its simplicity and efficiency, the cosine similarity is a widely used metric in machine learning and pattern recognition field \cite{XIA201539}. It often measures the document similarity in natural language processing tasks \cite{sidorov2014soft, li2013distance}. Cosine Similarity has proven to be an effective tool also in neural networks. In \cite{luo2018cosine}, to bound the pre-activations in a multi-layer neural network that might disturb the generalization, authors have proposed to use cosine similarity instead of the dot product and showed that it reaches a better performance than the simple dot product. In \cite{nguyen2010cosine}, authors have used this metric to improve face verification using deep learning.

%% file: sections/3methodology.tex
% Proposed Method
%%%%%%%%%%%%%%%%%%%% %%%%%%%%%%%%%%%%%%%%%%%%%%%%%%%%%
% section intro
%%%%%%%%%%%%%%%%%%%%%%%%%%%%%%%%%%%%%%%%%%%%%%%%%%%%%%
\section{Proposed Method}
\looseness=-1
In this section, we first formulate the problem. Secondly, we demonstrate the cosine similarity as a tool for determining the importance of weights in neural networks and how it relates to the Hebbian learning theory. Finally, we present two new sparse training algorithms using cosine similarity-based connection importance.

%%%%%%%%%%%%%%%%%%%%%%%%%%%%%%%%%%%%%%%%%%%%%%%%%%%%%%%%%
% Problem Definition
%%%%%%%%%%%%%%%%%%%%%%%%%%%%%%%%%%%%%%%%%%%%%%%%%%%%%%%
\subsection{Problem Definition}\label{ssec:problem_definition}
\looseness=-1
Given a set of training samples $\sX$ and target output $ \vy$, a dense neural network is trained to minimize $\label{eq:cost_func}
   J(\vtheta) = \frac{1}{m} \sum_{i=1}^{m} L( f(\vx^{(i)} ; \vtheta), \vy^{(i)}),$
where  $m$ is the number of training samples, $L$ is the loss function, $f$ is a neural network parametrized by $\vtheta$, $ f(\vx^{(i)} ; \vtheta)$ is the predicted output for input $ \vx^{(i)}$, and $\vy^{(i)}$ is the true label. $\vtheta \in \R^{N}$ is consisted of parameters of each layer $ l \in\{1,2, ..., H\}$ of the network as $\vtheta^l \in \R^{N^l}$, where $N^l = n^{l-1}\times n^l$ is the number of parameters of  layer $l$, $n^l$ is number of neurons at layer $l$, and the total number of parameters of the dense network is $N$. A sparse neural network, however, uses only a subset of $\vtheta^l$, and discards $s^{l}$ fraction of parameters of each layer $\vtheta^l$ (their weight values are equal to zero); $s^{l}$ is referred to as the \textit{sparsity} of layer $l$. The overall sparsity of the network is $S = 1- D$, where $ D = \frac{\sum_{l = 1}^H{(1-s^l)N^l}}{N}$ is the overall density of the network. We aim to obtain a sparse neural network with sparsity level of $S$ and parameters $\vtheta$. We aim to train this network to minimize the loss on the training set as follows:
\begin{equation}
        \mathbb{\vtheta}^{*} = \argmin_{ \vtheta \in \R^{N},\;  \norm{\vtheta}_0=D\times N} \frac{1}{m} \sum_{i=1}^{m} L( f(\vx^{(i)} ; \vtheta), \vy^{(i)}),
        \end{equation}
where $\norm{\vtheta}_0$ is the total number of non-zero connections of the network which is determined by the density level.\\
\looseness=-1
\textbf{Network Structure.} The architecture we consider is a Multi-layer Perceptron (MLP) with $H$ layers. Initially, sparse connections between two consecutive layers are initialized with an Erdős–Rényi random graph; each connection in this graph exists with a probability of $ P( \theta^{l}_{i}) = \frac{\varepsilon( n^{l-1}+ n^l)}{ n^{l-1}  n^l},\; i \in \{1, 2, ..., N^l\},
$ where $ \varepsilon \in \sR^+$ denotes the hyperparameter that controls the sparsity level. The lower the value of $\varepsilon$ is, the sparser the network would be. In other words, by increasing $\varepsilon$, the probability of $ P( \theta^{l}_{i}) $ would be higher which results in more connections and a denser network. Each existing connection is initialized with a small value from a normal distribution.

%%%%%%%%%%%%%%%%%%%%%%%%%%%%%%%%%%%%%%%%%%%%%%%%%%%%%%%%%%%%%%%%%%%%%%%%%%%%%%%
%% Cosine Similarity to Determine Connections Importance
%%%%%%%%%%%%%%%%%%%%%%%%%%%%%%%%%%%%%%%%%%%%%%%%%%%%%%%%%%%%%%%%%%%%%%%%%%%%%%%
\subsection{Cosine Similarity to Determine Connections Importance}\label{ssec:cosine_in_nn}
\looseness=-1
In this paper, we use the cosine similarity as a metric to derive the importance of non-existing connections and evolve the topology of a sparse neural network. We first demonstrate how we measure cosine similarity of two neurons. Then, we argue why this choice has been made and how it relates to the Hebbian Learning theory. We measure the similarity of two neurons $p$ and $q$ as:
\begin{equation}\label{eq:cosine_similarity}
 \emSim_{p,q}^{l} = \abs{\frac{ \mA_{:, p}^{l-1} \cdot  \mA_{:, q}^{l}}{\norm{ \mA_{:, p}^{l-1}}\norm{ \mA_{:, q}^{l}}}},
\end{equation}
where $\mSim^{l}$ is the similarity matrix between neurons in two successive layers $ l-1$ and $l$. $ \mA_{:, p}^{l-1}$ and $\mA_{:, q}^{l} \in \R^{m}$ are the activation vectors corresponding to neurons $ p$ and $ q$ in layers $ l-1$ and $ l$, respectively. If $\emSim_{p,q}^{l}$ is high for two unconnected neurons (close to $1$), it means that they have a high similarity among their activations; therefore, we prefer to add a connection between them as it suggests that this path contains important information about data. However, if $\emSim_{p,q}^{l}$ is low for two neurons (close to $0$), it means that the activations of neurons $p$ and $q$ are not similar, and the connection among them might not be beneficial for the network.

\looseness=-1
We now argue why cosine similarity can be used to measure the importance of a non-existing connection in sparse neural networks and how it connects to the Hebbian learning theory. Basically, by taking inspiration from the Hebbian learning theory, we aim to rewire the neurons that fire together in the context of sparse training algorithms, instead of only strengthening the existing connections among neurons that fire together \cite{schumacher2021livewired}. It has been discussed in \cite{schumacher2021livewired} that connecting a pair of neurons with strong coincident activations can be viewed as a natural extension of the Hebbian learning; it is necessary to wire the neurons that usually fire together in order to understand better the relationship among the higher-order representation of those neurons. If a causal connection between their higher-order representation does exist, growing a connection among them will enable an effective inference about the relationship between them. Therefore, we need to discover which pairs of neurons usually fire together and then rewire them.

We employ cosine similarity to measure the relation between the activation values of two neurons. Such as the Hebb's rule (Section \ref{ssec:background_hebb}), the importance of a connection in our method is also determined by multiplying the activations of its corresponding neurons, albeit normalized; in Equation \ref{eq:cosine_similarity}, $ \mA_{:, p}^{l-1}$ is the presynaptic activation and $ \mA_{:, q}^{l}$ is the postsynaptic activation. If the activations of two connected neurons agree, by computing the dot product of activations, both Equation \ref{eq:cosine_similarity} and Hebb's rule assign higher importance to the corresponding connection. This would result in increased weight and a better chance of adding this connection. Thus, both
methods reward connections between neurons that exhibit
similar behavior. As mentioned earlier, the main difference between the Hebb's rule and Equation \ref{eq:cosine_similarity} is normalization. We will discuss in Section \ref{ssec:discussion_normalization} why the normalization step is necessary for evolving the topology of a sparse neural network.

In summary, if the cosine similarity of the activation vector of two neurons is high, it indicates the necessity of the connection between them in the network's performance. Therefore, we use the cosine similarity information to find out if the link between a pair of neurons should be rewired or not. Based on this knowledge, we propose two new algorithms to evolve the sparse neural network in the following sections.

%%%%%%%%%%%%%%%%%%%%%%%%%%%%%%%%%%%%%%%%%%%%%%%%%%%%%%%%%%%%%%%%%%%%%%%%%%%%%%%
%% Sequential Cosine Similarity-based and Random Topology Exploration (CTRE_{seq})
%%%%%%%%%%%%%%%%%%%%%%%%%%%%%%%%%%%%%%%%%%%%%%%%%%%%%%%%%%%%%%%%%%%%%%%%%%%%%%%
\subsection{Sequential Cosine Similarity-based and Random Topology Exploration (\texorpdfstring{CTRE\textsubscript{seq}}))}\label{ssec:CTRE_seq}
\looseness=-1
Our first proposed algorithm, Sequential Cosine Similarity-based and Random Topology Exploration (CTRE\textsubscript{seq}) evolves the network topology using both cosine similarity between neurons of each pair of consecutive layers in the network and random search. Overall, in the beginning, at each training epoch, it removes unimportant connections based on their magnitude and adds new connections to the network based on their cosine similarity. When the network performance stops improving, the algorithm switches to random topology search. In the following, we will explain the algorithm in more detail.

\looseness=-1    
After initializing the sparse network with sparsity level determined by $\varepsilon$, the training begins. The training procedure consists of two consecutive phases: \textit{1. Cosine Similarity-based Exploration}: The training starts with this phase in which each epoch includes three steps: $(a)$ Firstly, a standard feed-forward and back-propagation are performed. $(b)$ Then, a proportion $\displaystyle \zeta$ of connections with the lowest magnitude in each layer is removed. In Section \ref{ssec:weight_removal}, we further discuss why this choice has been made. $(c)$ Subsequently, we add new connections to the network based on the neurons' similarity. Taking advantage of the cosine similarity metric, we measure the similarity of two neurons as formulated in Equation \ref{eq:cosine_similarity}. In each layer, we add connections (as many connections as the removed connections in this layer) with the highest similarity between the corresponding neurons; the new connections are initialized with a small value from a uniform distribution. \textit{2. Random Exploration}: The second phase begins when the performance of the network on a validation set does not improve in $e_{early\;stop}$ epochs ( $e_{early\;stop}$ is a hyperparameter of CTRE\textsubscript{seq}). This is due to the fact that the activation values might not change significantly after some epochs and, consequently, the similarity of neurons. As a result, the topology search using cosine similarity might stop as well. To prevent this, we begin a random search when the classification accuracy on the validation set stops increasing. This phase is almost similar to phase $1$, and they are different in the weight regrowing policy. In this phase, instead of using cosine similarity information, we add connections randomly to the network. In this way, we prevent early stopping of the topology search. Algorithm \ref{alg:CTRE_seq} summarizes this method.

\input{supplementary/algorithms/alg_ctre_seq}

%%%%%%%%%%%%%%%%%%%%%%%%%%%%%%%%%%%%%%%%%%%
\subsection{Simultaneous Cosine Similarity-based and Random Topology Exploration (\texorpdfstring{CTRE\textsubscript{sim}}))}\label{ssec:CTRE_sim}
%%%%%%%%%%%%%%%%%%%%%%%%%%%%%%%%%%%%%%%%%%
To constantly exploit the cosine similarity information during training and avoid early stopping of topology exploration, we propose another method for obtaining a sparse neural network, named Simultaneous Cosine Similarity-based and Random Topology Exploration (CTRE\textsubscript{sim}).

\looseness=-1
Prior to the training, we initialize a sparse neural network. After that, the training procedure starts with three steps in each epoch. The first two steps are similar to the CTRE\textsubscript{seq}, which are $(a)$ standard feed-forward and back-propagation, and $(b)$ magnitude-based weight removal. However, in step $(c)$, instead of relying solely on cosine similarity information or random addition, we combine both strategies. There are two reasons behind this choice: $(1)$ As discussed in Section \ref{ssec:CTRE_seq}, as the training proceeds, the activation values become stable and might not change significantly after a while and, consequently, the similarity values. In CTRE\textsubscript{seq}, we addressed this issue by switching completely to random search. However, the training speed might slow down if we rely only on the random search. $(2)$ If we rely only on cosine similarity information, there is a possibility to add some connections based on the similarity of the neurons, which have been removed based on the magnitude in the weight removal step. It means that in these cases, the path between these pairs of similar neurons does not contribute to the performance of the network. Therefore, we should not add such connections to the network. These are the potential limitations of CTRE\textsubscript{seq}.

\looseness=-1
To address these limitations, CTRE\textsubscript{sim} takes another approach to prevent adding the removed connections which have a high cosine similarity to the network, as follows. In step $c$, we add the connections with high similarities to the network; however, if some connections with high cosine similarity are earlier removed based on their magnitude in step $b$, we add random connections to the network. In other words, we split our budget between similarity-based and random exploration. More importantly, we let the network dynamically decide how much budget should be allocated to each exploration at each epoch. The benefits from this approach are twofold; we prevent early stopping of the topology search, and also prevent re-adding connections that have shown to be unhelpful for the network's performance. Algorithm \ref{alg:CTRE_sim} summarizes this method.

%% file: supplementary/algorithms/alg_ctre_seq.tex
\begin{figure}[!t]
%\vskip -0.4 in
\begin{minipage}[t]{0.49\textwidth}
\begin{algorithm}[H]
    \caption{CTRE\textsubscript{seq}}
    \label{alg:CTRE_seq}
     \scriptsize
    \begin{algorithmic}[1]
        \State \textbf{Input}: Dataset $\displaystyle \sX$, sparsity hyperparameter $\varepsilon$, drop fraction $\zeta$, early stop epoch $e_{early\;stop}$
        \State Initialize the network with sparsity determined by $\varepsilon$,  $flag_{random\;search} = False$
        \For{\texttt{$i \in \{1,\dots, \#epochs\}$}}
            \State perform standard feed-forward and back-propagation
            \For{\texttt{$l \in \{1,\dots, H\}$}}
                \State Remove $\zeta N^l$ of the weights with smallest magnitude.
                \If{$flag_{random\;search}$}
                    \State Add $\zeta N^l$ connections randomly
                \Else
                    \State Compute Similarity matrix $\displaystyle \mSim^{l}$ according to Equation \ref{eq:cosine_similarity}
                    \State \multiline{%
                    Add $\zeta N^l$ connections with the highest similarity value in $\displaystyle \mSim^{l}$}
                \EndIf
            \EndFor
        %    \State \texttt{${{s}_{i}}=\sum\limits_{j=1}^{n^h}{|{{W}^{1}_{ij}}}|$
        %    }
        \If{Accuracy on validation set does not improve in $e_{early\;stop}$}
            \State Set $flag_{random\;search} = True$
        \EndIf
        \EndFor
        \end{algorithmic}

    \end{algorithm}

\end{minipage}
\hfill
\begin{minipage}[t]{0.49\textwidth}

\begin{algorithm}[H]
    \caption{CTRE\textsubscript{sim }}
    \label{alg:CTRE_sim}
     \scriptsize
    \begin{algorithmic}[1]
        \State \textbf{Input}: Dataset $\displaystyle \sX$, sparsity hyperparameter $\varepsilon$, drop fraction $\zeta$
        \State Initialize the network with sparsity determined by $\varepsilon$
        \For{\texttt{$i \in \{1,\dots, \#epochs\}$}}
            \State perform standard feed-forward and back-propagation
            \For{\texttt{$l \in \{1,\dots, H\}$}}
                \State Remove $\zeta N^l$ of the weights with smallest magnitude.
                \State Compute Similarity matrix $\displaystyle \mSim^{l}$ according to Equation \ref{eq:cosine_similarity}
                \State \multiline{%
                $C_{sim}$ = Set of $\zeta N^l$ connections with the highest similarity value in $\displaystyle \mSim^{l}$}
                \For{each \texttt{$c \in C_{sim}$}} 
                    \If{$c$ was removed in the last weight removal step}
                        \State Add a random connection to the network
                    \Else
                        \State Add connection $c$ to the network
                    \EndIf 
                \EndFor
            \EndFor

        \EndFor
        \end{algorithmic}
 
    \end{algorithm}

\end{minipage}
\vskip -0.2 in
\end{figure}

%% file: sections/4Experiments.tex
% Experiments
\section{Experiments and Results}
\looseness=-1
In this section, we evaluate our proposed algorithms and compare them with several state-of-the-art algorithms for obtaining a sparse neural network. First, we describe the settings of the conducted experiments, including the hyperparameter values, implementation details, and datasets. Then, we compare them in terms of the classification accuracy on several datasets and networks with different sizes and sparsity levels. 

%%%%%%%%%%%%%%%%%%%%%%%%%%%%%%%%%%%%%%%%%%%%%%%%%%%%%%%%%%%%%%%%%%%%%%%%%%%%%%
%%%%%%%%%%%%%%%%%%%%%%%%%%%%%%%%%%%%%%%%%%%%%%%%%%%%%%%%%%%%%%%%%%%%%%%%%%%%%%
\subsection{Settings}\label{ssec:settings}
%%%%%%%%%%%%%%%%%%%%%%%%%%%%%%%%%%%%%%%%%%%%%%%%%%%%%%%%%%%%%%%%%%%%%%%%%%%%%%
%%%%%%%%%%%%%%%%%%%%%%%%%%%%%%%%%%%%%%%%%%%%%%%%%%%%%%%%%%%%%%%%%%%%%%%%%%%%%%
\looseness=-1
This section gives a brief overview of the experiment settings, including hyperparameter values, implementation details, and datasets used for the evaluation of the methods.     

%---------------------------------------------------------------------------  
\subsubsection{Hyperparameters.} \label{sssec:Hyperparameters}
%---------------------------------------------------------------------------  
The network that we use to perform experiments is a 3-layer MLP as described in Section \ref{ssec:problem_definition}. The activation functions used for hidden and output layers are \enquote{Relu} and \enquote{Softmax}, respectively, and the loss function used is \enquote{CrossEntropy}. The values for most hyperparameters have been selected using a grid search over a limited number of values. The hyperparameter $\zeta$ has been set to $0.2$. In Algorithm \ref{alg:CTRE_seq}, $e_{early\;stop}$ has been set to $40$. We train the network with Stochastic Gradient Decent (SGD) with momentum and L$_2$ regularizer. The momentum coefficient, the regularization coefficient, and learning rate are $0.9$, $0.0001$, and $0.01$, respectively. All the experiments are performed using $500$ training epochs. The datasets have been preprocessed using the Min-Max Scaler so that each feature is normalized between $0$ and $1$, except for Madelon, where we use standard scaler (each feature will have zero mean and unit variance). For the image datasets, data augmentation has not been performed unless it has been explicitly stated.

\looseness=-1
%--------------------------------------------------------------------------- 
\subsubsection{Comparison} \label{sssec:comparison}
%--------------------------------------------------------------------------- 
We compare the results with three state-of-the-art methods for obtaining sparse neural networks, including, SNIP, RigL, and SET. 

\begin{itemize}
    \item[$\bullet$] \textbf{SNIP} \cite{lee2018snip}. Single-shot network pruning (SNIP) is a dense-to-sparse sparsification algorithm that prunes the network prior to initialization based on connection sensitivity. It calculates this metric after a few iteration of dense training. After pruning, SNIP starts the training with the sparse neural network.
    \item[$\bullet$] \textbf{RigL} \cite{evci2020rigging}. The rigged lottery (RigL) is a sparse-to-sparse algorithm for obtaining a sparse neural network that uses the gradient information as the weight addition criteria.
    \item[$\bullet$] \textbf{SET} \cite{mocanu2018scalable}. Sparse evolutionary training (SET) is a sparse-to-sparse training algorithm that uses random weight addition for updating the topology.
\end{itemize}

Besides, we measure the classification performance of a fully-connected MLP as the baseline method.

\renewcommand\UrlFont{\small\rmfamily\itshape}
%--------------------------------------------------------------------------- 
%\vspace{-1 cm}
\looseness=-1
\subsubsection{Implementation} \label{sssec:Implementation}
%-----------------------------------------------------------------------
We evaluate our proposed methods and the considered baselines on eight datasets. We implemented our proposed method using Tensorflow \cite{tensorflow2015-whitepaper}. The baseline of this implementation is the RigL code from Github \footnote{The implementation of RigL, SNIP, and SET is available at \url{https://github.com/google-research/rigl}.}. It also includes the implementation for SNIP, SET, and fully-connected MLP. This code uses a binary mask over weights to implement sparsity. In addition, we provide a purely sparse implementation that uses Scipy library sparse matrices. This code is developed from the sparse implementation of SET, which is available on Github\footnote{The pure sparse implementation of SET  can be found on \\\url{https://github.com/dcmocanu/sparse-evolutionary-artificial-neural-networks}.}. For all the experiments, we use the Tensorflow implementation to have a fair comparison among methods. However, we provide the results using the sparse implementation in Appendix \ref{appendix:pure_sparse}. Most experiments were run on a CPU (Dell R730). For image datasets, we used a Tesla-P100 GPU. All the experiments were repeated with three random seeds. The only exception is the experiments from Section \ref{ssec:performance_evaluation} where we run 15 random seeds to analyze the statistical significance of the obtained results with respect to the considered algorithms (Section \ref{sssec:statistical_significance}). To ensure a fair comparison, for the sparse training methods (SET, RigL, and CTRE), the sparsity mask is updated at the end of each epoch, and drop fraction ($\zeta$) and learning rate are constant during training.

\input{supplementary/Tables/table_datasets}

%---------------------------------------------------------------------------  
\subsubsection{Datasets} \label{sssec:Datasets}
%---------------------------------------------------------------------------  

We conducted our experiments on eight benchmark datasets as follows:

\begin{itemize}[]
    % noitemsep,topsep=0pt
    %\itemsep-0.09em 
    \item[$\bullet$] \textbf{Madelon} \cite{guyon2008feature} is an artificial dataset with 20 informative features , and 480 noise features.
    \item[$\bullet$] \textbf{Isolet} \cite{fanty1991spoken} has been created with the spoken name of each letter of the English alphabet.
    \item[$\bullet$] \textbf{MNIST} \cite{lecun1998mnist} is a database of $28\times28$ images of handwritten digits.
    \item[$\bullet$] \textbf{Fashion\_MNIST} \cite{xiao2017} is a database of $28\times28$ images of Zalando's articles.
    \item[$\bullet$] \textbf{CIFAR10/100} \cite{krizhevsky2009learning} are two datasets of 32×32 colour images categorized in 10/100 classes.
    \item[$\bullet$] \textbf{PCMAC \& BASEHOCK} \cite{lang1995newsweeder} are two subsets of the 20 Newsgroups data.
\end{itemize}

More details about the datasets is presented in Table \ref{tab:datasets}.

\input{supplementary/Tables/table_num_hidden_epsilon_comparison2}

%%%%%%%%%%%%%%%%%%%%%%%%%%%%%%%%%%%%%%%%%%%%%%%%%
\subsection{Performance Evaluation}\label{ssec:performance_evaluation}
\begin{figure}[!t]
        \begin{center}
        \centerline{\includegraphics[width=\textwidth]{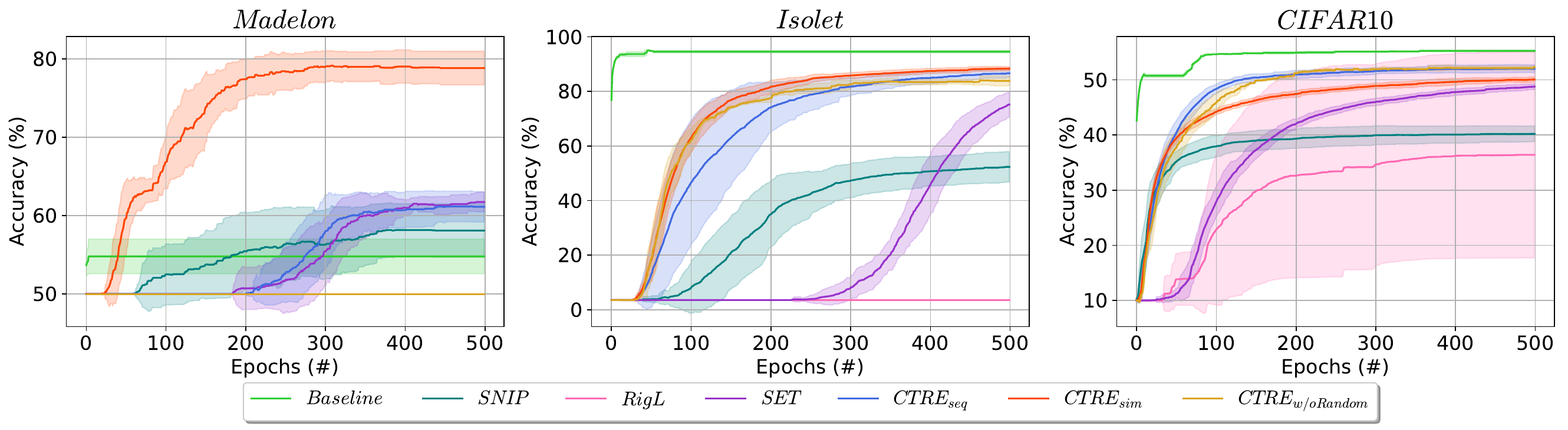}}
        \caption{Classification accuracy (\%) comparison among methods on a highly large and sparse 3-layer MLP with a density lower than $0.3\%$ ($n^l=1000$, $\varepsilon=1$). }
        \label{fig:speed}
        \end{center}
    \end{figure}

%%%%%%%%%%%%%%%%%%%%%%%%%%%%%%%%%%%%%%%%%%%%%%%
 \looseness=-1
In this experiment, we compare the methods in terms of classification accuracy on networks with varying sizes and sparsity levels. We consider three MLPs, each having three hidden layers with $100$, $500$, and $1000$ hidden neurons, respectively. By changing the value of $\varepsilon$ for each MLP, we study the effect of sparsity level on the performance of the methods. Table \ref{tab:results_comparison_rigl_implementation} summarizes the results of these experiments that are carried out on the five datasets, including tabular and image datasets that have different characteristics. We have also included the density (as percentage) and the number of connections (divided by $10^3$) for each network in this table. For training on each dataset, we allocate $10\%$ of the training set to a validation set. During training, each MLP is trained on the new training set. At each epoch, we measure the performance on the validation set. Finally, Table \ref{tab:results_comparison_rigl_implementation} presents the results of each algorithm on an unseen test set and using the model that gives the highest validation accuracy during training. The learning curves regarding each case are presented in Appendix \ref{appendix:learning_curves}; however, we present some interesting cases in Figure \ref{fig:speed}.

\looseness=-1
First, we analyze the performance of methods on the two tabular datasets. As can be seen in Table \ref{tab:results_comparison_rigl_implementation}, on Madelon dataset, CTRE\textsubscript{sim} is the best performer in most cases. Interestingly, the accuracy increases when the network becomes sparser. However, this can be explained intuitively; since the Madelon dataset contains many noise features ($> 95\%$), the higher the number of the connections is, the higher the risk for over-fitting the noise features will be. CTRE\textsubscript{sim} can find the most important information paths in the network, which most likely start from the input neurons corresponding to the informative features. As a result, it can reach an accuracy of $78.8\%$ with only $0.3\%$ of total connections of the equivalent dense network ($n^l=1000$), while the maximum accuracy achieved by other methods considered is $61.9\%$ (SET). On the second tabular dataset, Isolet, CTRE\textsubscript{sim} is the best performers on two very sparse models, including $0.4\%$ $(n^l=500)$ and $0.3\%$ $(n^l=1000)$ densities. In addition, in all the other cases, CTRE\textsubscript{sim} and CTRE\textsubscript{seq} are the second and third-best performers. In terms of learning speed, we can observe in Figure \ref{fig:speed} that CTRE\textsubscript{sim} can find a good topology much faster than other methods, which results in an increase in the accuracy within a short period after the training starts. From Figure \ref{fig:speed}, it can be seen that RigL fails to find an informative sub-network in these cases ($D<0.3\%$). This indicates that gradient information might not be informative in highly sparse networks.

\looseness=-1
On the image dataset, CTRE\textsubscript{sim} and CTRE\textsubscript{seq} are the best and second-best performers in most of the cases considered. When the network size is small ($n^l=100$), SET is the major competitor of CTRE. However, when the model size increases, CTRE outperforms SET. This indicates that the pure random weight addition policy in SET can perform well in networks with a higher density, while it is hard to find such sub-network randomly in high sparsity scenarios due to the very large search space. RigL also has a comparable performance to SET, except for very sparse models. As discussed in the previous paragraph, on a highly sparse network ($D<0.3\%$), RigL has poor performance. Besides, as shown in Figure \ref{fig:speed}, SNIP starts with a steep increase in the accuracy due to the few iterations of training a dense network and thus, starting with good topology. However, as the training proceeds, this topology cannot achieve the same performance as other methods. Therefore, it indicates that dynamic weight update is an essential factor in the sparse training of neural networks.

\looseness=-1
These observations confirm that the cosine similarity is an informative criterion for adding weight in the network compared to random (SET) and gradient-based addition (RigL) in very sparse neural networks. CTRE can reach a better performance than state-of-the-art sparse training algorithms in terms of learning speed and accuracy when the network is highly sparse. Besides, by comparing the results with the dense network, it is clear that it is possible to reach a comparable performance to the dense network even with a network with  $100$ times fewer connections which is an excellent choice for low-resource devices on edge. We further compare the computational cost of the algorithms in Appendix \ref{appendix:computational_complexity} and their learning speed in Appendix \ref{appendix:learning_curves}.

\subsubsection{Statistical Significance Analysis}\label{sssec:statistical_significance}
\input{supplementary/Tables/table_statistical_significance}
In this section, we analyze the statistical significance of the results obtained by CTRE compared to the other algorithms. To measure this, we perform Kolmogorov-Smirnov test (KS-test). The null hypothesis is that the two independent results/samples are drawn from the same continuous distribution. If the p-value is very small (p-value $< 0.05$), it suggests that the difference between the two sets of results is significant and the hypothesis is rejected. Otherwise, the obtained results are close together and the hypothesis is true. 

We perform KS-test between the results obtained by CTRE (for simplicity, we consider maximum results of $CTRE_{seq}$ and $CTRE_{sim}$) and the other considered algorithms for the experiments in Table \ref{tab:results_comparison_rigl_implementation}. The results of the KS-test is summarized in Table \ref{tab:results_statistical_significance}. In this table, \textit{Reject} shows that the results are sufficiently distinct and \textit{True} means that the obtained results are close together. The * sign in Table \ref{tab:results_statistical_significance} shows that an algorithm has achieved the maximum accuracy in the corresponding experiment. Finally, the entries colored as red shows an experiment where a compared method obtains a close result to CTRE while having lower mean accuracy.  

From Table \ref{tab:results_statistical_significance}, we can observe that in majority of the experiments, CTRE obtains higher mean accuracy than the other methods while being statistically different from them. The only dataset where the results in most cases are close is the Fashion-MNIST dataset where SET has comparable results to CTRE on this dataset. In addition, in high sparsity regime and large network size ($n^l=1000$, $\varepsilon = 1$), CTRE achieves the highest accuracy among the methods while being significantly distinct from them. Overall, Table \ref{tab:results_statistical_significance} indicates that CTRE is a well performing algorithm in terms of the classification accuracy that achieves significantly different results from the other methods.

\begin{figure}[!t]
    %\vskip -0.4in
    \begin{center}
    \centerline{\includegraphics[width=\textwidth]{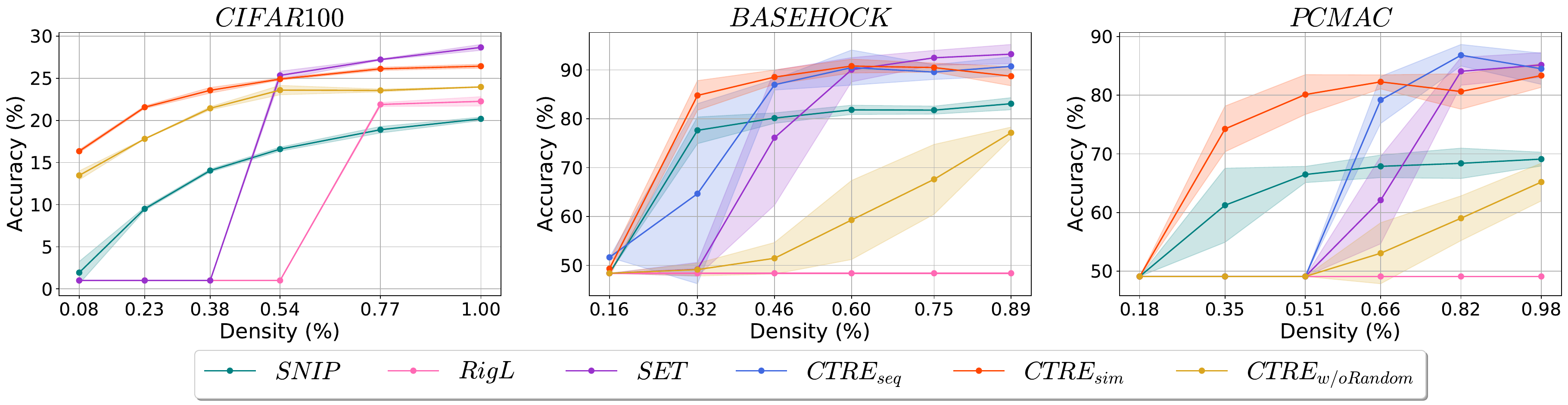}}
    \caption{Sparsity-accuracy trade-off on highly sparse neural  networks on three datasets.}
    \label{fig:high_sparsity_1_percent}
    \end{center}
\end{figure}

%%%%%%%%%%%%%%%%%%%%%%%%%%%%%%%%%%%%%%%%%%%%%%%%%%%%%%%%%%%%%%%%%%%%%%%%%%%%%%
%%%%%%%%%%%%%%%%%%%%%%%%%%%%%%%%%%%%%%%%%%%%%%%%%%%%%%%%%%%%%%%%%%%%%%%%%%%%%%

\subsection{Sparsity-Performance Trade-off Analysis in Highly Sparse MLPs}
%%%%%%%%%%%%%%%%%%%%%%%%%%%%%%%%%%%%%%%%%%%%%%%%%%%%%%%%%%%%%%%%%%%%%%%%%%%%%%
%%%%%%%%%%%%%%%%%%%%%%%%%%%%%%%%%%%%%%%%%%%%%%%%%%%%%%%%%%%%%%%%%%%%%%%%%%%%%%
\looseness=-1
We carry out another experiment to study the trade-off between sparsity and accuracy on very high sparsity cases. We perform this experiment for two difficult classification tasks including, image classification on CIFAR100, which is considered as a more difficult dataset than the earlier considered image datasets, and text classification on PCMAC and BASESHOCK that are subsets of 20-newsgroup dataset; they have a high number of features and a low number of samples. This experiment uses a 3-layer MLP with $1000$ and $3000$ hidden neurons for text datasets and CIFAR100 dataset, respectively. We change the density value between $0$ and $1$ and compare our proposed approaches to SNIP, RigL, and SET (due to the close performance of CTRE\textsubscript{sim} and CTRE\textsubscript{seq} on earlier considered image datasets, on CIFAR100, we perform the experiments with CTRE\textsubscript{sim}). We use data augmentation for CIFAR100. Also, as the network is considerably large on this dataset, we set the learning rate to $0.05$ to speed up the training. The results are presented in Figure \ref{fig:high_sparsity_1_percent}. 

\looseness=-1
As shown in Figure \ref{fig:high_sparsity_1_percent}, in highly sparse networks ($D<0.5\%$), CTRE\textsubscript{sim} outperforms other methods by a large gap. As discussed in Section \ref{ssec:performance_evaluation}, RigL performs poorly in these scenarios. SNIP outperforms SET and RigL at the very low densities while still has lower results than CTRE\textsubscript{sim} in all cases. While SET outperforms other methods for larger density values on CIFAR100 and BASESHOCK, it performs poorly on a very sparse network. On text datasets, CTRE\textsubscript{seq} has comparable performance to CTRE\textsubscript{sim} and SET on higher densities, and it achieves the highest accuracy on PCMAC. Overall, we can observe that CTRE\textsubscript{sim} has decent performance on these three datasets with a density value between $0.3\%$ and $0.5\%$.

%% file: supplementary/Tables/table_datasets.tex
       \begin{table}[!t]
        \centering
    
        \caption{Datasets characteristics.}
        \label{tab:datasets}
        \begin{scriptsize}
        \begin{tabularx}{0.65\textwidth}{c@{\hskip 0.07in}c@{\hskip 0.07in}c@{\hskip 0.07in}c@{\hskip 0.07in}c@{\hskip 0.07in}c@{\hskip 0.07in}c}
            \toprule

            \multicolumn{1}{c}{ \textbf{Dataset}} & \textbf{Dimensions} & \textbf{Type} & \textbf{Samples} & \textbf{Train} & \textbf{Test} & \textbf{Classes} \\  \midrule
            Isolet & 617 & Speech & 7737 & 6237 & 1560 & 26   \\ 
             Madelon & 500 & Artificial & 2600 & 2000 & 600 & 2  \\
            MNIST & 784 & Image & 70000 & 60000 & 10000 & 10   \\ 
            Fashion\_MNIST & 784 & Image & 70000 & 60000 & 10000 & 10   \\ 
            CIFAR10 & 3072  &Image  &60000  &50000  &10000  & 10   \\ 
            CIFAR100 & 3072  &Image  &60000  &50000  &10000  & 100   \\ 
            PCMAC & 3289 & Text & 1943 & 1554 & 389 & 2  \\ 
            BASEHOCK &4862 & Text &1993 &1594 &399 &2\\
            \bottomrule
            \end{tabularx}
            \end{scriptsize}
        \end{table}

%% file: supplementary/Tables/table_num_hidden_epsilon_comparison2.tex
\newrobustcmd{\bt}{\bfseries} % bold command
\begin{table}[!t]
    %\vskip -0.2 in
    \caption{Classification accuracy (\%) comparison among methods on networks with various sizes and sparsity levels. The density (\%) and number of connections for each case is indicated in the table. Please note that N (total number of parameters of the network) is scaled by $\times10^3$.} \label{tab:results_comparison_rigl_implementation}

    \begin{scriptsize}
    \scalebox{0.88}{ 
    \begin{tabular}{@{}c@{\hskip 0.02in}c@{\hskip 0.05in}c@{\hskip 0.05in}c@{\hskip 0.05in}c@{\hskip 0.07in}c@{\hskip 0.05in}c@{\hskip 0.05in}c@{\hskip 0.07in}c@{\hskip 0.05in}c@{\hskip 0.05in}c@{}}
        \toprule
         &  & \multicolumn{3}{c}{\pmb{ $n^l=100$}} & \multicolumn{3}{c}{\pmb{ $n^l=500$}} & \multicolumn{3}{c}{\pmb{ $n^l=1000$}} \\ \cmidrule(l){3-5}\cmidrule(l){6-8}\cmidrule(l){9-11}
         &  & \multicolumn{3}{c}{$\epsilon$} & \multicolumn{3}{c}{$\epsilon$} & \multicolumn{3}{c}{$\epsilon$}\\ %\cmidrule(c){3-5}
         %\cmidrule(c){6-8} \cmidrule(c){9-11}  \cmidrule(c){12-14}\\
        \bt Dataset & \bt Method &\bt 1 &\bt 5 & \bt13 & \bt1 & \bt5 & \bt13 & \bt1 & \bt5 & \bt13  \\ \midrule

\multirow{3}{*}{Madelon}&Baseline ($ N(\times10^3)$)&\multicolumn{3}{c}{$54.9\pm1.0$ (70.2)}&\multicolumn{3}{c}{$54.9\pm1.0$ (751.0)}&\multicolumn{3}{c}{$54.9\pm1.0$ (2502.0)}\\
\cmidrule(l){3-11} 
\multicolumn{1}{c}{}&$D(\%)$  ($N (\times10^3)$)&$1.6$ ($1.1$)&$7.8$ ($5.5$)&$20.4$ ($14.3$)&$0.5$ ($3.5$)&$2.3$ ($17.5$)&$6.1$ ($45.5$)&$0.3$ ($6.5$)&$1.3$ ($32.5$)&$3.4$ ($84.5$)
\\ \cmidrule(l){3-11} 
\multicolumn{1}{c}{}&SNIP&$56.5\pm3.5$&$56.1\pm2.9$&$54.1\pm2.6$&$57.3\pm3.0$&$56.4\pm2.5$&$57.5\pm1.9$&$58.1\pm3.4$&$58.4\pm1.5$&$57.8\pm1.6$\\
\multicolumn{1}{c}{}&RigL&$60.4\pm3.1$&$59.7\pm1.8$&$59.3\pm2.2$&$51.3\pm3.5$&$61.9\pm2.0$&$60.0\pm2.0$&$50.0\pm0.0$&$61.5\pm3.0$&$61.1\pm1.7$\\
\multicolumn{1}{c}{}&SET&$60.4\pm2.1$&$57.8\pm3.2$&$58.1\pm1.7$&$61.4\pm2.6$&$59.4\pm2.4$&$57.9\pm3.9$&$61.7\pm1.2$&$59.0\pm1.8$&$58.2\pm2.3$\\
\multicolumn{1}{c}{}&CTRE\textsubscript{seq}&\pmb{$82.2\pm2.4$}&$72.5\pm2.0$&$63.9\pm1.2$&$61.2\pm2.4$&\pmb{$81.5\pm1.4$}&$71.8\pm2.0$&$61.1\pm1.9$&\pmb{$83.9\pm2.0$}&\pmb{$76.5\pm1.9$}\\
\multicolumn{1}{c}{}&CTRE\textsubscript{sim}&$81.6\pm1.3$&\pmb{$73.0\pm1.6$}&\pmb{$65.6\pm3.0$}&\pmb{$79.4\pm1.7$}&$77.7\pm1.4$&\pmb{$73.0\pm1.5$}&\pmb{$78.8\pm2.2$}&$78.5\pm1.0$&$74.6\pm1.4$\\
\midrule

\multirow{3}{*}{Isolet}&Baseline ($ N(\times10^3)$)&\multicolumn{3}{c}{$94.4\pm0.1$ (84.3)}&\multicolumn{3}{c}{$94.3\pm0.0$ (821.5)}&\multicolumn{3}{c}{$94.6\pm0.4$ (2643.0)}\\
\cmidrule(l){3-11} 
\multicolumn{1}{c}{}&$D(\%)$  ($N (\times10^3)$)&$1.5$ ($1.2$)&$7.4$ ($6.2$)&$19.2$ ($16.2$)&$0.4$ ($3.6$)&$2.2$ ($18.2$)&$5.8$ ($47.4$)&$0.3$ ($6.6$)&$1.3$ ($33.2$)&$3.3$ ($86.4$)
\\ \cmidrule(l){3-11} 
\multicolumn{1}{c}{}&SNIP&$48.9\pm5.8$&$90.1\pm1.1$&$92.8\pm0.7$&$49.2\pm8.6$&$89.9\pm1.0$&$92.3\pm0.6$&$52.4\pm5.6$&$89.7\pm0.8$&$92.0\pm0.6$\\
\multicolumn{1}{c}{}&RigL&$66.0\pm28.6$&$90.0\pm0.8$&$92.3\pm0.7$&$3.5\pm0.0$&$88.4\pm1.0$&$91.9\pm1.2$&$3.5\pm0.0$&$87.2\pm1.8$&$91.2\pm1.5$\\
\multicolumn{1}{c}{}&SET&\pmb{$89.1\pm1.2$}&\pmb{$94.2\pm0.7$}&\pmb{$94.9\pm0.4$}&$86.5\pm1.2$&\pmb{$93.2\pm0.7$}&\pmb{$94.7\pm0.5$}&$75.2\pm4.7$&\pmb{$92.9\pm0.5$}&\pmb{$94.4\pm0.8$}\\
\multicolumn{1}{c}{}&CTRE\textsubscript{seq}&$86.7\pm1.8$&$92.4\pm1.1$&$94.3\pm0.5$&$87.2\pm2.0$&$92.3\pm0.6$&$94.0\pm0.4$&$86.7\pm1.6$&$91.5\pm1.0$&$93.7\pm0.5$\\
\multicolumn{1}{c}{}&CTRE\textsubscript{sim}&$87.5\pm0.8$&$93.4\pm0.7$&$94.3\pm0.8$&\pmb{$87.8\pm1.1$}&$91.7\pm1.1$&$93.9\pm0.5$&\pmb{$88.3\pm0.7$}&$91.3\pm1.3$&$93.1\pm0.6$\\
\midrule

\multirow{3}{*}{MNIST}&Baseline ($ N(\times10^3)$)&\multicolumn{3}{c}{$97.9\pm0.0$ (99.4)}&\multicolumn{3}{c}{$98.2\pm0.1$ (897.0)}&\multicolumn{3}{c}{$98.2\pm0.1$ (2794.0)}\\
\cmidrule(l){3-11} 
\multicolumn{1}{c}{}&$D(\%)$  ($N (\times10^3)$)&$1.4$ ($1.4$)&$7.0$ ($7.0$)&$18.2$ ($18.1$)&$0.4$ ($3.8$)&$2.1$ ($19.0$)&$5.5$ ($49.3$)&$0.2$ ($6.8$)&$1.2$ ($34.0$)&$3.2$ ($88.3$)
\\ \cmidrule(l){3-11} 
\multicolumn{1}{c}{}&SNIP&$91.1\pm0.5$&$96.3\pm0.2$&$97.2\pm0.1$&$93.8\pm0.3$&$96.6\pm0.2$&$97.2\pm0.2$&$94.4\pm0.3$&$96.7\pm0.2$&$97.3\pm0.2$\\
\multicolumn{1}{c}{}&RigL&$94.5\pm0.3$&$96.4\pm0.2$&$97.2\pm0.2$&$95.9\pm0.3$&$96.7\pm0.2$&$97.0\pm0.1$&$96.0\pm0.2$&$96.6\pm0.2$&$96.9\pm0.2$\\
\multicolumn{1}{c}{}&SET&$95.6\pm0.3$&$97.1\pm0.1$&$97.6\pm0.1$&$95.9\pm0.2$&$97.4\pm0.1$&$97.8\pm0.1$&$95.8\pm0.1$&$97.4\pm0.2$&$97.7\pm0.1$\\
\multicolumn{1}{c}{}&CTRE\textsubscript{seq}&\pmb{$95.7\pm0.2$}&\pmb{$97.3\pm0.2$}&$97.7\pm0.1$&\pmb{$97.0\pm0.2$}&$97.6\pm0.2$&$97.8\pm0.1$&\pmb{$97.3\pm0.1$}&\pmb{$97.7\pm0.1$}&$97.8\pm0.1$\\
\multicolumn{1}{c}{}&CTRE\textsubscript{sim}&$95.5\pm0.2$&\pmb{$97.3\pm0.1$}&\pmb{$97.8\pm0.1$}&$96.4\pm0.2$&\pmb{$97.7\pm0.1$}&\pmb{$98.0\pm0.1$}&$96.6\pm0.2$&\pmb{$97.7\pm0.1$}&\pmb{$97.9\pm0.1$}\\
\midrule

\multirow{3}{*}{ \shortstack{Fashion-\\MNIST} }&Baseline ($ N(\times10^3)$)&\multicolumn{3}{c}{$88.3\pm0.2$ (99.4)}&\multicolumn{3}{c}{$89.8\pm0.1$ (897.0)}&\multicolumn{3}{c}{$90.1\pm0.1$ (2794.0)}\\
\cmidrule(l){3-11} 
\multicolumn{1}{c}{}&$D(\%)$  ($N (\times10^3)$)&$1.4$ ($1.4$)&$7.0$ ($7.0$)&$18.2$ ($18.1$)&$0.4$ ($3.8$)&$2.1$ ($19.0$)&$5.5$ ($49.3$)&$0.2$ ($6.8$)&$1.2$ ($34.0$)&$3.2$ ($88.3$)
\\ \cmidrule(l){3-11} 
\multicolumn{1}{c}{}&SNIP&$79.5\pm2.3$&$86.0\pm0.3$&$87.1\pm0.3$&$81.1\pm1.2$&$86.3\pm0.3$&$87.4\pm0.4$&$82.2\pm1.1$&$86.5\pm0.2$&$87.3\pm0.2$\\
\multicolumn{1}{c}{}&RigL&$84.8\pm0.2$&$86.4\pm0.3$&$87.2\pm0.2$&$86.1\pm0.2$&$86.4\pm0.3$&$86.5\pm0.3$&$86.1\pm0.3$&$86.3\pm0.4$&$86.6\pm0.4$\\
\multicolumn{1}{c}{}&SET&\pmb{$85.8\pm0.3$}&$87.4\pm0.4$&$87.8\pm0.3$&$86.5\pm0.2$&$87.6\pm0.2$&$88.0\pm0.2$&$86.3\pm0.2$&$87.8\pm0.3$&$87.8\pm0.2$\\
\multicolumn{1}{c}{}&CTRE\textsubscript{seq}&\pmb{$85.8\pm0.5$}&\pmb{$87.5\pm0.3$}&$87.4\pm0.3$&\pmb{$87.1\pm0.4$}&$87.9\pm0.3$&$88.0\pm0.2$&\pmb{$87.3\pm0.2$}&$88.0\pm0.3$&\pmb{$88.3\pm0.2$}\\
\multicolumn{1}{c}{}&CTRE\textsubscript{sim}&\pmb{$85.8\pm0.3$}&\pmb{$87.5\pm0.3$}&\pmb{$88.0\pm0.2$}&$86.4\pm0.5$&\pmb{$88.1\pm0.2$}&\pmb{$88.3\pm0.2$}&$86.5\pm0.3$&\pmb{$88.1\pm0.3$}&\pmb{$88.3\pm0.2$}\\
\midrule

\multirow{3}{*}{CIFAR10}&Baseline ($ N(\times10^3)$)&\multicolumn{3}{c}{$51.2\pm0.5$ (328.2)}&\multicolumn{3}{c}{$53.2\pm0.2$ (2041.0)}&\multicolumn{3}{c}{$55.2\pm0.2$ (5082.0)}\\
\cmidrule(l){3-11} 
\multicolumn{1}{c}{}&$D(\%)$  ($N (\times10^3)$)&$1.1$ ($3.7$)&$5.6$ ($18.4$)&$14.6$ ($47.9$)&$0.3$ ($6.1$)&$1.5$ ($30.4$)&$3.9$ ($79.1$)&$0.2$ ($9.1$)&$0.9$ ($45.4$)&$2.3$ ($118.1$)
\\ \cmidrule(l){3-11} 
\multicolumn{1}{c}{}&SNIP&$35.8\pm3.8$&$47.6\pm0.7$&$49.5\pm0.7$&$39.4\pm1.8$&$48.9\pm0.6$&$50.7\pm0.6$&$40.2\pm1.4$&$49.3\pm1.0$&$50.9\pm0.6$\\
\multicolumn{1}{c}{}&RigL&$46.0\pm0.6$&$49.3\pm0.6$&$50.5\pm0.5$&$46.4\pm9.8$&$50.0\pm1.0$&$49.9\pm1.1$&$36.4\pm18.7$&$50.3\pm1.1$&$49.9\pm0.8$\\
\multicolumn{1}{c}{}&SET&\pmb{$48.3\pm0.5$}&$49.9\pm0.5$&\pmb{$50.6\pm0.5$}&$49.4\pm0.3$&$50.6\pm0.4$&$51.4\pm0.6$&$48.8\pm0.5$&$50.6\pm0.4$&$51.1\pm0.6$\\
\multicolumn{1}{c}{}&CTRE\textsubscript{seq}&$47.9\pm0.8$&$50.1\pm0.5$&$50.1\pm0.5$&\pmb{$51.3\pm0.5$}&\pmb{$52.7\pm0.5$}&\pmb{$52.6\pm0.6$}&\pmb{$52.0\pm0.7$}&\pmb{$53.2\pm0.6$}&\pmb{$53.6\pm0.6$}\\
\multicolumn{1}{c}{}&CTRE\textsubscript{sim}&$48.2\pm0.4$&\pmb{$50.3\pm0.3$}&$50.5\pm0.4$&$50.6\pm0.4$&$52.6\pm0.7$&$52.5\pm0.7$&$50.0\pm0.4$&$52.7\pm0.5$&$53.5\pm0.5$\\
\midrule

\multirow{3}{*}{CIFAR100}&Baseline ($ N(\times10^3)$)&\multicolumn{3}{c}{$23.1\pm0.6$ (337.2)}&\multicolumn{3}{c}{$23.1\pm0.6$ (2086.0)}&\multicolumn{3}{c}{$23.1\pm0.6$ (5172.0)}\\
\cmidrule(l){3-11} 
\multicolumn{1}{c}{}&$D(\%)$  ($N (\times10^3)$)&$1.1$ ($3.8$)&$5.6$ ($18.9$)&$14.5$ ($49.0$)&$0.3$ ($6.2$)&$1.5$ ($30.9$)&$3.8$ ($80.2$)&$0.2$ ($9.2$)&$0.9$ ($45.9$)&$2.3$ ($119.2$)
\\ \cmidrule(l){3-11} 
\multicolumn{1}{c}{}&SNIP&$5.9\pm0.6$&$15.7\pm0.7$&$20.4\pm0.3$&$6.6\pm0.8$&$17.9\pm0.7$&$22.1\pm0.4$&$6.2\pm0.8$&$18.3\pm0.6$&$22.5\pm0.5$\\
\multicolumn{1}{c}{}&RigL&$7.4\pm4.4$&$19.9\pm0.5$&$21.4\pm0.4$&$1.0\pm0.0$&$21.0\pm0.5$&$21.4\pm0.5$&$1.0\pm0.0$&$1.0\pm0.0$&$20.4\pm0.6$\\
\multicolumn{1}{c}{}&SET&\pmb{$14.7\pm0.4$}&$20.3\pm0.3$&$21.7\pm0.3$&$14.3\pm1.5$&$22.7\pm0.3$&$24.1\pm0.4$&$1.0\pm0.0$&$23.3\pm0.3$&$24.5\pm0.3$\\
\multicolumn{1}{c}{}&CTRE\textsubscript{seq}&$12.7\pm0.7$&\pmb{$21.1\pm0.3$}&$21.8\pm0.5$&\pmb{$18.7\pm0.4$}&\pmb{$23.6\pm0.4$}&$24.0\pm0.4$&\pmb{$21.4\pm0.4$}&\pmb{$23.9\pm0.5$}&$24.7\pm0.4$\\
\multicolumn{1}{c}{}&CTRE\textsubscript{sim}&$13.8\pm0.4$&$20.6\pm0.4$&\pmb{$21.9\pm0.4$}&$17.0\pm0.3$&$23.0\pm0.3$&\pmb{$24.6\pm0.4$}&$17.3\pm0.4$&$23.5\pm0.3$&\pmb{$25.1\pm0.3$}\\
\midrule

    \end{tabular}  }
    \end{scriptsize}
    %\vskip -0.2 in
\end{table}

%% file: supplementary/Tables/table_statistical_significance.tex
\begin{table}[!b]
    %\vskip -0.2 in
    \caption{Statistical significance of the results. Each entry show the result of KS-test among the results of CTRE and the compared methods for a specific network size and sparsity level. \textit{Reject} shows that the results are distinct, and \textit{True} indicates that the results are close together.} \label{tab:results_statistical_significance}

    \centering
    \begin{scriptsize}
    \scalebox{0.88}{ 
    \begin{tabular}{@{}c@{\hskip 0.02in}c@{\hskip 0.05in}c@{\hskip 0.05in}c@{\hskip 0.05in}c@{\hskip 0.07in}c@{\hskip 0.05in}c@{\hskip 0.05in}c@{\hskip 0.07in}c@{\hskip 0.05in}c@{\hskip 0.05in}c@{}}
        \toprule
         &  & \multicolumn{3}{c}{\pmb{ $n^l=100$}} & \multicolumn{3}{c}{\pmb{ $n^l=500$}} & \multicolumn{3}{c}{\pmb{ $n^l=1000$}} \\ \cmidrule(l){3-5}\cmidrule(l){6-8}\cmidrule(l){9-11}
         &  & \multicolumn{3}{c}{$\epsilon$} & \multicolumn{3}{c}{$\epsilon$} & \multicolumn{3}{c}{$\epsilon$}\\ %\cmidrule(c){3-5}
         %\cmidrule(c){6-8} \cmidrule(c){9-11}  \cmidrule(c){12-14}\\
        \bt Dataset & \bt Method &\bt 1 &\bt 5 & \bt13 & \bt1 & \bt5 & \bt13 & \bt1 & \bt5 & \bt13  \\ \midrule

\multicolumn{1}{c}{Madelon}&SNIP&Reject&Reject&Reject&Reject&Reject&Reject&Reject&Reject&Reject\\
\multicolumn{1}{c}{}&RigL&Reject&Reject&Reject&Reject&Reject&Reject&Reject&Reject&Reject\\
\multicolumn{1}{c}{}&SET&Reject&Reject&Reject&Reject&Reject&Reject&Reject&Reject&Reject\\
\multicolumn{1}{c}{}&CTRE&-*&-*&-*&-*&-*&-*&-*&-*&-*\\
\midrule
\multicolumn{1}{c}{Isolet}&SNIP&Reject&Reject&Reject&Reject&Reject&Reject&Reject&Reject&Reject\\
\multicolumn{1}{c}{}&RigL&Reject&Reject&Reject&Reject&Reject&Reject&Reject&Reject&Reject\\
\multicolumn{1}{c}{}&SET&Reject*&Reject*&Reject*&Reject&Reject*&Reject*&Reject&Reject*&Reject*\\
\multicolumn{1}{c}{}&CTRE&-&-&-&-*&-&-&-*&-&-\\
\midrule
\multicolumn{1}{c}{MNIST}&SNIP&Reject&Reject&Reject&Reject&Reject&Reject&Reject&Reject&Reject\\
\multicolumn{1}{c}{}&RigL&Reject&Reject&Reject&Reject&Reject&Reject&Reject&Reject&Reject\\
\multicolumn{1}{c}{}&SET&\textcolor{red}{True}&Reject&Reject&Reject&Reject&Reject&Reject&Reject&Reject\\
\multicolumn{1}{c}{}&CTRE&-*&-*&-*&-*&-*&-*&-*&-*&-*\\
\midrule
\multicolumn{1}{c}{\shortstack{Fashion-\\MNIST} }&SNIP&Reject&Reject&Reject&Reject&Reject&Reject&Reject&Reject&Reject\\
\multicolumn{1}{c}{}&RigL&Reject&Reject&Reject&Reject&Reject&Reject&Reject&Reject&Reject\\
\multicolumn{1}{c}{}&SET&True*&\textcolor{red}{True}&\textcolor{red}{True}&Reject&Reject&Reject&Reject&\textcolor{red}{True}&Reject\\
\multicolumn{1}{c}{}&CTRE&-*&-*&-*&-*&-*&-*&-*&-*&-*\\
\midrule
\multicolumn{1}{c}{CIFAR10}&SNIP&Reject&Reject&Reject&Reject&Reject&Reject&Reject&Reject&Reject\\
\multicolumn{1}{c}{}&RigL&Reject&Reject&\textcolor{red}{True}&Reject&Reject&Reject&Reject&Reject&Reject\\
\multicolumn{1}{c}{}&SET&True*&Reject&True*&Reject&Reject&Reject&Reject&Reject&Reject\\
\multicolumn{1}{c}{}&CTRE&-&-*&-&-*&-*&-*&-*&-*&-*\\
\midrule
\multicolumn{1}{c}{CIFAR100}&SNIP&Reject&Reject&Reject&Reject&Reject&Reject&Reject&Reject&Reject\\
\multicolumn{1}{c}{}&RigL&Reject&Reject&Reject&Reject&Reject&Reject&Reject&Reject&Reject\\
\multicolumn{1}{c}{}&SET&Reject*&Reject&\textcolor{red}{True}&Reject&Reject&Reject&Reject&Reject&Reject\\
\multicolumn{1}{c}{}&CTRE&-&-*&-*&-*&-*&-*&-*&-*&-*\\
\midrule

    \end{tabular}}
    \end{scriptsize}
    %\vskip -0.2 in
\end{table}

%% file: sections/Discussion.tex
\section{Discussion}
In this section, we perform an in-depth analysis to understand the behavior of CTRE better. First, in Section \ref{ssec:ablation}, we perform two ablation studies to study the effectiveness of both random topology search and similarity importance metric in the performance of CTRE. In Section \ref{ssec:weight_removal}, we discuss why we have chosen magnitude over cosine similarity for the weight removal step. In Section \ref{ssec:discussion_normalization}, we discuss why the insensitivity of cosine similarity to the vector's magnitude is important in the performance of CTRE. Finally, we discuss the convergence of CTRE in Section \ref{ssec:discussion_convergence}.

%%%%%%%%%%%%%%%%%%%%%%%%%%%%%%%%%%%%%%%%%%%%%%%%%%%%%%%%%%%%%
\subsection{Ablation Study: Analysis of Topology Search Policies} \label{ssec:ablation}
%%%%%%%%%%%%%%%%%%%%%%%%%%%%%%%%%%%%%%%%%%%%%%%%%%%%%%%%%%%%%
\looseness=-1
This section presents and discusses the results of two ablation studies designed to understand better the effect of different topology search policies in CTRE. In the following, we describe each ablation experiment separately.

%%%%%%%%%%%%%%%%%%%%%%%%%%%%%%%%%%%%%%%%%%%%%%%%%%%%%%%%%%%%%%%%
%%%%% Ablation 1
\subsubsection{ Ablation Study 1: Random Topology Search}\label{sssec:ablation1}
The first ablation study aims to analyze the effect of random connection addition on the behavior of CTRE. Therefore, instead of using the similarity information and random search (simultaneously in CTRE\textsubscript{sim} and sequentially in CTRE\textsubscript{seq}), we only use the cosine similarity information at each epoch. We call this approach CTRE$_{w/oRandom}$ and repeat the experiments from Section \ref{ssec:performance_evaluation}. The detailed results are available in Table \ref{tab:results_comparison_cosine}.

\input{supplementary/Tables/table_cosine_comparison}

As can be seen in Table \ref{tab:results_comparison_cosine}, in most cases considered, CTRE$_{w/oRandom}$ has been outperformed by CTRE\textsubscript{sim} and CTRE\textsubscript{seq}. On the other hand, we can observe that on image datasets, CTRE$_{w/oRandom}$ has comparable performance to the other two methods; this indicates the effectiveness of similarity information on the image datasets. However, on tabular datasets, it performs poorly on high sparsity cases ($\varepsilon = 1$). Therefore, using only cosine information in these scenarios can cause the topology search to be stuck in a local minimum. This might have been originated by the early stopping of changes in the activation values, which leads to an early stop in topology search. CTRE\textsubscript{seq} solves this by changing the weight update policy to random search. However, there is a risk of early switching to random search when the cosine information has not been fully exploited. Finally, by considering both random and cosine information in each epoch, the CTRE\textsubscript{sim} algorithm will minimize the risk of staying in the local minimum or switching to a completely random search, both of which might slow the training process. In the context of network topology search, these components can also be characterized as exploitation (local information based on the similarity between neurons) and exploration (random search). As a result, CTRE\textsubscript{sim} can mitigate the limitations of CTRE\textsubscript{seq} and find a performant sub-network by leveraging these two components, which outperform state-of-the-art algorithms.

%%%%%%%%%%%%%%%%%%%%%%%%%%%%%%%%%%%%%%%%%%%%%%%%%%%%%%%%%%%%%%%%
%%%%% Ablation 2
\subsubsection{Ablation Study 2: Cosine similarity-based Topology Search}\label{sssec:ablation2} 
To study the effectiveness of cosine similarity addition in the performance of CTRE, we design an experiment; in this experiment, we add connections in the reverse order of importance to the network. We expect that adding weights in this order would result in poor performance. We perform this experiment on CTRE\textsubscript{sim}. Concretely, at each step, we add a number of weights with the lowest similarity among the corresponding neurons; if a weight with a very low similarity has been removed in the last weight removal step, we add a random connection instead. We call this method CTRE\textsubscript{sim/LTH} (LTH refers to low to high importance).

As can be seen in Table \ref{tab:results_comparison_cosine}, CTRE\textsubscript{sim/LTH} has been outperformed by CTRE\textsubscript{sim} and CTRE\textsubscript{seq} in most of the cases considered. This shows that cosine similarity is a useful metric to detect the most important weights in the network. By comparing CTRE\textsubscript{sim/LTH} with SET (Table \ref{tab:results_comparison_rigl_implementation}), it is clear that in most cases CTRE\textsubscript{sim/LTH} has a close or slightly worse accuracy than SET. Therefore, it can be inferred that CTRE\textsubscript{sim/LTH} is selecting non-informative weights, which can be similar to or worse than a random search. As a result, this can indicate the effectiveness of the introduced similarity metric (Equation \ref{eq:cosine_similarity}) in finding a well-performing sparse neural network. It is worth noting that on the Isolet dataset, CTRE\textsubscript{sim/LTH} outperforms CTRE\textsubscript{sim} and CTRE\textsubscript{seq} in some cases, particularly in the networks with higher density. This is similar to the results of SET as well. Therefore, we can conclude that random search outperforms other methods on the Isolet dataset and low sparsity levels. However, it is not easy to find a highly sparse network using the random search policy.

%%%%%%%%%%%%%%%%%%%%%%%%%%%%%%%%%%%%%%%%%%%%%%%%%%%%%%%%%%%%%%
\subsection{Analysis of Weight Removal Policy}\label{ssec:weight_removal}
%%%%%%%%%%%%%%%%%%%%%%%%%%%%%%%%%%%%%%%%%%%%%%%%%%%%%%%%%%%%%
\looseness=-1
In this section, we aim to analyze the weight removal policy and further explain the reason behind choosing magnitude-based pruning over the cosine similarity (discussed in Section \ref{ssec:cosine_in_nn}). In many previous studies, magnitude-based pruning has been commonly used as a criterion to remove unimportant weight from a neural network. We design an experiment to compare the performance of magnitude-based and cosine similarity-based pruning in neural networks.    

\begin{figure}[!t]
        \vskip 0.1in
        \begin{center}
           \centerline{\includegraphics[width=\textwidth]{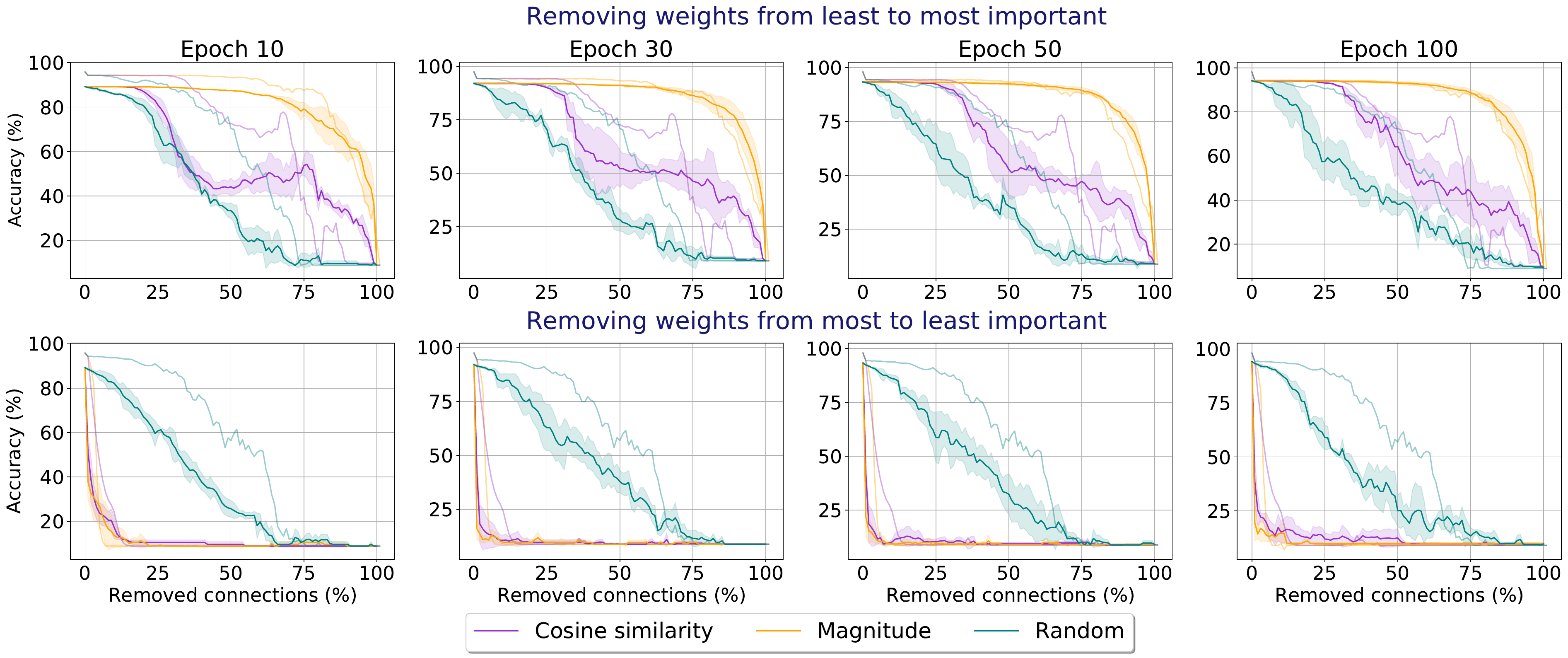}}
        \caption{Effect of weight removal using three criteria including magnitude, cosine similarity, and random, on the classification accuracy (\%) at different epochs. The lines with higher transparency corresponds to the weight removal of the SET-MLP and the lines with lower transparency corresponds to the dense-MLP}
        \label{fig:weight_removal}
        \end{center}
        \vskip -0.35in
    \end{figure}
    
\looseness=-1    
In this experiment, we start with a trained network and gradually remove weights based on the magnitude and cosine similarity value (Using Equation \ref{eq:cosine_similarity}) of the corresponding connection. We also consider random pruning as the baseline. 

\textbf{Settings.} We perform this experiment using two networks: $(1)$ A 3-layer dense MLP with $1000$ neurons in each layer, and $(2)$ A 3-layer sparse MLP with $1000$ neurons in each layer that is trained using the SET approach \cite{mocanu2018scalable} (3.2\% density). The choice of SET instead of CTRE was made to avoid any biases on the cosine similarity weight removal, as CTRE uses cosine information to add weights. Both of these networks are trained on the MNIST dataset. 

\textbf{Weight Removal.} We remove weights with two orders on each of the sparse and dense networks: least to most important and vice versa. We gradually remove weights; at each step, we remove 1\% of the connections and measure the accuracy of the pruned network until no connection remains in the network. 

\textbf{Results.} The results when the two networks are trained for $10$, $30$, $50$, and $100$ epochs are available in Figure \ref{fig:weight_removal}. In this figure, the lines with higher transparency correspond to the weight removal of the SET-MLP, and the lines with lower transparency correspond to the dense-MLP. This experiment has been repeated with three seeds for each case.

As shown in Figure \ref{fig:weight_removal}, when weights are removed from least to most important, magnitude-based pruning can order weights better than cosine similarity-based pruning. When the networks are trained for $100$ epochs, by dropping the unimportant weights using magnitude, the major accuracy drop starts almost after removing 70\% of the connections, while it happens after removing 30\%  for cosine similarity. This behavior exists in both the dense and the sparse networks. As expected, the drop for random removal happens from the beginning of the pruning procedure. In earlier epochs ($10$, $30$, and $50$), the drop in the accuracy happens earlier for both magnitude and cosine similarity. 

It can be seen in Figure \ref{fig:weight_removal}, by removing weights in the opposite order (from most to least important), the behavior of drop in the accuracy is almost similar for cosine similarity-based and magnitude-based pruning in SET-MLP, particularly in the earlier epochs. Therefore, both magnitude and cosine similarity can identify the most important connections in good order. However, this behavior is different in the dense network; magnitude-based pruning can better detect the most important weights. In the dense network, the drop in the accuracy for magnitude-based pruning happens earlier than cosine similarity pruning. 

\textbf{Conclusions.} These observations can lead us to conclude that, firstly, the magnitude can be a good metric for weight removal in sparse training. Secondly, it can be inferred that cosine similarity can be a good metric for adding the most important connections in the weight addition phase in sparse neural networks in the absence of magnitude. As discussed earlier, the cosine similarity information of each connection is an informative criterion to detect the most important weights in a sparse neural network and has similar behavior to magnitude-based pruning in these scenarios. Therefore, in the absence of magnitude for non-existing connections in a sparse neural network (during weight addition), cosine similarity can be a useful criterion to detect the most important weights without requiring computing dense gradient information.

%%%%%%%%%%%%%%%%%%%%%%%%%%%%%%%%%%%%%%%%%%%%%%%%%%%%%%%%%%%%%%
\subsection{Magnitude Insensitivity: The Favorable Feature of Cosine Similarity in Noisy Environments }\label{ssec:discussion_normalization}
%%%%%%%%%%%%%%%%%%%%%%%%%%%%%%%%%%%%%%%%%%%%%%%%%%%%%%%%%%%%%%
This section further discusses why cosine similarity has been chosen as a metric to determine the importance of non-existing connections. Specifically, we mainly focus on analyzing the importance of normalization in Equation \ref{eq:cosine_similarity} in the performance of the algorithm. While based on the Hebbian learning rule, the connection among a pair of neurons with high activations should be strengthened, we argue that in the search for a performant sparse neural network, the magnitude of the activations should be ignored.

Based on Hebb's rule (Section \ref{ssec:background_hebb}), the connection among the neurons with high activations receives higher synaptic updates. Therefore, if we evolve the topology using this rule (without any normalization) the importance of a non-existing connection should be determined by: $\abs{{ \mA_{:, p}^{l-1} \cdot  \mA_{:, q}^{l}}}$. We evaluate the performance of this metric by replacing it with Equation \ref{eq:cosine_similarity} in CTRE\textsubscript{sim} and CTRE\textsubscript{seq}; we name these algorithms CTRE\textsubscript{sim-Hebb} and CTRE\textsubscript{seq-Hebb}, respectively. 

\begin{figure}[!b]
        %\vskip -0.45in
        \begin{center}
        \centerline{\includegraphics[width=0.8\textwidth]{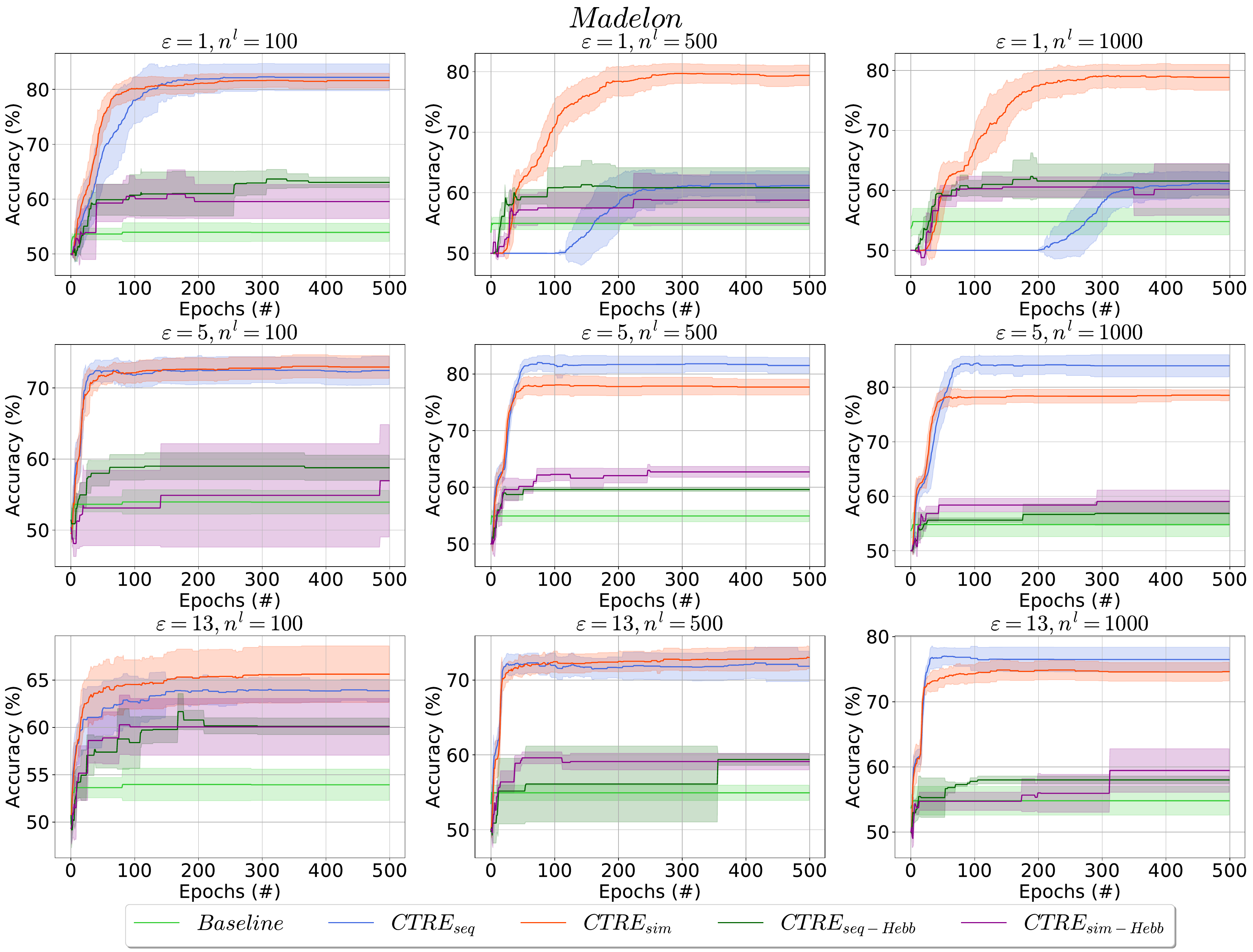}}
        \caption{Classification accuracy (\%) comparison on Madelon for CTRE and pure Hebbian-based updates. }
        \label{fig:cosinevshebb}
        \end{center}
        %\vskip -0.45in
    \end{figure}

We evaluate these methods on the Madelon dataset. The reason behind choosing this dataset is due to its interesting properties; it contains $480$ noisy features (out of the $500$ features). Therefore, finding informative information paths through the network is considered to be a challenging task. The settings of this experiment are similar to Section \ref{ssec:performance_evaluation}; we measure the performance on networks with different sizes and sparsity levels. The results are presented in Table \ref{tab:results_hebb} and the accuracy during training is plotted in Figure \ref{fig:cosinevshebb}. CTRE\textsubscript{sim-Hebb} and CTRE\textsubscript{seq-Hebb} have been outperformed by CTRE\textsubscript{sim} and CTRE\textsubscript{seq} in all cases considered. Particularly, we can observe that as the network becomes sparser, the gap between the performance of the pure Hebbian-based methods and the cosine similarity-based methods increases. 

\input{supplementary/Tables/table_madelon_hebb}

The poor performance of CTRE\textsubscript{sim-Hebb} and CTRE\textsubscript{seq-Hebb} on the Madelon dataset is resulted from their sensitivity to the magnitude of activation values. As Madelon contains many noisy features, some uninformative neurons likely receive a high activation value. Therefore, if we use only the activation magnitude to find the informative paths of information, the algorithm will be biased on the neurons with very high activation, which might not be informative. Therefore, it is likely to assign new connections to noisy features with high activation. This would cause the algorithm to be stuck in a local minimum which might be difficult to escape as these neurons continue to receive more and more connections at each epoch. Furthermore, as the networks become sparser, the informative features have a lower chance of receiving more connections (there are more noisy features compared to the informative ones). Therefore, in sparse networks, the gap between the performance of these methods is much larger than in denser networks. Based on these observations, it can be concluded that the insensitivity of cosine similarity to the vector's magnitude helps CTRE to be more robust in noisy environments.

%%%%%%%%%%%%%%%%%%%%%%%%%%%%%%%%%%%%%%%%%%%%%%%%%%%%%%%%%%%%%%
\subsection{Convergence Analysis}\label{ssec:discussion_convergence}
%%%%%%%%%%%%%%%%%%%%%%%%%%%%%%%%%%%%%%%%%%%%%%%%%%%%%%%%%%%%%%
This section discusses the convergence of the proposed algorithm for training sparse neural networks from scratch, CTRE. In short, we first discuss the effect of the weight evolution process on the algorithm's convergence. Secondly, we explore whether cosine similarity causes CTRE to converge into a local minimum or not.

First, we analyze if the weight evolution process in the CTRE algorithm interferes with the convergence of the back-propagation algorithm or not. In the CTRE algorithm, a number of connections of the weights are removed at each training epoch, and the same number of connections are added based on the cosine or random search policies. The weight evolution process is performed at each epoch after the standard feed-forward and back-propagation steps. The removed connections have a small magnitude compared to the other connections, and newly activated connections also get a small value. Therefore, they do not change the loss value significantly. The new weights will be updated in the next feed-forward and back-propagation step, and they will grow or shrink. Therefore, the weight evolution process does not disrupt the convergence of the model.

\begin{figure}[!t]
    \begin{center}
    \centerline{\includegraphics[width=0.8\textwidth]{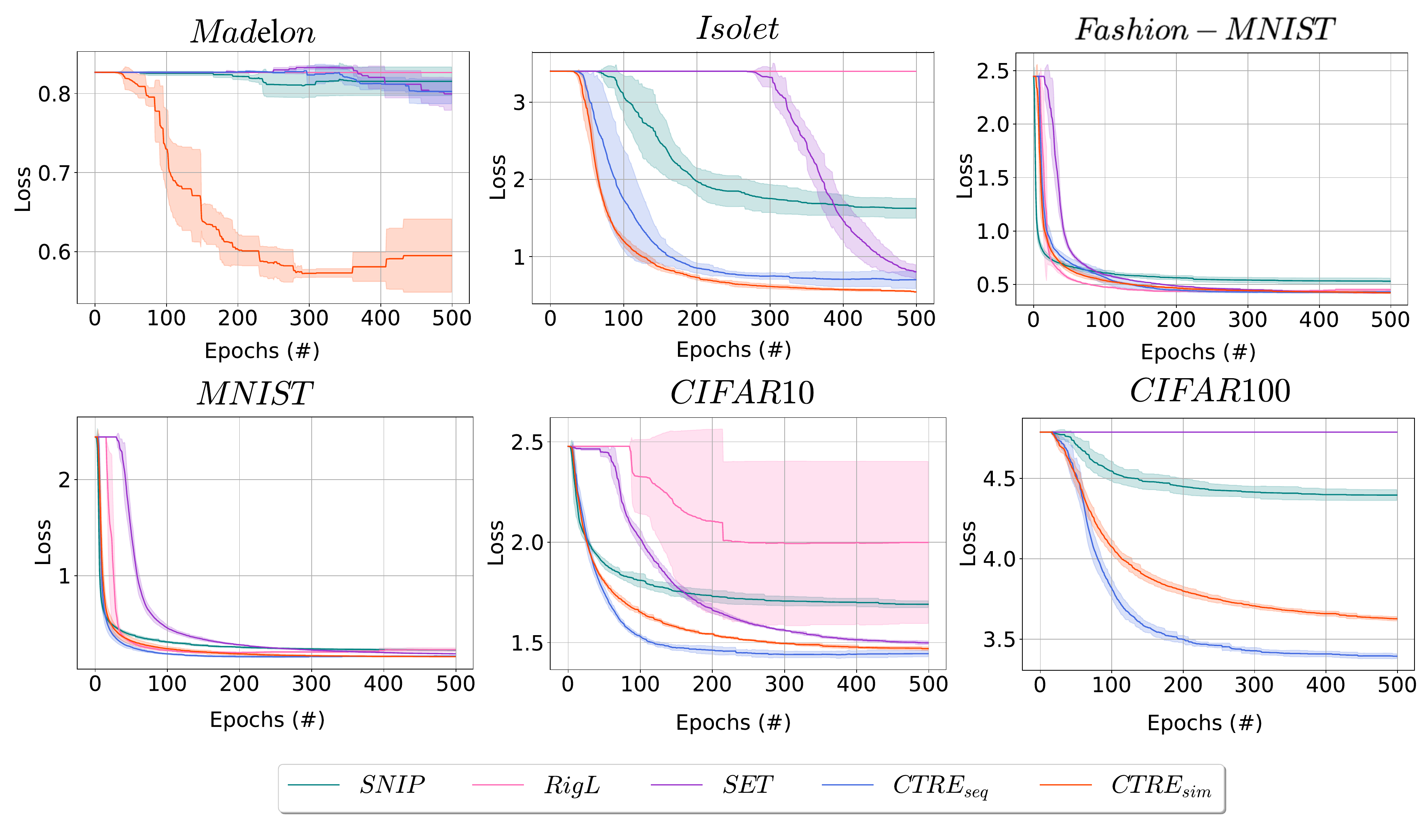}}
    \caption{Test loss comparison during training for the high sparsity regime and a large network ($\varepsilon=1$, $n^l=1000$).}
    \label{fig:test_loss_all}
    \end{center}
\end{figure}

To validate this, we depict the test loss during training in Figure \ref{fig:test_loss_all} for the high sparsity regime and a large network ($\varepsilon=1$, $n^l=1000$). It can be observed that the loss function converges for the CTRE algorithm on all the datasets. In addition, in most cases, its convergence speed is much faster than for the other algorithms.

\begin{figure}[!t]
    \begin{center}
    \centerline{\includegraphics[width=0.8\textwidth]{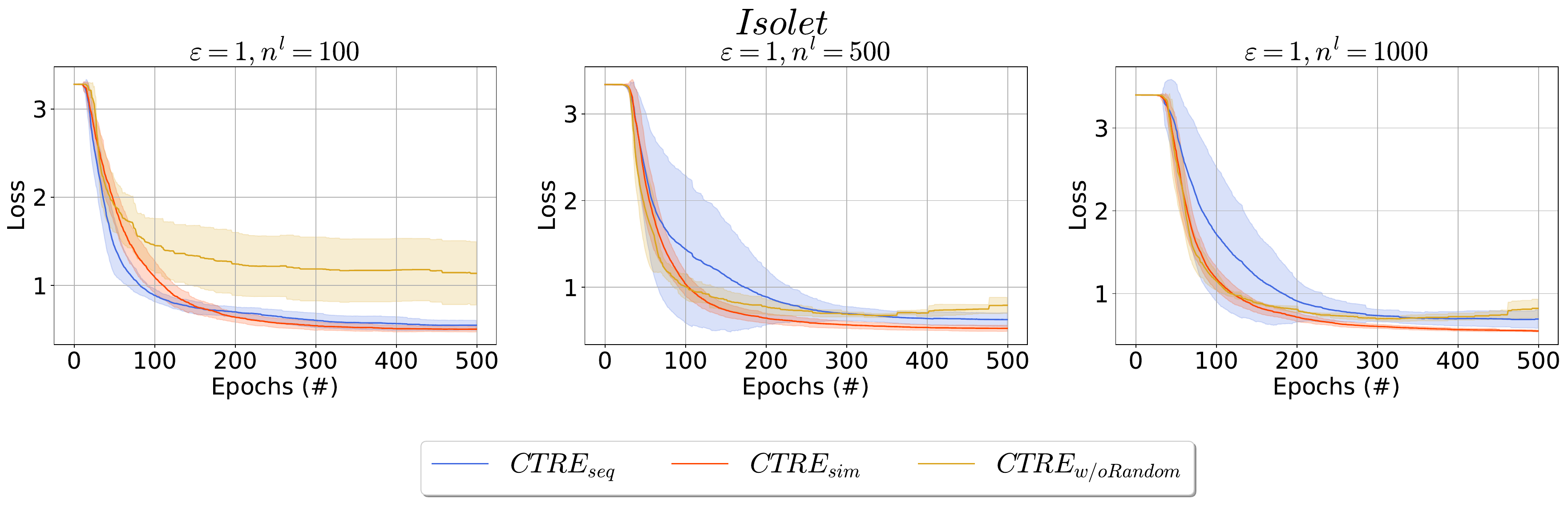}}
    \caption{Test loss comparison during training for the high sparsity regime ($\varepsilon=1$) on the Isolet dataset.}
    \label{fig:loss_isolet}
    \end{center}
\end{figure}

Secondly, we analyze whether CTRE is prone to converge to a local optima. As discussed in Section \ref{ssec:weight_removal}, cosine similarity is very successful at determining the most and least important connections in the network. However, in the mid-importance range, it might not be able to rank connections as well as the magnitude criterion; therefore, it might add some connections that do not contribute to decreasing the network loss. In such cases, the cosine similarity metric might prevent topology exploration and get stuck in local minima. CTRE explores other weights and exits this local minimum by using a random search. To validate this, in Figure \ref{fig:loss_isolet}, we have presented the loss during training for CTRE\textsubscript{seq}, CTRE\textsubscript{sim}, and CTRE\textsubscript{w/oRandom} on three highly sparse neural networks trained on the Isolet dataset. The fast decrease in the loss in these plots indicates that all three methods quickly find a good-performing sub-network. However, the loss value of CTRE\textsubscript{w/oRandom} does not improve significantly after 200 epochs, and it converges to a higher value than the other two methods. Therefore, it is important to use random exploration to keep improving the topology and avoid local minima as it is done in CTRE.

%% file: supplementary/Tables/table_cosine_comparison.tex
%\newrobustcmd{\bt}{\bfseries} % bold command
\begin{table}[!t]
    
    \caption{Classification accuracy (\%) comparison among Cosine similarity-based methods.} \label{tab:results_comparison_cosine}

    \begin{scriptsize}
    %\scalebox{0.9}{ 
    \scalebox{0.88}{ 
    \begin{tabular}{@{}c@{\hskip 0.02in}c@{\hskip 0.04in}c@{\hskip 0.04in}c@{\hskip 0.04in}c@{\hskip 0.07in}c@{\hskip 0.04in}c@{\hskip 0.04in}c@{\hskip 0.07in}c@{\hskip 0.04in}c@{\hskip 0.04in}c@{}}
        \toprule
         &  & \multicolumn{3}{c}{\pmb{ $n^l=100$}} & \multicolumn{3}{c}{\pmb{ $n^l=500$}} & \multicolumn{3}{c}{\pmb{ $n^l=1000$}} \\ \cmidrule(l){3-5}\cmidrule(l){6-8}\cmidrule(l){9-11}
         &  & \multicolumn{3}{c}{$\epsilon$} & \multicolumn{3}{c}{$\epsilon$} & \multicolumn{3}{c}{$\epsilon$}\\ %\cmidrule(c){3-5}
         %\cmidrule(c){6-8} \cmidrule(c){9-11}  \cmidrule(c){12-14}\\
        \bt Dataset & \bt Method &\bt 1 &\bt 5 & \bt13 & \bt1 & \bt5 & \bt13 & \bt1 & \bt5 & \bt13  \\ \midrule

\multicolumn{1}{c}{Madelon}&CTRE\textsubscript{seq}&\pmb{$82.2\pm2.4$}&$72.5\pm2.0$&$63.9\pm1.2$&$61.2\pm2.4$&\pmb{$81.5\pm1.4$}&$71.8\pm2.0$&$61.1\pm1.9$&\pmb{$83.9\pm2.0$}&\pmb{$76.5\pm1.9$}\\
\multicolumn{1}{c}{}&CTRE\textsubscript{sim}&$81.6\pm1.3$&\pmb{$73.0\pm1.6$}&\pmb{$65.6\pm3.0$}&\pmb{$79.4\pm1.7$}&$77.7\pm1.4$&\pmb{$73.0\pm1.5$}&\pmb{$78.8\pm2.2$}&$78.5\pm1.0$&$74.6\pm1.4$\\
\multicolumn{1}{c}{}&CTRE\textsubscript{w/oRandom}&$79.6\pm2.0$&$59.0\pm2.3$&$57.6\pm1.2$&$58.6\pm12.0$&$72.7\pm5.3$&$61.4\pm1.6$&$50.0\pm0.0$&$71.2\pm1.4$&$56.3\pm3.7$\\
\multicolumn{1}{c}{}&CTRE\textsubscript{sim/LTH}&$61.3\pm3.4$&$58.0\pm2.6$&$56.9\pm1.7$&$53.2\pm0.7$&$59.6\pm1.4$&$58.2\pm1.7$&$51.1\pm0.2$&$58.9\pm1.6$&$58.3\pm1.2$\\
\midrule

\multicolumn{1}{c}{Isolet}&CTRE\textsubscript{seq}&$86.7\pm1.8$&$92.4\pm1.1$&$94.3\pm0.5$&$87.2\pm2.0$&\pmb{$92.3\pm0.6$}&$94.0\pm0.4$&$86.7\pm1.6$&$91.5\pm1.0$&$93.7\pm0.5$\\
\multicolumn{1}{c}{}&CTRE\textsubscript{sim}&\pmb{$87.5\pm0.8$}&\pmb{$93.4\pm0.7$}&$94.3\pm0.8$&\pmb{$87.8\pm1.1$}&$91.7\pm1.1$&$93.9\pm0.5$&\pmb{$88.3\pm0.7$}&$91.3\pm1.3$&$93.1\pm0.6$\\
\multicolumn{1}{c}{}&CTRE\textsubscript{w/oRandom}&$66.7\pm11.3$&$91.9\pm0.3$&$93.5\pm0.2$&$84.5\pm0.8$&$89.9\pm0.3$&$92.8\pm0.5$&$83.9\pm1.9$&$87.6\pm1.1$&$92.1\pm0.9$\\
\multicolumn{1}{c}{}&CTRE\textsubscript{sim/LTH}&$81.1\pm1.3$&$93.3\pm0.6$&\pmb{$95.0\pm0.2$}&$66.3\pm4.9$&$91.7\pm0.6$&\pmb{$94.6\pm0.4$}&$61.2\pm11.8$&\pmb{$91.8\pm1.1$}&\pmb{$93.8\pm0.2$}\\
\midrule

\multicolumn{1}{c}{MNIST}&CTRE\textsubscript{seq}&\pmb{$95.7\pm0.2$}&\pmb{$97.3\pm0.2$}&$97.7\pm0.1$&\pmb{$97.0\pm0.2$}&$97.6\pm0.2$&$97.8\pm0.1$&\pmb{$97.3\pm0.1$}&\pmb{$97.7\pm0.1$}&$97.8\pm0.1$\\
\multicolumn{1}{c}{}&CTRE\textsubscript{sim}&$95.5\pm0.2$&\pmb{$97.3\pm0.1$}&\pmb{$97.8\pm0.1$}&$96.4\pm0.2$&\pmb{$97.7\pm0.1$}&\pmb{$98.0\pm0.1$}&$96.6\pm0.2$&\pmb{$97.7\pm0.1$}&\pmb{$97.9\pm0.1$}\\
\multicolumn{1}{c}{}&CTRE\textsubscript{w/oRandom}&$94.8\pm0.3$&$97.0\pm0.2$&$97.5\pm0.1$&$96.9\pm0.0$&$97.4\pm0.1$&$97.4\pm0.1$&$97.2\pm0.3$&$97.5\pm0.2$&$97.5\pm0.1$\\
\multicolumn{1}{c}{}&CTRE\textsubscript{sim/LTH}&$94.6\pm0.2$&$96.9\pm0.1$&$97.4\pm0.1$&$94.2\pm0.1$&$97.3\pm0.2$&$97.4\pm0.0$&$93.9\pm0.0$&$97.1\pm0.0$&$97.5\pm0.1$\\
\midrule

\multirow{2}{*}{\shortstack{Fashion-\\MNIST}}&CTRE\textsubscript{seq}&\pmb{$85.8\pm0.5$}&\pmb{$87.5\pm0.3$}&$87.4\pm0.3$&\pmb{$87.1\pm0.4$}&$87.9\pm0.3$&$88.0\pm0.2$&\pmb{$87.3\pm0.2$}&$88.0\pm0.3$&\pmb{$88.3\pm0.2$}\\
\multicolumn{1}{c}{}&CTRE\textsubscript{sim}&\pmb{$85.8\pm0.3$}&\pmb{$87.5\pm0.3$}&\pmb{$88.0\pm0.2$}&$86.4\pm0.5$&\pmb{$88.1\pm0.2$}&\pmb{$88.3\pm0.2$}&$86.5\pm0.3$&\pmb{$88.1\pm0.3$}&\pmb{$88.3\pm0.2$}\\
\multicolumn{1}{c}{}&CTRE\textsubscript{w/oRandom}&$83.9\pm0.4$&$86.9\pm0.3$&$86.9\pm0.1$&$86.2\pm0.3$&$87.9\pm0.2$&$88.0\pm0.2$&$86.8\pm0.3$&$87.7\pm0.4$&$88.2\pm0.2$\\
\multicolumn{1}{c}{}&CTRE\textsubscript{sim/LTH}&$85.0\pm0.2$&$87.3\pm0.3$&$87.6\pm0.2$&$84.4\pm0.3$&$87.5\pm0.3$&$87.8\pm0.3$&$84.2\pm0.2$&$87.6\pm0.2$&$88.0\pm0.2$\\
\midrule

\multicolumn{1}{c}{CIFAR10}&CTRE\textsubscript{seq}&$47.9\pm0.8$&$50.1\pm0.5$&$50.1\pm0.5$&\pmb{$51.3\pm0.5$}&\pmb{$52.7\pm0.5$}&$52.6\pm0.6$&$52.0\pm0.7$&$53.2\pm0.6$&\pmb{$53.6\pm0.6$}\\
\multicolumn{1}{c}{}&CTRE\textsubscript{sim}&\pmb{$48.2\pm0.4$}&\pmb{$50.3\pm0.3$}&\pmb{$50.5\pm0.4$}&$50.6\pm0.4$&$52.6\pm0.7$&$52.5\pm0.7$&$50.0\pm0.4$&$52.7\pm0.5$&$53.5\pm0.5$\\
\multicolumn{1}{c}{}&CTRE\textsubscript{w/oRandom}&$45.4\pm0.2$&$49.4\pm0.3$&$49.9\pm0.4$&$50.9\pm0.3$&$52.2\pm0.6$&\pmb{$52.7\pm0.4$}&\pmb{$52.3\pm0.2$}&\pmb{$53.3\pm0.5$}&$53.5\pm0.4$\\
\multicolumn{1}{c}{}&CTRE\textsubscript{sim/LTH}&$46.7\pm0.5$&$49.6\pm0.5$&$50.4\pm0.3$&$45.4\pm0.5$&$50.9\pm0.2$&$51.2\pm0.4$&$44.9\pm0.5$&$50.6\pm0.1$&$50.9\pm0.3$\\
\midrule

    \end{tabular} }
    \end{scriptsize}
\end{table}

%% file: supplementary/Tables/table_madelon_hebb.tex
%\newrobustcmd{\bt}{\bfseries} % bold command
\begin{table}[!t]
    \centering
    \caption{Classification accuracy (\%) comparison of Cosine similarity-based methods and pure Hebbian-based evolution, on the Madelon dataset. Please note that N (total number of parameters of the network) is scaled by $\times10^3$.} \label{tab:results_hebb}

    \begin{scriptsize}
    \scalebox{0.88}{ 
    \begin{tabular}{@{}c@{\hskip 0.02in}c@{\hskip 0.04in}c@{\hskip 0.04in}c@{\hskip 0.04in}c@{\hskip 0.07in}c@{\hskip 0.04in}c@{\hskip 0.04in}c@{\hskip 0.07in}c@{\hskip 0.04in}c@{\hskip 0.04in}c@{}}
        \toprule
         &  & \multicolumn{3}{c}{\pmb{ $n^l=100$}} & \multicolumn{3}{c}{\pmb{ $n^l=500$}} & \multicolumn{3}{c}{\pmb{ $n^l=1000$}} \\ \cmidrule(l){3-5}\cmidrule(l){6-8}\cmidrule(l){9-11}
         &  & \multicolumn{3}{c}{$\epsilon$} & \multicolumn{3}{c}{$\epsilon$} & \multicolumn{3}{c}{$\epsilon$}\\ %\cmidrule(c){3-5}
         %\cmidrule(c){6-8} \cmidrule(c){9-11}  \cmidrule(c){12-14}\\
        \bt Dataset & \bt Method &\bt 1 &\bt 5 & \bt13 & \bt1 & \bt5 & \bt13 & \bt1 & \bt5 & \bt13  \\ \midrule
\multicolumn{1}{c}{}&$D(\%)$  ($N (\times10^3)$)&$1.6$ ($1.1$)&$7.8$ ($5.5$)&$20.4$ ($14.3$)&$0.5$ ($3.5$)&$2.3$ ($17.5$)&$6.1$ ($45.5$)&$0.3$ ($6.5$)&$1.3$ ($32.5$)&$3.4$ ($84.5$)
\\ \cmidrule(l){3-11} 

\multicolumn{1}{c}{Madelon}&CTRE\textsubscript{seq}&\pmb{$82.2\pm2.4$}&$72.5\pm2.0$&$63.9\pm1.2$&$61.2\pm2.4$&\pmb{$81.5\pm1.4$}&$71.8\pm2.0$&$61.1\pm1.9$&\pmb{$83.9\pm2.0$}&\pmb{$76.5\pm1.9$}\\
\multicolumn{1}{c}{}&CTRE\textsubscript{sim}&$81.6\pm1.3$&\pmb{$73.0\pm1.6$}&\pmb{$65.6\pm3.0$}&\pmb{$79.4\pm1.7$}&$77.7\pm1.4$&\pmb{$73.0\pm1.5$}&\pmb{$78.8\pm2.2$}&$78.5\pm1.0$&$74.6\pm1.4$\\
\multicolumn{1}{c}{}&CTRE\textsubscript{seq-Hebb}&$63.1\pm1.0$&$58.8\pm1.8$&$60.1\pm0.9$&$60.8\pm3.3$&$59.6\pm0.3$&$59.4\pm0.8$&$61.6\pm2.9$&$56.8\pm2.1$&$58.0\pm0.6$\\
\multicolumn{1}{c}{}&CTRE\textsubscript{sim-Hebb}&$59.6\pm3.1$&$56.9\pm7.9$&$60.1\pm3.0$&$58.8\pm4.2$&$62.7\pm1.0$&$59.1\pm1.1$&$60.2\pm4.3$&$59.1\pm2.1$&$59.4\pm3.3$\\
\midrule

    \end{tabular}}
    \end{scriptsize}
\end{table}

%% file: sections/appendix.tex
\newpage
\appendix
\section*{Appendix}

%%%%%%%%%%%%%%%%%%%%%%%%%%%%%%%%%%%%%%%%%%%%%
\section{Performance Evaluation}\label{appendix:learning_curves}
\begin{figure}[H]
        \vskip 0.1in
        \begin{center}
        \centerline{\includegraphics[width=0.9\textwidth]{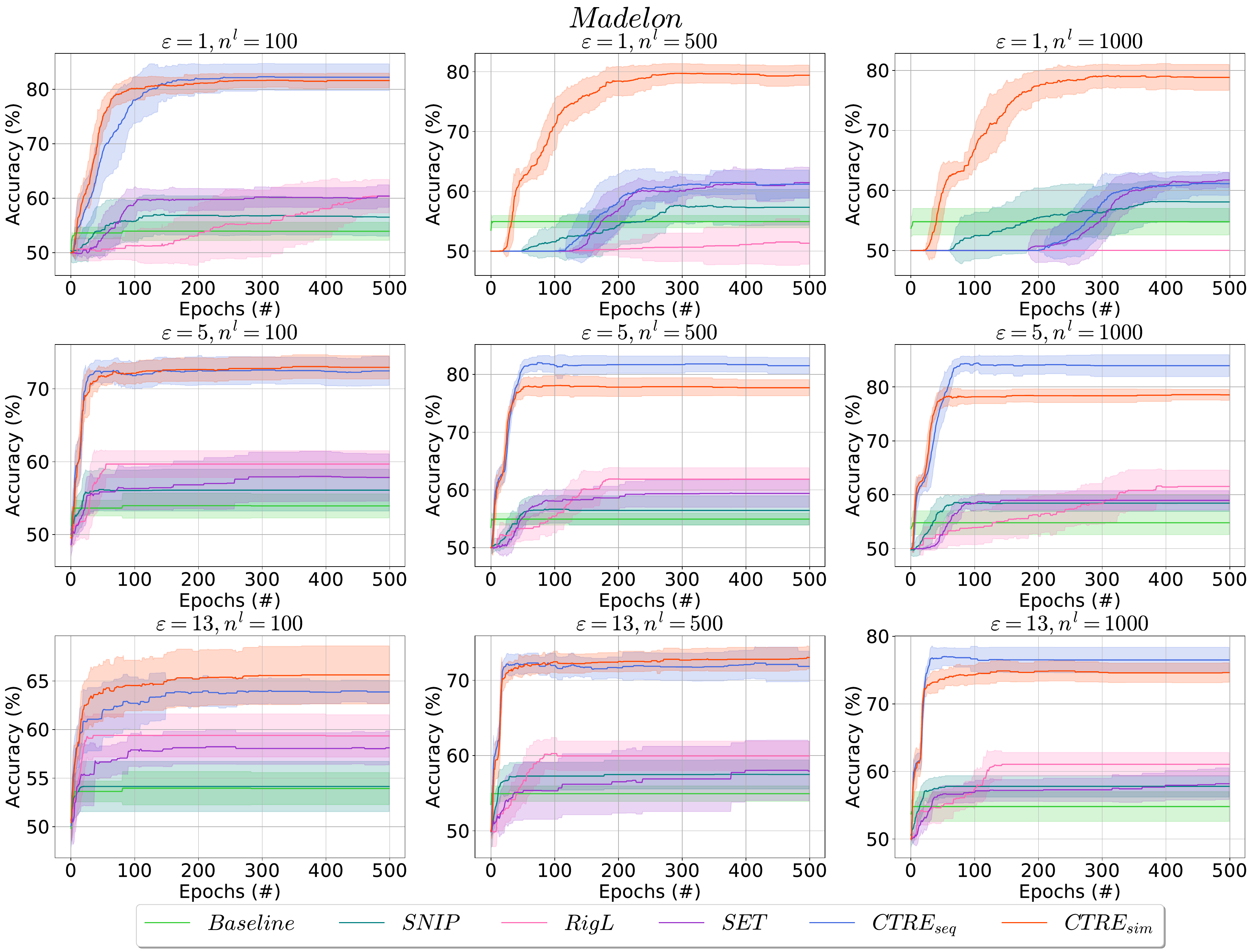}}
        \caption{Classification accuracy (\%) results on Madelon.}
        \label{fig:acc_madelon}
        \end{center}
        \vskip -0.35in
    \end{figure}

\begin{figure}[H]
        \vskip 0.1in
        \begin{center}
        \centerline{\includegraphics[width=0.9\textwidth]{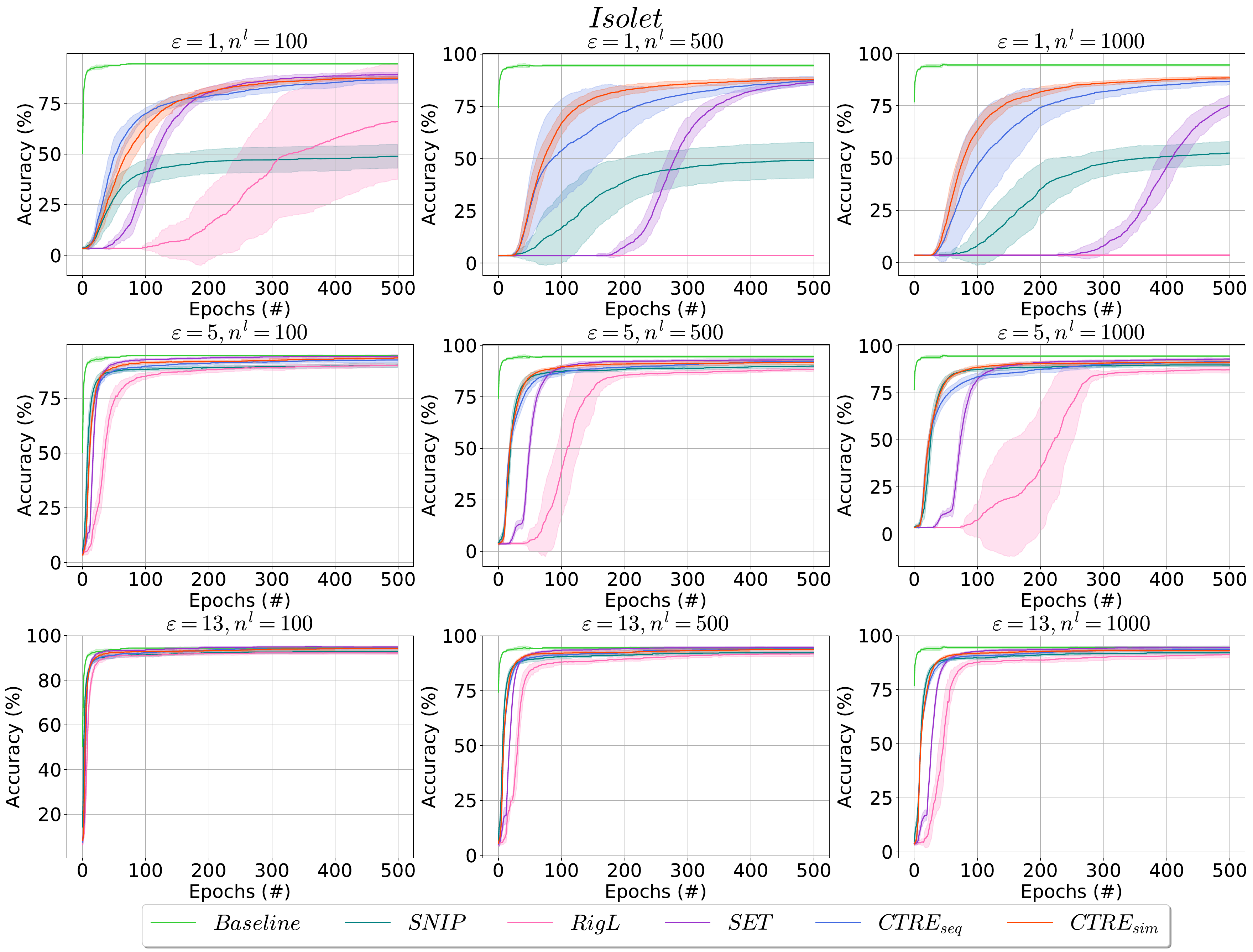}}
        \caption{Classification accuracy (\%) results on Isolet.}
        \label{fig:acc_isolet}
        \end{center}
        \vskip -0.35in
    \end{figure}

\begin{figure}[H]
        \vskip 0.1in
        \begin{center}
        \centerline{\includegraphics[width=0.9\textwidth]{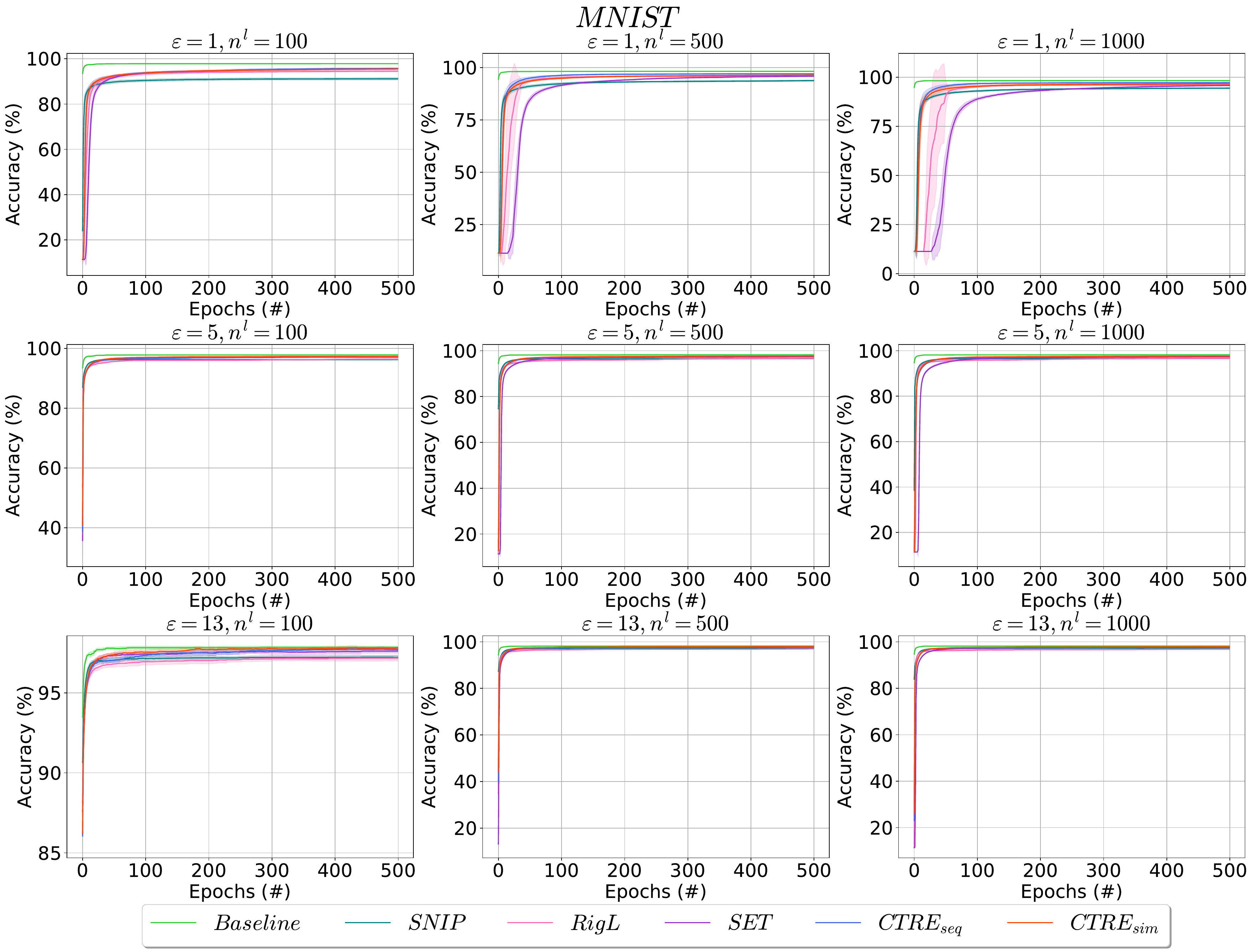}}
        \caption{Classification accuracy (\%) results on MNIST.}
        \label{fig:acc_mnist}
        \end{center}
        \vskip -0.35in
    \end{figure}

\begin{figure}[H]
        \vskip 0.1in
        \begin{center}
        \centerline{\includegraphics[width=0.9\textwidth]{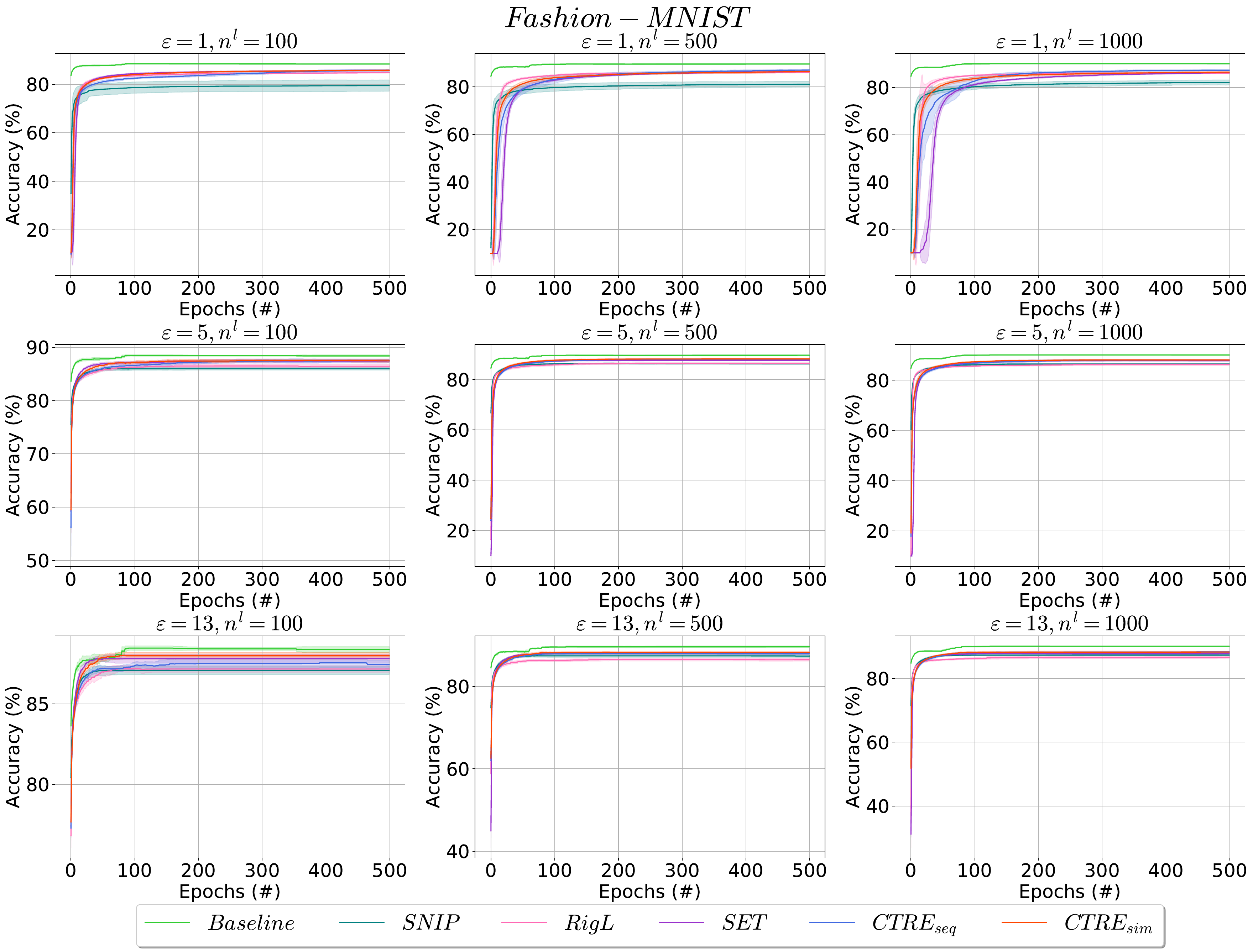}}
        \caption{Classification accuracy (\%) results on Fashion-MNIST.}
        \label{fig:acc_fashion_mnist}
        \end{center}
        \vskip -0.35in
    \end{figure}

\begin{figure}[H]
        \vskip 0.1in
        \begin{center}
        \centerline{\includegraphics[width=0.9\textwidth]{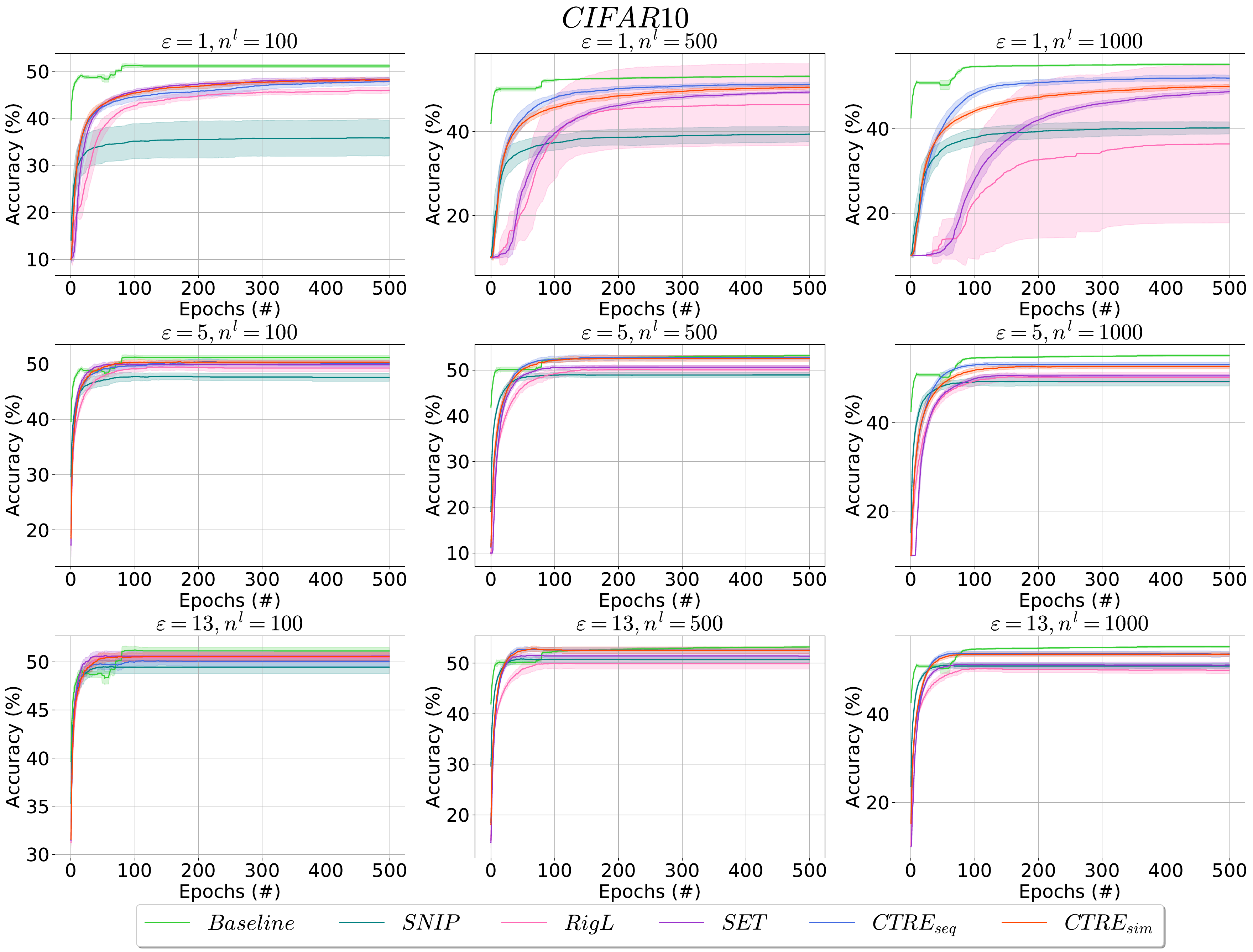}}
        \caption{Classification accuracy (\%) results on CIFAR10.}
        \label{fig:acc_cifar10}
        \end{center}
        \vskip -0.35in
    \end{figure}

\begin{figure}[H]
        \vskip 0.1in
        \begin{center}
        \centerline{\includegraphics[width=0.9\textwidth]{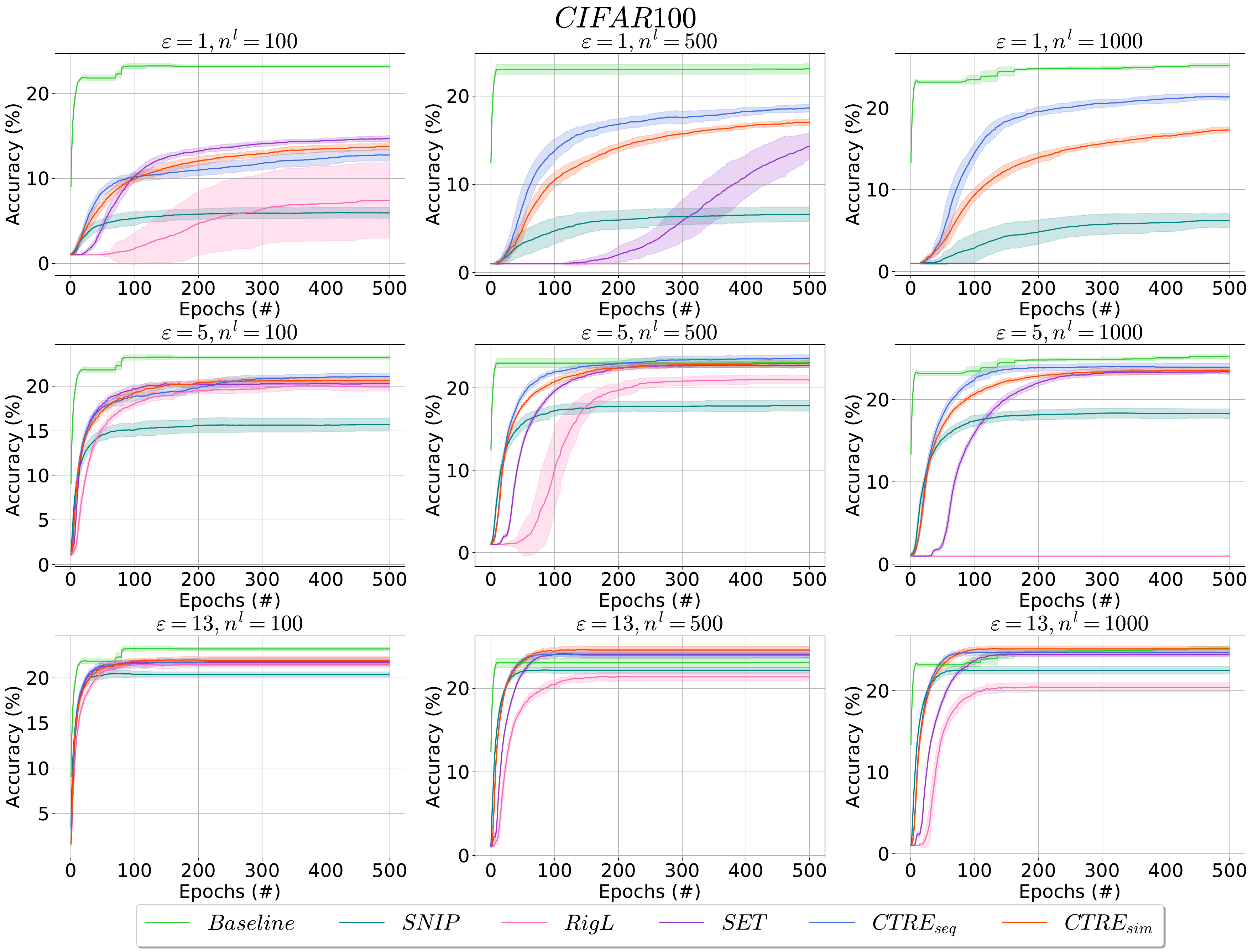}}
        \caption{Classification accuracy (\%) results on CIFAR100.}
        \label{fig:acc_cifar100}
        \end{center}
        \vskip -0.35in
    \end{figure}

%%%%%%%%%%%%%%%%%%%%%%%%%%%%%%%%%%%%%%%%%%%%%
In this appendix, we compare the performance of the algorithms in terms of the accuracy, the learning speed, and computational complexity. Particularly, we analyze the results obtained in Section \ref{ssec:performance_evaluation} in the manuscript. We first introduce a new metric for comparing the learning speed of the methods. We also include the learning curves for the experiments of Section \ref{ssec:performance_evaluation} in Figures \ref{fig:acc_madelon}, \ref{fig:acc_isolet}, \ref{fig:acc_mnist}, \ref{fig:acc_fashion_mnist}, \ref{fig:acc_cifar10}, and \ref{fig:acc_cifar100}, which corresponds to Madelon, Isolet, MNIST, Fashion-MNIST, CIFAR10, and CIFAR100, respectively. The characteristics of these datasets are presented in Table \ref{tab:datasets}.

%%%%%%%%%%%%%%%%%%%%%%%%%%%%%%%%%%%%%%%%%%%%%%%%%%%%%%%%%%%%%
%%%%%%%%%%%%%%%%%%%%%%%%%%%%%%%%%%%%%%%%%%%%%%%%%%%%%%%%%%%%%%
\looseness=-1
To compare the training speed, we define a metric that computes the fraction of the total training time required to reach a certain level of accuracy. We call this metric Training Delay (\emph{TD}) and compute it as follows:
\begin{equation}
    TD = \frac{\min_{acc_{i} 	\geqslant {th \times acc_{max} }, \;i \in\{1, 2, ..., \#\;epochs\}}{i}}{\#\;epochs},
\end{equation}

where $acc_{i}$ is the test accuracy at epoch $i$, \emph{th} is the threshold hyperparameter between $0$ and $1$, $acc_{max}$ is the maximum accuracy achieved by the training methods for the model with $n^l$ hidden neurons and sparsity level $\varepsilon$, and $\#\;epochs$ is the total number of training epochs. In other words, \emph{TD} shows the trade-off between accuracy and learning speed. The lower \emph{TD} is for a method, the faster it can be trained to reach a certain desired level of accuracy (determined by \emph{th}); therefore, it has a better trade-off between accuracy and learning speed. We believe that minimizing this metric is crucial for low-resource devices where accuracy is not the only important aspect for evaluating the performance of the method. Instead, achieving a decent level of accuracy within a minimum number of training epochs is the primary concern.

For each network with different sizes and each training method, we measure \emph{TD} on all datasets. We consider only the high sparsity case (when $\varepsilon = 1$) since when the network is dense, all the methods have very low \emph{TD}, and the difference between them is negligible. However, when we are looking for a highly sparse sub-network, it takes longer for each method to find the well-performing sub-network, and the difference among the methods is more apparent. We set the threshold $th$ to $0.9$; therefore, we compute the training delay for reaching $0.9$ of the maximum accuracy achieved on this model. The results are presented in Table \ref{tab:speed}. If a method cannot reach the $th \times acc_{max}$ within the total number of epochs ($500$ in these experiments), we keep the corresponding entry empty.

As can be seen in Table \ref{tab:speed}, CTRE (including CTRE\textsubscript{sim} and CTRE\textsubscript{seq}) has the lowest training delay (\emph{TD}) in $13$ out of $18$ cases considered. On the Isolet and the Madelon datasets, some methods cannot reach the required level of accuracy ($0.9$ of the maximum accuracy) within the $500$ training epochs. SNIP has the worst performance among these methods and cannot reach the required level of accuracy on Madelon, Isolet, and CIFAR10. RigL has a similar performance to SNIP; while it has a decent performance on Fashion-MNIST and MNIST, it has a poor performance on the other datasets. Finally, SET has comparable performance to other methods on Fashion-MNIST and MNIST. However, when the network is highly sparse and large ($\varepsilon=1$ and $n^l > 100$) on the Isolet dataset, it does not have a good performance.

\input{supplementary/Tables/table_speed}

% Madelon

% Isolet    
    
% mnist    

% Fashion mnist    
% CIFAR10    

% CIFAR100 
    
%%%%%%%%%%%%%%%%%%%%%%%%%%%%%%%%%%%%%%%%%%%%%%%%%%%%%%%%%%%%%%%%%%
\newpage
\section{Computational Complexity}\label{appendix:computational_complexity}
%%%%%%%%%%%%%%%%%%%%%%%%%%%%%%%%%%%%%%%%%%%%%%%%%%%%%%%%%%%%%%%%%%
In this appendix, we compare the algorithms in terms of the computational complexity. While the computational cost during inference is equal for all methods (in the case of having the same sparsity level), the computational complexity during training is different.

We compare the computational complexity with the two closest sparse training algorithms to CTRE: SET and RigL. Our proposed methods require an extra cost of computing the cosine similarity matrix for the connections compared to SET. For each layer in each epoch, CTRE requires computing three dot products of size $m$ (number of samples) for each connection in this layer to compute similarity matrix in Equation \ref{eq:cosine_similarity}. Therefore, for each layer $l$, CTRE requires in the order of $\mathcal{O}(mN^l)$ extra computations at each epoch, where $N^l$ is the number of parameters of layer $l$. However, this additional cost considerably improves the accuracy and the learning speed (discussed in Appendix \ref{appendix:learning_curves}), particularly on tabular datasets and highly sparse neural networks. Therefore, depending on the application, the specialists should decide about the trade-off between accuracy and the computational cost when finding highly sparse neural networks. Compared to RigL, which requires computing occasional dense gradients, CTRE has the same order of complexity. This is because the order of computing gradients for back-propagation is also $\mathcal{O}(mN^l)$. However, CTRE outperforms RigL, especially in the high sparsity region. 

To further decrease the computational cost of CTRE, we have tried to reduce the cost of cosine similarity computation by considering a proportion of the samples to compute the similarity matrix. We run  CTRE\textsubscript{sim} with half of the samples ( CTRE\textsubscript{sample2}) and a quarter of samples ( CTRE\textsubscript{sample4}) to compute the cosine similarity matrix. The results can be observed in Table \ref{tab:results_samples}. It is clear that even with half of the samples, CTRE can still achieve a close performance as the original method on all datasets. Based on these observations, it can be concluded that only a fraction of samples can be used to compute the similarity matrix. In this way, we would be able to decrease the computational cost without affecting the performance.

\input{supplementary/Tables/table_cosine_sampling}

Further studies can be performed to decrease the computational cost of deriving the similarity matrix. In short, CTRE is a first step in finding highly sparse neural networks using neuron characteristics and can be further explored in future works.

\input{supplementary/Tables/table_num_hidden_epsilon_comparison}

%%%%%%%%%%%%%%%%%%%%%%%%%%%%%%%%%%%%%%%%%%%%%%%%%%%%%%%%%%%%%
\section{Performance Evaluation Using Pure Sparse Implementation}\label{appendix:pure_sparse}
%%%%%%%%%%%%%%%%%%%%%%%%%%%%%%%%%%%%%%%%%%%%%%%%%%%%%%%%%%%%%%
In this appendix, we present the results using the pure sparse implementation. This code is developed from the sparse implementation of SET\footnote{The pure sparse implementation of SET  can be found on \\\url{https://github.com/dcmocanu/sparse-evolutionary-artificial-neural-networks}.}. While the other training methods for obtaining sparse neural networks mostly use a binary mask over weights to simulate sparsity, this code is implemented in a purely sparse manner using Cython and SciPy sparse matrices. We have implemented our proposed method using this sparse implementation and repeated the experiments from Section \ref{ssec:performance_evaluation} in the manuscript. The results are summarized in Table \ref{tab:table_num_hidden_epsilon_comparison}.

As can be seen in Table \ref{tab:table_num_hidden_epsilon_comparison}, the results are subtly different from Table \ref{tab:results_comparison_rigl_implementation} in Section \ref{ssec:performance_evaluation}. This difference arise from some small differences in the implementation. One of the main differences is that in Section \ref{ssec:performance_evaluation}, Tensoflow library is used for implementing the neural network; however, this implementation uses Numpy, Scipy, and Cython to perform sparse matrix operations. Another difference is the weight initialization policy. While in the experiments of Section \ref{ssec:performance_evaluation}, weights are initialized using a uniform distribution, in the sparse implementation weights are initialized using a normal distribution which seems more beneficial to this implementation. Overall, in most cases, the results in Table \ref{tab:table_num_hidden_epsilon_comparison} are higher than the results in Table \ref{tab:results_comparison_rigl_implementation}.

%%%%%%%%%%%%%%%%%%%%%%%%%%%%%%%%%%%%%%%%%%%%%%%%%%%%%%%%%%%%%
\section{Cosine vs. Euclidean-based Similarity Metric}\label{appendix:ablation_euclidean}
%%%%%%%%%%%%%%%%%%%%%%%%%%%%%%%%%%%%%%%%%%%%%%%%%%%%%%%%%%%%%%
In this section, we analyze the effectiveness of cosine similarity metric in evolving the topology of the sparse neural network compared to other similarity metrics. To achieve this, we consider Euclidean-based similarity metric. Instead of computing the importance of the unexisting connections using cosine similarity (Equation \ref{eq:cosine_similarity}), we compute it as:

\begin{equation}\label{eq:euclidean_similarity}
 \emSim_{p,q}^{l} = \frac{1}{1 + d(\mA_{:, q}^{l}, \mA_{:, p}^{l-1})},
\end{equation}

where $d(a, b)$ is the Euclidean distance between vectors $a$ and $b$. We replace Equation \ref{eq:euclidean_similarity} with Equation \ref{eq:cosine_similarity} in Algorithm \ref{alg:CTRE_sim} and we call this method as CTRE\textsubscript{sim-euclidean}. We compare CTRE\textsubscript{sim-euclidean} with CTRE\textsubscript{sim}. The results are presented in Table \ref{tab:results_euclidean}.
\input{supplementary/Tables/table_cosine_sim_euclidean}

As shown in Table \ref{tab:results_euclidean}, CTRE\textsubscript{sim} outperforms CTRE\textsubscript{sim-euclidean} in most cases considered. Particularly, in the high sparsity regime ($\varepsilon = 1$), there is a considerable gap between the performance of these two methods. It can be concluded that Euclidean-based similarity metric is not very informative in evolving the topology of sparse neural networks. This might stem from the sensitivity of this metric to the vectors' magnitude. Cosine-similarity metric, a magnitude insensitive metric which also presents a biologically-plausible approach for measuring the importance of the connections, is a good choice for obtaining sparse neural networks in the $CTRE$ algorithm. 
%%%%%%%%%%%%%%%%%%%%%%%%%%%%%%%%%%%%%%%%%%%%%%%%%%%%%%%%%%%%%
\section{Ablation Study: Cosine and Random-based Weight Addition Order in CTRE\textsubscript{seq}}\label{appendix:ablation_CTRE_seq}
%%%%%%%%%%%%%%%%%%%%%%%%%%%%%%%%%%%%%%%%%%%%%%%%%%%%%%%%%%%%%%
In this section, we perform an ablation study on the CTRE\textsubscript{seq} algorithm to measure the effectiveness of the cosine-based and random search order in the performance of the algorithm. With this aim, we measure the performance of two variants of CTRE\textsubscript{seq} (Algorithm \ref{alg:CTRE_seq}) that are different from CTRE\textsubscript{seq} in the order of the cosine and random weight addition:

\begin{itemize}[]
    \item[$\bullet$] \textbf{CTRE\textsubscript{abl1:seq}} starts training with random weight addition, and switches to cosine-based addition when there is no improvement in the validation accuracy for $e_{early\;stop}$ epochs. 
    \item[$\bullet$] \textbf{CTRE\textsubscript{abl2:seq}} splits the training process into two equal phases in terms of the number of epochs. In the first phase it add weights randomly, and in the second phase it uses the cosine-similarity information for weight addition.
\end{itemize}

We compare the performance of these two algorithms with CTRE\textsubscript{seq} in Table \ref{tab:results_comparison_ablation_ctre_seq}. CTRE\textsubscript{seq} outperforms CTRE\textsubscript{abl1:seq} and CTRE\textsubscript{abl2:seq} in the majority of the experiments. In high sparsity regime and large network size ($\varepsilon =1$, $n^l=1000$), there is a considerable gap between their performance. The reason behind this difference is that by using cosine similarity-based weight addition at the beginning of the algorithm, the algorithm finds a well-performing sub-network very fast. Then, in the rest of the algorithm it improves this topology using cosine information and random search. However, by starting with random weight addition, it might take longer for the algorithm to reach a reasonable level of performance. This results in a lower accuracy than CTRE\textsubscript{seq} at the end of training.

To summarize, while starting with cosine similarity-based weight addition and switching to random search might seem counter-intuitive, we show that this strategy is beneficial for the CTRE algorithm.

\input{supplementary/Tables/table_cosine_seq_random_ablation}

%% file: supplementary/Tables/table_speed.tex
\begin{table}[!b]
    \caption{Training delay (TD) (\%) comparison among methods. Empty fields indicate that the method cannot reach the considered level of accuracy in $500$ training epochs. } \label{tab:speed}

    \begin{center}
    \begin{scriptsize}
    \scalebox{0.9}{
    \begin{tabular}{c@{\hskip 0.07in}c@{\hskip 0.07in}c@{\hskip 0.05in}c@{\hskip 0.05in}c@{\hskip 0.12in}c@{\hskip 0.05in}c@{\hskip 0.05in}c@{\hskip 0.12in}c@{\hskip 0.05in}c@{\hskip 0.05in}c@{\hskip 0.12in}c@{\hskip 0.05in}c@{\hskip 0.0in}c@{\hskip 0.12in}c@{\hskip 0.05in}c@{\hskip 0.05in}c@{\hskip 0.12in}c@{\hskip 0.05in}c@{\hskip 0.05in}c@{}}
        \toprule
        & & \multicolumn{3}{c}{\bt Madelon} & \multicolumn{3}{c}{\bt Isolet} & \multicolumn{3}{c}{\bt MNIST}& \multicolumn{3}{c}{\bt Fashion-MNIST}& \multicolumn{3}{c}{\bt CIFAR10} & \multicolumn{3}{c}{\bt CIFAR100} \\ \cmidrule(lr){3-5}\cmidrule(lr){6-8}\cmidrule(lr){9-11}\cmidrule(lr){12-14}\cmidrule(lr){15-17} \cmidrule(lr){18-20}
         %\bt Threshold & 
         %\multicolumn{3}{c}{$\epsilon$} & \multicolumn{3}{c}{$\epsilon$} &
         %\multicolumn{3}{c}{$\epsilon$}&
         %\multicolumn{3}{c}{$\epsilon$} &
         %\multicolumn{3}{c}{$\epsilon$}\\ \midrule
        & & \multicolumn{3}{c}{ $ \pmb{n^l}$} & \multicolumn{3}{c}{ $ \pmb{n^l}$} &
         \multicolumn{3}{c}{$ \pmb{n^l}$} &\multicolumn{3}{c}{$ \pmb{n^l}$} &\multicolumn{3}{c}{$ \pmb{n^l}$}\\ 
         $\pmb{\varepsilon}$&\bt Method &\bt 100 &\bt 500 & \bt1000 & \bt100 & \bt500 & \bt1000 & \bt100 & \bt500 & \bt1000 & \bt100 & \bt500 & \bt1000 & \bt100 & \bt500 & \bt1000  & \bt100 & \bt500 & \bt1000  \\\midrule

&\multicolumn{1}{c}{SNIP}&$-$&$-$&$-$&$-$&$-$&$-$&\bt 1.8 &$2.8$&$3.2$&$5.6$&$8.8$&$8.6$&$-$&$-$&$-$&$-$&$-$&$-$\\
&\multicolumn{1}{c}{RigL}&$-$&$-$&$-$&$-$&$-$&$-$&$3.2$&$5.8$&$9.6$&\bt2.8 &\bt 3.6 & \bt4.6 &$25.2$&$65.8$&$-$&$-$&$-$&$-$\\
1&\multicolumn{1}{c}{SET}&$-$&$-$&$-$&$40.0$&$74.0$&$-$&$4.4$&$11.8$&$18.2$&$3.4$&$9.4$&$14.2$& \bt11.4&$39.4$&$68.4$&\bt 40.2 &$-$&$-$\\
&\multicolumn{1}{c}{CTRE\textsubscript{seq}}&$16.0$&$-$&$-$&$48.0$&$52.8$&$52.4$&$2.2$& \bt2.6 &\bt 3.0 &$3.6$&$9.2$&$12.8$&$13.4$& \bt 14.6 &\bt 16.8 &$-$& \bt40.4 &\bt 36.6\\
&\multicolumn{1}{c}{CTRE\textsubscript{sim}}&\bt 9.8 & \bt 20.6 &\bt 24.8&\bt 38.6&\bt30.6&\bt35.6 &$2.2$&$3.0$&$3.2$&$3.0$&$6.2$&$7.2$& \bt11.4 &$22.4$&$33.2$&$66.8$&$88.6$&$-$\\
\midrule

    \end{tabular}}
    \end{scriptsize}
    \end{center}
\end{table}

%% file: supplementary/Tables/table_cosine_sampling.tex
%\newrobustcmd{\bt}{\bfseries} % bold command
\begin{table}[!b]
    \centering
    \caption{Classification accuracy (\%) of CTRE\textsubscript{sim} with different number of training samples for computing the similarity matrices.
    } \label{tab:results_samples}

    \begin{scriptsize}
    \scalebox{0.96}{
    \begin{tabular}{@{}c@{\hskip 0.02in}c@{\hskip 0.03in}c@{\hskip 0.03in}c@{\hskip 0.03in}c@{\hskip 0.07in}c@{\hskip 0.03in}c@{\hskip 0.03in}c@{\hskip 0.07in}c@{\hskip 0.03in}c@{\hskip 0.03in}c@{}}
        \toprule
         &  & \multicolumn{3}{c}{\pmb{ $n^l=100$}} & \multicolumn{3}{c}{\pmb{ $n^l=500$}} & \multicolumn{3}{c}{\pmb{ $n^l=1000$}} \\ \cmidrule(l){3-5}\cmidrule(l){6-8}\cmidrule(l){9-11}
         &  & \multicolumn{3}{c}{$\epsilon$} & \multicolumn{3}{c}{$\epsilon$} & \multicolumn{3}{c}{$\epsilon$}\\ %\cmidrule(c){3-5}
         %\cmidrule(c){6-8} \cmidrule(c){9-11}  \cmidrule(c){12-14}\\
        \bt Dataset & \bt Method &\bt 1 &\bt 5 & \bt13 & \bt1 & \bt5 & \bt13 & \bt1 & \bt5 & \bt13  \\ \midrule

\multicolumn{1}{c}{Madelon}&CTRE\textsubscript{sim}&$81.6\pm1.3$&\pmb{$73.0\pm1.6$}&\pmb{$65.6\pm3.0$}&$79.4\pm1.7$&$77.7\pm1.4$&\pmb{$73.0\pm1.5$}&$78.8\pm2.2$&$78.5\pm1.0$&\pmb{$74.6\pm1.4$}\\
\multicolumn{1}{c}{}&CTRE\textsubscript{sample2}&$81.0\pm2.3$&$72.4\pm1.4$&$65.3\pm1.6$&\pmb{$79.7\pm0.5$}&\pmb{$77.9\pm0.9$}&$71.7\pm2.8$&\pmb{$80.6\pm0.7$}&$78.6\pm1.0$&$74.1\pm0.4$\\
\multicolumn{1}{c}{}&CTRE\textsubscript{sample4}&\pmb{$81.9\pm1.0$}&$71.3\pm1.8$&$64.4\pm1.0$&$79.0\pm1.1$&$77.8\pm0.8$&$72.7\pm1.6$&$78.7\pm1.1$&\pmb{$79.1\pm1.4$}&$74.1\pm0.7$\\
\midrule

\multicolumn{1}{c}{Isolet}&CTRE\textsubscript{sim}&\pmb{$87.5\pm0.8$}&\pmb{$93.4\pm0.7$}&$94.3\pm0.8$&$87.8\pm1.1$&$91.7\pm1.1$&$93.9\pm0.5$&$88.3\pm0.7$&$91.3\pm1.3$&$93.1\pm0.6$\\
\multicolumn{1}{c}{}&CTRE\textsubscript{sample2}&$81.9\pm4.2$&$92.8\pm0.3$&$94.7\pm0.3$&\pmb{$88.4\pm0.8$}&\pmb{$92.6\pm0.4$}&\pmb{$94.0\pm1.1$}&$88.7\pm0.5$&$92.4\pm0.7$&\pmb{$94.0\pm0.2$}\\
\multicolumn{1}{c}{}&CTRE\textsubscript{sample4}&$85.0\pm1.3$&$93.0\pm0.8$&\pmb{$94.9\pm0.5$}&$87.9\pm0.5$&$91.7\pm1.1$&$93.8\pm0.1$&\pmb{$89.4\pm1.0$}&\pmb{$92.5\pm0.2$}&$93.8\pm0.5$\\
\midrule

\multicolumn{1}{c}{MNIST}&CTRE\textsubscript{sim}&$95.5\pm0.2$&\pmb{$97.3\pm0.1$}&\pmb{$97.8\pm0.1$}&\pmb{$96.4\pm0.2$}&$97.7\pm0.1$&\pmb{$98.0\pm0.1$}&\pmb{$96.6\pm0.2$}&$97.7\pm0.1$&\pmb{$97.9\pm0.1$}\\
\multicolumn{1}{c}{}&CTRE\textsubscript{sample2}&\pmb{$95.7\pm0.0$}&$97.0\pm0.1$&$97.7\pm0.0$&$96.3\pm0.0$&\pmb{$97.8\pm0.0$}&$97.8\pm0.0$&$96.2\pm0.0$&\pmb{$97.9\pm0.1$}&$97.8\pm0.1$\\
\multicolumn{1}{c}{}&CTRE\textsubscript{sample4}&$95.1\pm0.2$&\pmb{$97.3\pm0.1$}&$97.6\pm0.2$&$96.1\pm0.2$&$97.7\pm0.1$&$97.8\pm0.1$&\pmb{$96.6\pm0.1$}&$97.7\pm0.1$&$97.8\pm0.0$\\
\midrule

\multirow{2}{*}{\shortstack{Fashion-\\MNIST}}&CTRE\textsubscript{sim}&\pmb{$85.8\pm0.3$}&$87.5\pm0.3$&\pmb{$88.0\pm0.2$}&$86.4\pm0.5$&\pmb{$88.1\pm0.2$}&$88.3\pm0.2$&$86.5\pm0.3$&$88.1\pm0.3$&$88.3\pm0.2$\\
\multicolumn{1}{c}{}&CTRE\textsubscript{sample2}&$85.7\pm0.1$&$87.4\pm0.1$&$87.7\pm0.3$&$86.3\pm0.7$&$87.9\pm0.1$&$88.1\pm0.2$&$86.3\pm0.2$&\pmb{$88.3\pm0.0$}&$88.0\pm0.2$\\
\multicolumn{1}{c}{}&CTRE\textsubscript{sample4}&$85.6\pm0.4$&\pmb{$87.6\pm0.1$}&\pmb{$88.0\pm0.1$}&\pmb{$87.2\pm0.1$}&$87.5\pm0.1$&\pmb{$88.6\pm0.1$}&\pmb{$86.6\pm0.3$}&$88.0\pm0.1$&\pmb{$88.4\pm0.2$}\\
\midrule

\multicolumn{1}{c}{CIFAR10}&CTRE\textsubscript{sim}&\pmb{$48.2\pm0.4$}&$50.3\pm0.3$&$50.5\pm0.4$&$50.6\pm0.4$&$52.6\pm0.7$&$52.5\pm0.7$&$50.0\pm0.4$&$52.7\pm0.5$&$53.5\pm0.5$\\
\multicolumn{1}{c}{}&CTRE\textsubscript{sample2}&$47.8\pm0.1$&$49.4\pm0.4$&$50.9\pm0.4$&$50.9\pm0.3$&\pmb{$52.8\pm0.4$}&\pmb{$52.9\pm0.4$}&$50.4\pm0.5$&\pmb{$53.9\pm0.1$}&$53.4\pm0.2$\\
\multicolumn{1}{c}{}&CTRE\textsubscript{sample4}&$47.8\pm0.1$&\pmb{$50.8\pm0.2$}&\pmb{$51.1\pm0.4$}&\pmb{$52.0\pm0.1$}&$52.7\pm0.3$&\pmb{$52.9\pm0.6$}&\pmb{$51.6\pm0.5$}&$53.6\pm0.3$&\pmb{$53.6\pm0.5$}\\
\midrule

    \end{tabular}}
    \end{scriptsize}
\end{table}

%% file: supplementary/Tables/table_num_hidden_epsilon_comparison.tex
\begin{table}[!t]
    
    \caption{Classification accuracy (\%) comparison using pure sparse implementation.
    }
    \label{tab:table_num_hidden_epsilon_comparison}

    \begin{scriptsize}
    \scalebox{0.9}{  
    \begin{tabular}{@{}c@{\hskip 0.02in}c@{\hskip 0.03in}c@{\hskip 0.03in}c@{\hskip 0.03in}c@{\hskip 0.07in}c@{\hskip 0.03in}c@{\hskip 0.03in}c@{\hskip 0.07in}c@{\hskip 0.03in}c@{\hskip 0.03in}c@{}}
        \toprule
         &  & \multicolumn{3}{c}{\pmb{ $n^l=100$}} & \multicolumn{3}{c}{\pmb{ $n^l=500$}} & \multicolumn{3}{c}{\pmb{ $n^l=1000$}} \\ \cmidrule(l){3-5}\cmidrule(l){6-8}\cmidrule(l){9-11}
         &  & \multicolumn{3}{c}{$\epsilon$} & \multicolumn{3}{c}{$\epsilon$} & \multicolumn{3}{c}{$\epsilon$}\\ %\cmidrule(c){3-5}
         %\cmidrule(c){6-8} \cmidrule(c){9-11}  \cmidrule(c){12-14}\\
        \bt Dataset & \bt Method &\bt 1 &\bt 5 & \bt13 & \bt1 & \bt5 & \bt13 & \bt1 & \bt5 & \bt13  \\ \midrule

\multicolumn{1}{c}{Madelon}&SET&$58.7\pm2.0$&$60.4\pm3.3$&$58.1\pm1.5$&$65.1\pm2.2$&$59.6\pm5.1$&$61.8\pm1.0$&$61.8\pm1.9$&$62.1\pm4.7$&$61.6\pm3.9$\\
\multicolumn{1}{c}{}&CTRE\textsubscript{seq}&$85.3\pm0.3$&$75.1\pm0.6$&$67.1\pm1.4$&\pmb{$87.2\pm1.2$}&$82.1\pm2.0$&$75.3\pm1.0$&$87.2\pm0.2$&\pmb{$86.6\pm0.7$}&$75.7\pm2.7$\\
\multicolumn{1}{c}{}&CTRE\textsubscript{sim}&$82.9\pm1.0$&$73.9\pm2.3$&$66.8\pm2.7$&$82.5\pm1.1$&$79.4\pm1.0$&$74.9\pm1.2$&$82.9\pm1.1$&$80.4\pm1.6$&$74.6\pm0.6$\\
\multicolumn{1}{c}{}&CTRE\textsubscript{w/oRandom}&\pmb{$86.4\pm1.2$}&\pmb{$75.6\pm1.2$}&\pmb{$68.5\pm0.6$}&\pmb{$87.2\pm1.2$}&\pmb{$82.8\pm1.1$}&\pmb{$77.0\pm0.6$}&\pmb{$88.2\pm0.6$}&$84.5\pm0.6$&\pmb{$77.9\pm1.4$}\\
\midrule

\multicolumn{1}{c}{Isolet}&SET&\pmb{$87.6\pm1.0$}&\pmb{$94.1\pm0.1$}&$94.5\pm0.4$&$86.2\pm1.2$&$93.7\pm0.7$&$94.4\pm0.1$&$86.1\pm0.5$&$94.1\pm0.1$&$94.7\pm0.1$\\
\multicolumn{1}{c}{}&CTRE\textsubscript{seq}&$83.1\pm1.2$&$93.7\pm0.4$&\pmb{$94.6\pm0.7$}&\pmb{$91.1\pm0.3$}&\pmb{$94.2\pm0.3$}&\pmb{$94.5\pm0.4$}&\pmb{$90.9\pm1.1$}&\pmb{$94.3\pm0.4$}&\pmb{$94.8\pm0.3$}\\
\multicolumn{1}{c}{}&CTRE\textsubscript{sim}&$85.6\pm0.6$&$93.2\pm0.7$&\pmb{$94.6\pm0.6$}&$88.6\pm1.3$&$93.7\pm0.4$&$94.4\pm0.2$&$88.8\pm0.6$&$93.5\pm0.1$&$94.0\pm0.1$\\
\multicolumn{1}{c}{}&CTRE\textsubscript{w/oRandom}&$81.1\pm1.1$&$92.9\pm1.6$&\pmb{$94.6\pm0.2$}&$89.1\pm0.5$&$93.8\pm0.1$&$93.9\pm0.3$&$90.6\pm0.9$&$93.9\pm0.4$&$93.4\pm0.7$\\
\midrule

\multicolumn{1}{c}{MNIST}&SET&$95.0\pm0.2$&$97.5\pm0.1$&$97.8\pm0.1$&$94.9\pm0.2$&$97.5\pm0.0$&$97.8\pm0.1$&$95.2\pm0.2$&$97.3\pm0.1$&$97.6\pm0.0$\\
\multicolumn{1}{c}{}&CTRE\textsubscript{seq}&\pmb{$95.7\pm0.2$}&\pmb{$97.8\pm0.1$}&\pmb{$97.9\pm0.0$}&$97.3\pm0.2$&\pmb{$97.9\pm0.2$}&\pmb{$98.0\pm0.0$}&\pmb{$97.6\pm0.0$}&\pmb{$97.9\pm0.1$}&\pmb{$97.9\pm0.1$}\\
\multicolumn{1}{c}{}&CTRE\textsubscript{sim}&$95.4\pm0.1$&$97.6\pm0.0$&$97.7\pm0.0$&$96.4\pm0.6$&$97.4\pm0.1$&$97.8\pm0.0$&$96.7\pm0.6$&$97.2\pm0.0$&$97.7\pm0.1$\\
\multicolumn{1}{c}{}&CTRE\textsubscript{w/oRandom}&$95.5\pm0.1$&$97.4\pm0.0$&$97.7\pm0.1$&\pmb{$97.4\pm0.1$}&$97.7\pm0.1$&$97.5\pm0.2$&\pmb{$97.6\pm0.1$}&$97.8\pm0.0$&$97.8\pm0.2$\\
\midrule

\multirow{2}{*}{\shortstack{Fashion-\\MNIST} }&SET&\pmb{$85.6\pm0.1$}&$87.7\pm0.1$&\pmb{$88.5\pm0.1$}&$85.2\pm0.2$&$87.9\pm0.1$&$88.4\pm0.3$&$85.2\pm0.2$&$87.8\pm0.0$&$88.5\pm0.2$\\
\multicolumn{1}{c}{}&CTRE\textsubscript{seq}&$85.4\pm0.4$&\pmb{$88.0\pm0.1$}&$88.4\pm0.0$&\pmb{$86.8\pm0.1$}&\pmb{$88.5\pm0.2$}&\pmb{$88.6\pm0.1$}&\pmb{$87.1\pm0.4$}&\pmb{$88.6\pm0.2$}&\pmb{$88.7\pm0.2$}\\
\multicolumn{1}{c}{}&CTRE\textsubscript{sim}&$84.9\pm0.8$&$87.4\pm0.3$&$88.1\pm0.1$&$85.7\pm0.6$&$87.7\pm0.1$&$88.2\pm0.1$&$85.9\pm0.7$&$87.5\pm0.1$&$88.4\pm0.2$\\
\multicolumn{1}{c}{}&CTRE\textsubscript{w/oRandom}&$83.7\pm0.2$&$87.3\pm0.2$&$87.7\pm0.2$&$85.9\pm0.2$&$87.9\pm0.3$&$88.0\pm0.1$&$86.5\pm0.4$&$88.0\pm0.2$&$87.9\pm0.0$\\
\midrule

\multicolumn{1}{c}{CIFAR10}&SET&$48.5\pm0.4$&$52.7\pm0.5$&\pmb{$54.0\pm0.4$}&$47.7\pm0.4$&$53.3\pm0.3$&$54.3\pm0.4$&$46.2\pm0.2$&$52.8\pm0.2$&$54.5\pm0.7$\\
\multicolumn{1}{c}{}&CTRE\textsubscript{seq}&$48.5\pm0.5$&\pmb{$53.3\pm0.3$}&$53.0\pm0.6$&\pmb{$52.2\pm0.3$}&\pmb{$55.7\pm0.6$}&\pmb{$55.6\pm0.3$}&\pmb{$54.2\pm0.3$}&\pmb{$55.8\pm0.1$}&\pmb{$56.2\pm0.1$}\\
\multicolumn{1}{c}{}&CTRE\textsubscript{sim}&\pmb{$48.6\pm0.0$}&$51.0\pm0.0$&$51.3\pm0.0$&$50.1\pm0.6$&$55.4\pm0.1$&$54.3\pm0.2$&$50.2\pm0.9$&\pmb{$55.8\pm0.4$}&$55.8\pm0.0$\\
\multicolumn{1}{c}{}&CTRE\textsubscript{w/oRandom}&$45.5\pm0.5$&$49.6\pm0.3$&$49.0\pm0.4$&$51.2\pm0.3$&$53.5\pm0.3$&$53.0\pm0.1$&$53.8\pm0.4$&$53.6\pm0.0$&$54.7\pm0.1$\\
\midrule

    \end{tabular}}
    \end{scriptsize}
\end{table}

%% file: supplementary/Tables/table_cosine_sim_euclidean.tex
%\newrobustcmd{\bt}{\bfseries} % bold command
\begin{table}[!t]
    
    \caption{Classification accuracy (\%) comparison among cosine and euclidean-based similarity metrics in $CTRE$ algorithm.} \label{tab:results_euclidean}

    \begin{scriptsize}
    \scalebox{0.9}{  
    \begin{tabular}{@{}c@{\hskip 0.02in}c@{\hskip 0.04in}c@{\hskip 0.04in}c@{\hskip 0.04in}c@{\hskip 0.07in}c@{\hskip 0.04in}c@{\hskip 0.04in}c@{\hskip 0.07in}c@{\hskip 0.04in}c@{\hskip 0.04in}c@{}}
        \toprule
         &  & \multicolumn{3}{c}{\pmb{ $n^l=100$}} & \multicolumn{3}{c}{\pmb{ $n^l=500$}} & \multicolumn{3}{c}{\pmb{ $n^l=1000$}} \\ \cmidrule(l){3-5}\cmidrule(l){6-8}\cmidrule(l){9-11}
         &  & \multicolumn{3}{c}{$\epsilon$} & \multicolumn{3}{c}{$\epsilon$} & \multicolumn{3}{c}{$\epsilon$}\\ 
        \bt Dataset & \bt Method &\bt 1 &\bt 5 & \bt13 & \bt1 & \bt5 & \bt13 & \bt1 & \bt5 & \bt13  \\ \midrule

\multicolumn{1}{c}{Madelon}&CTRE\textsubscript{sim}&\pmb{$81.6\pm1.3$}&\pmb{$73.0\pm1.6$}&\pmb{$65.6\pm3.0$}&\pmb{$79.4\pm1.7$}&\pmb{$77.7\pm1.4$}&\pmb{$73.0\pm1.5$}&\pmb{$78.8\pm2.2$}&\pmb{$78.5\pm1.0$}&\pmb{$74.6\pm1.4$}\\
\multicolumn{1}{c}{}&CTRE\textsubscript{sim-euclidean}&$77.2\pm1.7$&$65.6\pm1.3$&$63.7\pm0.7$&$71.5\pm4.2$&$74.2\pm1.6$&$65.8\pm4.3$&$70.6\pm4.5$&$76.9\pm1.8$&$67.8\pm2.2$\\
\midrule

\multicolumn{1}{c}{Isolet}&CTRE\textsubscript{sim}&\pmb{$87.5\pm0.8$}&$93.4\pm0.7$&$94.3\pm0.8$&\pmb{$87.8\pm1.1$}&\pmb{$91.7\pm1.1$}&\pmb{$93.9\pm0.5$}&\pmb{$88.3\pm0.7$}&$91.3\pm1.3$&$93.1\pm0.6$\\
\multicolumn{1}{c}{}&CTRE\textsubscript{sim-euclidean}&$83.0\pm1.3$&\pmb{$93.7\pm0.6$}&\pmb{$94.5\pm0.7$}&$72.0\pm4.1$&$91.6\pm0.3$&$93.7\pm0.8$&$49.8\pm6.1$&\pmb{$91.4\pm0.7$}&\pmb{$93.6\pm0.5$}\\
\midrule

\multicolumn{1}{c}{MNIST}&CTRE\textsubscript{sim}&\pmb{$95.5\pm0.2$}&\pmb{$97.3\pm0.1$}&\pmb{$97.8\pm0.1$}&\pmb{$96.4\pm0.2$}&\pmb{$97.7\pm0.1$}&\pmb{$98.0\pm0.1$}&\pmb{$96.6\pm0.2$}&\pmb{$97.7\pm0.1$}&\pmb{$97.9\pm0.1$}\\
\multicolumn{1}{c}{}&CTRE\textsubscript{sim-euclidean}&$94.6\pm0.2$&$97.0\pm0.1$&$97.3\pm0.1$&$94.3\pm0.2$&$97.2\pm0.1$&$97.4\pm0.0$&$93.8\pm0.1$&$96.9\pm0.0$&$97.1\pm0.3$\\
\midrule

\multirow{2}{*}{\shortstack{Fashion-\\MNIST}}&CTRE\textsubscript{sim}&\pmb{$85.8\pm0.3$}&\pmb{$87.5\pm0.3$}&\pmb{$88.0\pm0.2$}&\pmb{$86.4\pm0.5$}&\pmb{$88.1\pm0.2$}&\pmb{$88.3\pm0.2$}&\pmb{$86.5\pm0.3$}&\pmb{$88.1\pm0.3$}&\pmb{$88.3\pm0.2$}\\
\multicolumn{1}{c}{}&CTRE\textsubscript{sim-euclidean}&$84.9\pm0.2$&$87.1\pm0.1$&$87.7\pm0.1$&$84.6\pm0.2$&$87.4\pm0.2$&$87.8\pm0.2$&$83.8\pm0.3$&$87.4\pm0.1$&$87.4\pm0.4$\\
\midrule

\multicolumn{1}{c}{CIFAR10}&CTRE\textsubscript{sim}&\pmb{$48.2\pm0.4$}&\pmb{$50.3\pm0.3$}&$50.5\pm0.4$&\pmb{$50.6\pm0.4$}&\pmb{$52.6\pm0.7$}&\pmb{$52.5\pm0.7$}&\pmb{$50.0\pm0.4$}&\pmb{$52.7\pm0.5$}&\pmb{$53.5\pm0.5$}\\
\multicolumn{1}{c}{}&CTRE\textsubscript{sim-euclidean}&$46.1\pm0.4$&$50.1\pm0.4$&\pmb{$50.7\pm0.4$}&$46.1\pm1.4$&$50.5\pm0.5$&$51.6\pm0.2$&$44.8\pm0.5$&$49.5\pm0.6$&$51.2\pm0.3$\\
\midrule

\multicolumn{1}{c}{CIFAR100}&CTRE\textsubscript{sim}&\pmb{$13.8\pm0.4$}&$20.6\pm0.4$&\pmb{$21.9\pm0.4$}&\pmb{$17.0\pm0.3$}&\pmb{$23.0\pm0.3$}&\pmb{$24.6\pm0.4$}&\pmb{$17.3\pm0.4$}&\pmb{$23.5\pm0.3$}&\pmb{$25.1\pm0.3$}\\
\multicolumn{1}{c}{}&CTRE\textsubscript{sim-euclidean}&$12.2\pm0.4$&\pmb{$20.7\pm0.4$}&$21.8\pm0.3$&$11.8\pm0.7$&$22.5\pm0.3$&$24.0\pm0.2$&$9.4\pm1.0$&$21.9\pm0.3$&$23.8\pm0.4$\\
\midrule

    \end{tabular}}
    \end{scriptsize}
\end{table}

%% file: supplementary/Tables/table_cosine_seq_random_ablation.tex
%\newrobustcmd{\bt}{\bfseries} % bold command
\begin{table}[!t]
    
    \caption{Classification accuracy (\%) comparison among variants of $CTRE_{seq}$ algorithm.} \label{tab:results_comparison_ablation_ctre_seq}

    \begin{scriptsize}
    \scalebox{0.9}{  
    \begin{tabular}{@{}c@{\hskip 0.02in}c@{\hskip 0.04in}c@{\hskip 0.04in}c@{\hskip 0.04in}c@{\hskip 0.07in}c@{\hskip 0.04in}c@{\hskip 0.04in}c@{\hskip 0.07in}c@{\hskip 0.04in}c@{\hskip 0.04in}c@{}}
        \toprule
         &  & \multicolumn{3}{c}{\pmb{ $n^l=100$}} & \multicolumn{3}{c}{\pmb{ $n^l=500$}} & \multicolumn{3}{c}{\pmb{ $n^l=1000$}} \\ \cmidrule(l){3-5}\cmidrule(l){6-8}\cmidrule(l){9-11}
         &  & \multicolumn{3}{c}{$\epsilon$} & \multicolumn{3}{c}{$\epsilon$} & \multicolumn{3}{c}{$\epsilon$}\\ %\cmidrule(c){3-5}
         %\cmidrule(c){6-8} \cmidrule(c){9-11}  \cmidrule(c){12-14}\\
        \bt Dataset & \bt Method &\bt 1 &\bt 5 & \bt13 & \bt1 & \bt5 & \bt13 & \bt1 & \bt5 & \bt13  \\ \midrule

\multicolumn{1}{c}{Madelon}&CTRE\textsubscript{seq}&\pmb{$82.2\pm2.4$}&\pmb{$72.5\pm2.0$}&\pmb{$63.9\pm1.2$}&\pmb{$61.2\pm2.4$}&\pmb{$81.5\pm1.4$}&\pmb{$71.8\pm2.0$}&$61.1\pm1.9$&\pmb{$83.9\pm2.0$}&\pmb{$76.5\pm1.9$}\\
\multicolumn{1}{c}{}&CTRE\textsubscript{abl1:seq}&$57.8\pm2.3$&$58.1\pm3.1$&$57.8\pm2.1$&$61.0\pm3.1$&$59.6\pm3.0$&$59.4\pm2.0$&$61.1\pm2.9$&$59.7\pm1.1$&$58.3\pm2.4$\\
\multicolumn{1}{c}{}&CTRE\textsubscript{abl2:seq}&$59.2\pm3.0$&$58.0\pm3.5$&$58.4\pm2.3$&$61.1\pm2.0$&$58.7\pm2.9$&$58.8\pm2.4$&\pmb{$61.6\pm0.9$}&$58.4\pm1.6$&$58.7\pm1.9$\\
\midrule

\multicolumn{1}{c}{Isolet}&CTRE\textsubscript{seq}&$86.7\pm1.8$&$92.4\pm1.1$&$94.3\pm0.5$&\pmb{$87.2\pm2.0$}&$92.3\pm0.6$&$94.0\pm0.4$&\pmb{$86.7\pm1.6$}&$91.5\pm1.0$&\pmb{$93.7\pm0.5$}\\
\multicolumn{1}{c}{}&CTRE\textsubscript{abl1:seq}&\pmb{$89.5\pm1.1$}&$93.7\pm0.5$&\pmb{$94.9\pm0.5$}&$85.8\pm1.0$&\pmb{$93.3\pm0.8$}&\pmb{$94.4\pm0.4$}&$72.5\pm2.3$&$92.3\pm0.6$&$93.5\pm0.9$\\
\multicolumn{1}{c}{}&CTRE\textsubscript{abl2:seq}&$89.0\pm1.3$&\pmb{$93.8\pm0.6$}&$94.6\pm0.7$&$85.8\pm1.5$&$93.1\pm0.8$&$94.2\pm0.9$&$75.1\pm3.9$&\pmb{$92.4\pm0.9$}&$93.3\pm0.7$\\
\midrule

\multicolumn{1}{c}{MNIST}&CTRE\textsubscript{seq}&\pmb{$95.7\pm0.2$}&\pmb{$97.3\pm0.2$}&\pmb{$97.7\pm0.1$}&\pmb{$97.0\pm0.2$}&\pmb{$97.6\pm0.2$}&\pmb{$97.8\pm0.1$}&\pmb{$97.3\pm0.1$}&\pmb{$97.7\pm0.1$}&\pmb{$97.8\pm0.1$}\\
\multicolumn{1}{c}{}&CTRE\textsubscript{abl1:seq}&$95.5\pm0.3$&$97.1\pm0.1$&$97.6\pm0.1$&$95.8\pm0.2$&$97.3\pm0.1$&$97.5\pm0.1$&$95.9\pm0.1$&$97.3\pm0.1$&$97.2\pm0.1$\\
\multicolumn{1}{c}{}&CTRE\textsubscript{abl2:seq}&$95.5\pm0.2$&$97.1\pm0.1$&$97.6\pm0.1$&$96.1\pm0.1$&$97.4\pm0.2$&$97.5\pm0.1$&$96.0\pm0.1$&$97.4\pm0.1$&$97.3\pm0.0$\\
\midrule

\multirow{2}{*}{\shortstack{Fashion-\\MNIST}}&CTRE\textsubscript{seq}&$85.8\pm0.5$&\pmb{$87.5\pm0.3$}&$87.4\pm0.3$&\pmb{$87.1\pm0.4$}&\pmb{$87.9\pm0.3$}&\pmb{$88.0\pm0.2$}&\pmb{$87.3\pm0.2$}&\pmb{$88.0\pm0.3$}&\pmb{$88.3\pm0.2$}\\
\multicolumn{1}{c}{}&CTRE\textsubscript{abl1:seq}&$85.7\pm0.2$&\pmb{$87.5\pm0.2$}&$87.8\pm0.2$&$86.3\pm0.3$&$87.7\pm0.2$&$87.9\pm0.3$&$86.0\pm0.2$&$87.7\pm0.3$&$87.5\pm0.3$\\
\multicolumn{1}{c}{}&CTRE\textsubscript{abl2:seq}&\pmb{$85.9\pm0.3$}&$87.4\pm0.2$&\pmb{$87.9\pm0.2$}&$86.4\pm0.2$&$87.7\pm0.0$&$87.7\pm0.2$&$86.2\pm0.1$&$87.7\pm0.1$&$87.6\pm0.2$\\
\midrule

\multicolumn{1}{c}{CIFAR10}&CTRE\textsubscript{seq}&$47.9\pm0.8$&\pmb{$50.1\pm0.5$}&$50.1\pm0.5$&\pmb{$51.3\pm0.5$}&\pmb{$52.7\pm0.5$}&\pmb{$52.6\pm0.6$}&\pmb{$52.0\pm0.7$}&\pmb{$53.2\pm0.6$}&\pmb{$53.6\pm0.6$}\\
\multicolumn{1}{c}{}&CTRE\textsubscript{abl1:seq}&\pmb{$48.1\pm0.4$}&$49.3\pm0.6$&\pmb{$50.5\pm0.4$}&$49.3\pm0.4$&$50.9\pm0.4$&$51.1\pm0.5$&$48.9\pm0.4$&$50.9\pm0.4$&$50.8\pm0.7$\\
\multicolumn{1}{c}{}&CTRE\textsubscript{abl2:seq}&$47.8\pm0.2$&$49.9\pm0.4$&$50.3\pm0.2$&$49.2\pm0.4$&$50.8\pm0.4$&$51.1\pm0.2$&$48.6\pm0.1$&$51.0\pm0.1$&$50.7\pm0.3$\\
\midrule

\multicolumn{1}{c}{CIFAR100}&CTRE\textsubscript{seq}&$12.7\pm0.7$&\pmb{$21.1\pm0.3$}&\pmb{$21.8\pm0.5$}&\pmb{$18.7\pm0.4$}&\pmb{$23.6\pm0.4$}&\pmb{$24.0\pm0.4$}&\pmb{$21.4\pm0.4$}&\pmb{$23.9\pm0.5$}&\pmb{$24.7\pm0.4$}\\
\multicolumn{1}{c}{}&CTRE\textsubscript{abl1:seq}&\pmb{$14.9\pm0.6$}&$20.5\pm0.3$&$21.5\pm0.5$&$13.9\pm2.3$&$22.8\pm0.3$&$23.8\pm0.6$&$1.0\pm0.0$&$23.6\pm0.3$&$24.3\pm0.4$\\
\multicolumn{1}{c}{}&CTRE\textsubscript{abl2:seq}&$14.5\pm0.3$&$20.4\pm0.1$&$21.4\pm0.3$&$14.1\pm0.9$&$22.8\pm0.0$&\pmb{$24.0\pm0.1$}&$1.0\pm0.0$&$23.4\pm0.4$&$24.6\pm0.2$\\
\midrule

    \end{tabular}}
    \end{scriptsize}
\end{table}

%% file: main.bbl
\begin{thebibliography}{77}
\providecommand{\natexlab}[1]{#1}
\providecommand{\url}[1]{\texttt{#1}}
\expandafter\ifx\csname urlstyle\endcsname\relax
  \providecommand{\doi}[1]{doi: #1}\else
  \providecommand{\doi}{doi: \begingroup \urlstyle{rm}\Url}\fi

\bibitem[Abadi et~al.(2015)Abadi, Agarwal, Barham, Brevdo, Chen, Citro,
  Corrado, Davis, Dean, Devin, Ghemawat, Goodfellow, Harp, Irving, Isard, Jia,
  Jozefowicz, Kaiser, Kudlur, Levenberg, Man\'{e}, Monga, Moore, Murray, Olah,
  Schuster, Shlens, Steiner, Sutskever, Talwar, Tucker, Vanhoucke, Vasudevan,
  Vi\'{e}gas, Vinyals, Warden, Wattenberg, Wicke, Yu, and
  Zheng]{tensorflow2015-whitepaper}
Mart\'{\i}n Abadi, Ashish Agarwal, Paul Barham, Eugene Brevdo, Zhifeng Chen,
  Craig Citro, Greg~S. Corrado, Andy Davis, Jeffrey Dean, Matthieu Devin,
  Sanjay Ghemawat, Ian Goodfellow, Andrew Harp, Geoffrey Irving, Michael Isard,
  Yangqing Jia, Rafal Jozefowicz, Lukasz Kaiser, Manjunath Kudlur, Josh
  Levenberg, Dandelion Man\'{e}, Rajat Monga, Sherry Moore, Derek Murray, Chris
  Olah, Mike Schuster, Jonathon Shlens, Benoit Steiner, Ilya Sutskever, Kunal
  Talwar, Paul Tucker, Vincent Vanhoucke, Vijay Vasudevan, Fernanda Vi\'{e}gas,
  Oriol Vinyals, Pete Warden, Martin Wattenberg, Martin Wicke, Yuan Yu, and
  Xiaoqiang Zheng.
\newblock {TensorFlow}: Large-scale machine learning on heterogeneous systems,
  2015.
\newblock URL \url{https://www.tensorflow.org/}.
\newblock Software available from tensorflow.org.

\bibitem[Arora et~al.(2014)Arora, Bhaskara, Ge, and Ma]{arora2014provable}
Sanjeev Arora, Aditya Bhaskara, Rong Ge, and Tengyu Ma.
\newblock Provable bounds for learning some deep representations.
\newblock In \emph{International conference on machine learning}, pages
  584--592. PMLR, 2014.

\bibitem[Atashgahi et~al.(2022)Atashgahi, Sokar, van~der Lee, Mocanu, Mocanu,
  Veldhuis, and Pechenizkiy]{atashgahi2020quick}
Zahra Atashgahi, Ghada Sokar, Tim van~der Lee, Elena Mocanu, Decebal~Constantin
  Mocanu, Raymond Veldhuis, and Mykola Pechenizkiy.
\newblock Quick and robust feature selection: the strength of energy-efficient
  sparse training for autoencoders.
\newblock \emph{Machine Learning (ECML-PKDD 2022 journal track)}, pages 1--38,
  2022.

\bibitem[Bartunov et~al.(2018)Bartunov, Santoro, Richards, Marris, Hinton, and
  Lillicrap]{bartunov2018assessing}
Sergey Bartunov, Adam Santoro, Blake~A Richards, Luke Marris, Geoffrey~E
  Hinton, and Timothy~P Lillicrap.
\newblock Assessing the scalability of biologically-motivated deep learning
  algorithms and architectures.
\newblock In \emph{Proceedings of the 32nd International Conference on Neural
  Information Processing Systems}, pages 9390--9400, 2018.

\bibitem[Bellec et~al.(2018)Bellec, Kappel, Maass, and
  Legenstein]{bellec2018deep}
Guillaume Bellec, David Kappel, Wolfgang Maass, and Robert Legenstein.
\newblock Deep rewiring: Training very sparse deep networks.
\newblock In \emph{International Conference on Learning Representations}, 2018.
\newblock URL \url{https://openreview.net/forum?id=BJ_wN01C-}.

\bibitem[Brown et~al.(2020)Brown, Mann, Ryder, Subbiah, Kaplan, Dhariwal,
  Neelakantan, Shyam, Sastry, Askell, Agarwal, Herbert-Voss, Krueger, Henighan,
  Child, Ramesh, Ziegler, Wu, Winter, Hesse, Chen, Sigler, Litwin, Gray, Chess,
  Clark, Berner, McCandlish, Radford, Sutskever, and
  Amodei]{NEURIPS2020_1457c0d6}
Tom Brown, Benjamin Mann, Nick Ryder, Melanie Subbiah, Jared~D Kaplan, Prafulla
  Dhariwal, Arvind Neelakantan, Pranav Shyam, Girish Sastry, Amanda Askell,
  Sandhini Agarwal, Ariel Herbert-Voss, Gretchen Krueger, Tom Henighan, Rewon
  Child, Aditya Ramesh, Daniel Ziegler, Jeffrey Wu, Clemens Winter, Chris
  Hesse, Mark Chen, Eric Sigler, Mateusz Litwin, Scott Gray, Benjamin Chess,
  Jack Clark, Christopher Berner, Sam McCandlish, Alec Radford, Ilya Sutskever,
  and Dario Amodei.
\newblock Language models are few-shot learners.
\newblock In H.~Larochelle, M.~Ranzato, R.~Hadsell, M.~F. Balcan, and H.~Lin,
  editors, \emph{Advances in Neural Information Processing Systems}, volume~33,
  pages 1877--1901. Curran Associates, Inc., 2020.
\newblock URL
  \url{https://proceedings.neurips.cc/paper/2020/file/1457c0d6bfcb4967418bfb8ac142f64a-Paper.pdf}.

\bibitem[Dai et~al.(2019)Dai, Yin, and Jha]{dai2019nest}
Xiaoliang Dai, Hongxu Yin, and Niraj~K Jha.
\newblock Nest: A neural network synthesis tool based on a grow-and-prune
  paradigm.
\newblock \emph{IEEE Transactions on Computers}, 68\penalty0 (10):\penalty0
  1487--1497, 2019.

\bibitem[de~Jorge et~al.(2020)de~Jorge, Sanyal, Behl, Torr, Rogez, and
  Dokania]{de2020progressive}
Pau de~Jorge, Amartya Sanyal, Harkirat~S Behl, Philip~HS Torr, Gregory Rogez,
  and Puneet~K Dokania.
\newblock Progressive skeletonization: Trimming more fat from a network at
  initialization.
\newblock \emph{arXiv preprint arXiv:2006.09081}, 2020.

\bibitem[Dettmers and Zettlemoyer(2019)]{dettmers2019sparse}
Tim Dettmers and Luke Zettlemoyer.
\newblock Sparse networks from scratch: Faster training without losing
  performance.
\newblock \emph{arXiv preprint arXiv:1907.04840}, 2019.

\bibitem[Evci et~al.(2020)Evci, Gale, Menick, Castro, and
  Elsen]{evci2020rigging}
Utku Evci, Trevor Gale, Jacob Menick, Pablo~Samuel Castro, and Erich Elsen.
\newblock Rigging the lottery: Making all tickets winners.
\newblock In \emph{International Conference on Machine Learning}, pages
  2943--2952. PMLR, 2020.

\bibitem[Fanty and Cole(1991)]{fanty1991spoken}
Mark Fanty and Ronald Cole.
\newblock Spoken letter recognition.
\newblock In \emph{Advances in Neural Information Processing Systems}, pages
  220--226, 1991.

\bibitem[Frankle and Carbin(2018)]{frankle2018lottery}
Jonathan Frankle and Michael Carbin.
\newblock The lottery ticket hypothesis: Finding sparse, trainable neural
  networks.
\newblock \emph{arXiv preprint arXiv:1803.03635}, 2018.

\bibitem[Friston(2008)]{friston2008hierarchical}
Karl Friston.
\newblock Hierarchical models in the brain.
\newblock \emph{PLoS computational biology}, 4\penalty0 (11):\penalty0
  e1000211, 2008.

\bibitem[Gale et~al.(2019)Gale, Elsen, and Hooker]{gale2019state}
Trevor Gale, Erich Elsen, and Sara Hooker.
\newblock The state of sparsity in deep neural networks.
\newblock \emph{arXiv preprint arXiv:1902.09574}, 2019.

\bibitem[Galke and Scherp(2021)]{galke2021forget}
Lukas Galke and Ansgar Scherp.
\newblock Forget me not: A gentle reminder to mind the simple multi-layer
  perceptron baseline for text classification.
\newblock \emph{arXiv preprint arXiv:2109.03777}, 2021.

\bibitem[Gordon et~al.(2018)Gordon, Eban, Nachum, Chen, Wu, Yang, and
  Choi]{gordon2018morphnet}
Ariel Gordon, Elad Eban, Ofir Nachum, Bo~Chen, Hao Wu, Tien-Ju Yang, and Edward
  Choi.
\newblock Morphnet: Fast \& simple resource-constrained structure learning of
  deep networks.
\newblock In \emph{Proceedings of the IEEE conference on computer vision and
  pattern recognition}, pages 1586--1595, 2018.

\bibitem[Gorishniy et~al.(2021)Gorishniy, Rubachev, Khrulkov, and
  Babenko]{gorishniy2021revisiting}
Yury Gorishniy, Ivan Rubachev, Valentin Khrulkov, and Artem Babenko.
\newblock Revisiting deep learning models for tabular data.
\newblock \emph{arXiv preprint arXiv:2106.11959}, 2021.

\bibitem[Graves et~al.(2013)Graves, Mohamed, and Hinton]{graves2013speech}
Alex Graves, Abdel-rahman Mohamed, and Geoffrey Hinton.
\newblock Speech recognition with deep recurrent neural networks.
\newblock In \emph{2013 IEEE international conference on acoustics, speech and
  signal processing}, pages 6645--6649. Ieee, 2013.

\bibitem[Group(2020)]{AIethic2019}
AI~High-Level~Expert Group.
\newblock Assessment list for trustworthy artificial intelligence ({ALTAI}) for
  self-assessment, 2020.

\bibitem[Guo et~al.(2016)Guo, Yao, and Chen]{guo2016dynamic}
Yiwen Guo, Anbang Yao, and Yurong Chen.
\newblock Dynamic network surgery for efficient dnns.
\newblock In \emph{Proceedings of the 30th International Conference on Neural
  Information Processing Systems}, NIPS'16, page 1387–1395, Red Hook, NY,
  USA, 2016. Curran Associates Inc.
\newblock ISBN 9781510838819.

\bibitem[Guyon et~al.(2008)Guyon, Gunn, Nikravesh, and Zadeh]{guyon2008feature}
Isabelle Guyon, Steve Gunn, Masoud Nikravesh, and Lofti~A Zadeh.
\newblock \emph{Feature extraction: foundations and applications}, volume 207.
\newblock Springer, 2008.

\bibitem[Han et~al.(2012)Han, Kamber, Pei, et~al.]{han2012getting}
Jiawei Han, Micheline Kamber, Jian Pei, et~al.
\newblock Getting to know your data.
\newblock In \emph{Data mining}, pages 39--82. Elsevier Amsterdam, Netherlands,
  2012.

\bibitem[Han et~al.(2015)Han, Pool, Tran, and Dally]{han2015learning}
Song Han, Jeff Pool, John Tran, and William~J Dally.
\newblock Learning both weights and connections for efficient neural networks.
\newblock In \emph{Proceedings of the 28th International Conference on Neural
  Information Processing Systems-Volume 1}, pages 1135--1143, 2015.

\bibitem[Hassibi and Stork(1993)]{hassibi1993second}
Babak Hassibi and David~G Stork.
\newblock Second order derivatives for network pruning: Optimal brain surgeon.
\newblock In \emph{Advances in neural information processing systems}, pages
  164--171, 1993.

\bibitem[Hebb(2005)]{hebb2005organization}
Donald~Olding Hebb.
\newblock \emph{The organization of behavior: A neuropsychological theory}.
\newblock Psychology Press, 2005.

\bibitem[Hestness et~al.(2017)Hestness, Narang, Ardalani, Diamos, Jun,
  Kianinejad, Patwary, Ali, Yang, and Zhou]{hestness2017deep}
Joel Hestness, Sharan Narang, Newsha Ardalani, Gregory Diamos, Heewoo Jun,
  Hassan Kianinejad, Md~Patwary, Mostofa Ali, Yang Yang, and Yanqi Zhou.
\newblock Deep learning scaling is predictable, empirically.
\newblock \emph{arXiv preprint arXiv:1712.00409}, 2017.

\bibitem[Hoefler et~al.(2021)Hoefler, Alistarh, Ben-Nun, Dryden, and
  Peste]{hoefler2021sparsity}
Torsten Hoefler, Dan Alistarh, Tal Ben-Nun, Nikoli Dryden, and Alexandra Peste.
\newblock Sparsity in deep learning: Pruning and growth for efficient inference
  and training in neural networks.
\newblock \emph{arXiv preprint arXiv:2102.00554}, 2021.

\bibitem[Jayakumar et~al.(2020)Jayakumar, Pascanu, Rae, Osindero, and
  Elsen]{jayakumar2020top}
Siddhant Jayakumar, Razvan Pascanu, Jack Rae, Simon Osindero, and Erich Elsen.
\newblock Top-kast: Top-k always sparse training.
\newblock \emph{Advances in Neural Information Processing Systems},
  33:\penalty0 20744--20754, 2020.

\bibitem[Jouppi et~al.(2017)Jouppi, Young, Patil, Patterson, Agrawal, Bajwa,
  Bates, Bhatia, Boden, Borchers, et~al.]{jouppi2017datacenter}
Norman~P Jouppi, Cliff Young, Nishant Patil, David Patterson, Gaurav Agrawal,
  Raminder Bajwa, Sarah Bates, Suresh Bhatia, Nan Boden, Al~Borchers, et~al.
\newblock In-datacenter performance analysis of a tensor processing unit.
\newblock In \emph{Proceedings of the 44th annual international symposium on
  computer architecture}, pages 1--12, 2017.

\bibitem[Junjie et~al.(2019)Junjie, Zhe, Runbin, Cheung, and
  So]{junjie2019dynamic}
LIU Junjie, XU~Zhe, SHI Runbin, Ray~CC Cheung, and Hayden~KH So.
\newblock Dynamic sparse training: Find efficient sparse network from scratch
  with trainable masked layers.
\newblock In \emph{International Conference on Learning Representations}, 2019.

\bibitem[Kepner and Robinett(2019)]{kepner2019radix}
Jeremy Kepner and Ryan Robinett.
\newblock Radix-net: Structured sparse matrices for deep neural networks.
\newblock In \emph{2019 IEEE International Parallel and Distributed Processing
  Symposium Workshops (IPDPSW)}, pages 268--274. IEEE, 2019.

\bibitem[Krizhevsky et~al.(2009)Krizhevsky, Hinton,
  et~al.]{krizhevsky2009learning}
Alex Krizhevsky, Geoffrey Hinton, et~al.
\newblock Learning multiple layers of features from tiny images.
\newblock 2009.

\bibitem[Kuriscak et~al.(2015)Kuriscak, Marsalek, Stroffek, and
  Toth]{kuriscak2015biological}
Eduard Kuriscak, Petr Marsalek, Julius Stroffek, and Peter~G Toth.
\newblock Biological context of hebb learning in artificial neural networks, a
  review.
\newblock \emph{Neurocomputing}, 152:\penalty0 27--35, 2015.

\bibitem[Kusupati et~al.(2020)Kusupati, Ramanujan, Somani, Wortsman, Jain,
  Kakade, and Farhadi]{pmlr-v119-kusupati20a}
Aditya Kusupati, Vivek Ramanujan, Raghav Somani, Mitchell Wortsman, Prateek
  Jain, Sham Kakade, and Ali Farhadi.
\newblock Soft threshold weight reparameterization for learnable sparsity.
\newblock In Hal~Daumé III and Aarti Singh, editors, \emph{Proceedings of the
  37th International Conference on Machine Learning}, volume 119 of
  \emph{Proceedings of Machine Learning Research}, pages 5544--5555. PMLR,
  13--18 Jul 2020.
\newblock URL \url{http://proceedings.mlr.press/v119/kusupati20a.html}.

\bibitem[Lang(1995)]{lang1995newsweeder}
Ken Lang.
\newblock Newsweeder: Learning to filter netnews.
\newblock In \emph{Machine Learning Proceedings 1995}, pages 331--339.
  Elsevier, 1995.

\bibitem[LeCun(1998)]{lecun1998mnist}
Yann LeCun.
\newblock The mnist database of handwritten digits.
\newblock \emph{http://yann. lecun. com/exdb/mnist/}, 1998.

\bibitem[LeCun et~al.(1990)LeCun, Denker, and Solla]{lecun1990optimal}
Yann LeCun, John~S Denker, and Sara~A Solla.
\newblock Optimal brain damage.
\newblock In \emph{Advances in neural information processing systems}, pages
  598--605, 1990.

\bibitem[Lee et~al.(2019)Lee, Ajanthan, and Torr]{lee2018snip}
Namhoon Lee, Thalaiyasingam Ajanthan, and Philip Torr.
\newblock {SNIP}: {SINGLE}-{SHOT} {NETWORK} {PRUNING} {BASED} {ON} {CONNECTION}
  {SENSITIVITY}.
\newblock In \emph{International Conference on Learning Representations}, 2019.
\newblock URL \url{https://openreview.net/forum?id=B1VZqjAcYX}.

\bibitem[Li and Han(2013)]{li2013distance}
Baoli Li and Liping Han.
\newblock Distance weighted cosine similarity measure for text classification.
\newblock In \emph{International conference on intelligent data engineering and
  automated learning}, pages 611--618. Springer, 2013.

\bibitem[Li et~al.(2020)Li, Gu, Mayer, Gool, and Timofte]{li2020group}
Yawei Li, Shuhang Gu, Christoph Mayer, Luc~Van Gool, and Radu Timofte.
\newblock Group sparsity: The hinge between filter pruning and decomposition
  for network compression.
\newblock In \emph{Proceedings of the IEEE/CVF Conference on Computer Vision
  and Pattern Recognition}, pages 8018--8027, 2020.

\bibitem[Liang and Hu(2015)]{liang2015recurrent}
Ming Liang and Xiaolin Hu.
\newblock Recurrent convolutional neural network for object recognition.
\newblock In \emph{Proceedings of the IEEE conference on computer vision and
  pattern recognition}, pages 3367--3375, 2015.

\bibitem[Liu and Wu(2019)]{LIU201984}
Congcong Liu and Huaming Wu.
\newblock Channel pruning based on mean gradient for accelerating convolutional
  neural networks.
\newblock \emph{Signal Processing}, 156:\penalty0 84--91, 2019.
\newblock ISSN 0165-1684.
\newblock \doi{https://doi.org/10.1016/j.sigpro.2018.10.019}.
\newblock URL
  \url{https://www.sciencedirect.com/science/article/pii/S0165168418303517}.

\bibitem[Liu et~al.(2017)Liu, Gong, and Miao]{liu2017modeling}
Jia Liu, Maoguo Gong, and Qiguang Miao.
\newblock Modeling hebb learning rule for unsupervised learning.
\newblock In \emph{IJCAI}, pages 2315--2321, 2017.

\bibitem[Liu et~al.(2020)Liu, van~der Lee, Yaman, Atashgahi, Ferraro, Sokar,
  Pechenizkiy, and Mocanu]{liu2020topological}
Shiwei Liu, Tim van~der Lee, Anil Yaman, Zahra Atashgahi, Davide Ferraro, Ghada
  Sokar, Mykola Pechenizkiy, and Decebal~Constantin Mocanu.
\newblock Topological insights into sparse neural networks.
\newblock In \emph{Proceedings of the European Conference on Machine Learning
  and Principles and Practice of Knowledge Discovery in Databases (ECML PKDD)
  2020.}, pages 2006--14085, 2020.

\bibitem[Liu et~al.(2021{\natexlab{a}})Liu, Mocanu, Matavalam, Pei, and
  Pechenizkiy]{liu2021sparse}
Shiwei Liu, Decebal~Constantin Mocanu, Amarsagar Reddy~Ramapuram Matavalam,
  Yulong Pei, and Mykola Pechenizkiy.
\newblock Sparse evolutionary deep learning with over one million artificial
  neurons on commodity hardware.
\newblock \emph{Neural Computing and Applications}, 33\penalty0 (7):\penalty0
  2589--2604, 2021{\natexlab{a}}.

\bibitem[Liu et~al.(2021{\natexlab{b}})Liu, Mocanu, Pei, and
  Pechenizkiy]{liu2021selfish}
Shiwei Liu, Decebal~Constantin Mocanu, Yulong Pei, and Mykola Pechenizkiy.
\newblock Selfish sparse rnn training.
\newblock In Marina Meila and Tong Zhang, editors, \emph{Proceedings of the
  38th International Conference on Machine Learning}, volume 139 of
  \emph{Proceedings of Machine Learning Research}, pages 6893--6904. PMLR,
  18--24 Jul 2021{\natexlab{b}}.
\newblock URL \url{https://proceedings.mlr.press/v139/liu21p.html}.

\bibitem[Liu et~al.(2021{\natexlab{c}})Liu, Yin, Mocanu, and
  Pechenizkiy]{liu2021we}
Shiwei Liu, Lu~Yin, Decebal~Constantin Mocanu, and Mykola Pechenizkiy.
\newblock Do we actually need dense over-parameterization? in-time
  over-parameterization in sparse training.
\newblock In Marina Meila and Tong Zhang, editors, \emph{Proceedings of the
  38th International Conference on Machine Learning}, volume 139 of
  \emph{Proceedings of Machine Learning Research}, pages 6989--7000. PMLR,
  18--24 Jul 2021{\natexlab{c}}.
\newblock URL \url{https://proceedings.mlr.press/v139/liu21y.html}.

\bibitem[Louizos et~al.(2018)Louizos, Welling, and Kingma]{louizos2018learning}
Christos Louizos, Max Welling, and Diederik~P. Kingma.
\newblock Learning sparse neural networks through l0 regularization.
\newblock In \emph{International Conference on Learning Representations}, 2018.
\newblock URL \url{https://openreview.net/forum?id=H1Y8hhg0b}.

\bibitem[Luo et~al.(2018)Luo, Zhan, Xue, Wang, Ren, and Yang]{luo2018cosine}
Chunjie Luo, Jianfeng Zhan, Xiaohe Xue, Lei Wang, Rui Ren, and Qiang Yang.
\newblock Cosine normalization: Using cosine similarity instead of dot product
  in neural networks.
\newblock In \emph{International Conference on Artificial Neural Networks},
  pages 382--391. Springer, 2018.

\bibitem[Masi et~al.(2018)Masi, Wu, Hassner, and Natarajan]{masi2018deep}
Iacopo Masi, Yue Wu, Tal Hassner, and Prem Natarajan.
\newblock Deep face recognition: A survey.
\newblock In \emph{2018 31st SIBGRAPI conference on graphics, patterns and
  images (SIBGRAPI)}, pages 471--478. IEEE, 2018.

\bibitem[Mocanu et~al.(2016)Mocanu, Mocanu, Nguyen, Gibescu, and
  Liotta]{mocanu2016topological}
Decebal~Constantin Mocanu, Elena Mocanu, Phuong~H Nguyen, Madeleine Gibescu,
  and Antonio Liotta.
\newblock A topological insight into restricted boltzmann machines.
\newblock \emph{Machine Learning}, 104\penalty0 (2-3):\penalty0 243--270, 2016.

\bibitem[Mocanu et~al.(2018)Mocanu, Mocanu, Stone, Nguyen, Gibescu, and
  Liotta]{mocanu2018scalable}
Decebal~Constantin Mocanu, Elena Mocanu, Peter Stone, Phuong~H Nguyen,
  Madeleine Gibescu, and Antonio Liotta.
\newblock Scalable training of artificial neural networks with adaptive sparse
  connectivity inspired by network science.
\newblock \emph{Nature communications}, 9\penalty0 (1):\penalty0 2383, 2018.

\bibitem[Mocanu et~al.(2021)Mocanu, Mocanu, Pinto, Curci, Nguyen, Gibescu,
  Ernst, and Vale]{mocanu2021sparse}
Decebal~Constantin Mocanu, Elena Mocanu, Tiago Pinto, Selima Curci, Phuong~H
  Nguyen, Madeleine Gibescu, Damien Ernst, and Zita~A Vale.
\newblock Sparse training theory for scalable and efficient agents.
\newblock In \emph{Proceedings of the 20th International Conference on
  Autonomous Agents and MultiAgent Systems}, pages 34--38, 2021.

\bibitem[Molchanov et~al.(2017)Molchanov, Ashukha, and
  Vetrov]{molchanov2017variational}
Dmitry Molchanov, Arsenii Ashukha, and Dmitry Vetrov.
\newblock Variational dropout sparsifies deep neural networks.
\newblock In \emph{International Conference on Machine Learning}, pages
  2498--2507. PMLR, 2017.

\bibitem[Molchanov et~al.(2016)Molchanov, Tyree, Karras, Aila, and
  Kautz]{molchanov2016pruning}
Pavlo Molchanov, Stephen Tyree, Tero Karras, Timo Aila, and Jan Kautz.
\newblock Pruning convolutional neural networks for resource efficient
  inference.
\newblock 2016.

\bibitem[Molchanov et~al.(2019)Molchanov, Mallya, Tyree, Frosio, and
  Kautz]{Molchanov_2019_CVPR}
Pavlo Molchanov, Arun Mallya, Stephen Tyree, Iuri Frosio, and Jan Kautz.
\newblock Importance estimation for neural network pruning.
\newblock In \emph{Proceedings of the IEEE/CVF Conference on Computer Vision
  and Pattern Recognition (CVPR)}, June 2019.

\bibitem[Mostafa and Wang(2019)]{pmlr-v97-mostafa19a}
Hesham Mostafa and Xin Wang.
\newblock Parameter efficient training of deep convolutional neural networks by
  dynamic sparse reparameterization.
\newblock In Kamalika Chaudhuri and Ruslan Salakhutdinov, editors,
  \emph{Proceedings of the 36th International Conference on Machine Learning},
  volume~97 of \emph{Proceedings of Machine Learning Research}, pages
  4646--4655. PMLR, 09--15 Jun 2019.
\newblock URL \url{http://proceedings.mlr.press/v97/mostafa19a.html}.

\bibitem[Neyshabur et~al.(2019)Neyshabur, Li, Bhojanapalli, LeCun, and
  Srebro]{neyshabur2018the}
Behnam Neyshabur, Zhiyuan Li, Srinadh Bhojanapalli, Yann LeCun, and Nathan
  Srebro.
\newblock The role of over-parametrization in generalization of neural
  networks.
\newblock In \emph{International Conference on Learning Representations}, 2019.
\newblock URL \url{https://openreview.net/forum?id=BygfghAcYX}.

\bibitem[Nguyen and Bai(2010)]{nguyen2010cosine}
Hieu~V Nguyen and Li~Bai.
\newblock Cosine similarity metric learning for face verification.
\newblock In \emph{Asian conference on computer vision}, pages 709--720.
  Springer, 2010.

\bibitem[Pogodin et~al.(2021)Pogodin, Mehta, Lillicrap, and
  Latham]{pogodin2021towards}
Roman Pogodin, Yash Mehta, Timothy~P Lillicrap, and Peter~E Latham.
\newblock Towards biologically plausible convolutional networks.
\newblock \emph{arXiv preprint arXiv:2106.13031}, 2021.

\bibitem[Popov et~al.(2019)Popov, Morozov, and Babenko]{popov2019neural}
Sergei Popov, Stanislav Morozov, and Artem Babenko.
\newblock Neural oblivious decision ensembles for deep learning on tabular
  data.
\newblock \emph{arXiv preprint arXiv:1909.06312}, 2019.

\bibitem[Raihan and Aamodt(2020)]{raihan2020sparse}
Md~Aamir Raihan and Tor~M Aamodt.
\newblock Sparse weight activation training.
\newblock \emph{arXiv preprint arXiv:2001.01969}, 2020.

\bibitem[Savarese et~al.(2020)Savarese, Silva, and Maire]{NEURIPS2020_83004190}
Pedro Savarese, Hugo Silva, and Michael Maire.
\newblock Winning the lottery with continuous sparsification.
\newblock In H.~Larochelle, M.~Ranzato, R.~Hadsell, M.~F. Balcan, and H.~Lin,
  editors, \emph{Advances in Neural Information Processing Systems}, volume~33,
  pages 11380--11390. Curran Associates, Inc., 2020.
\newblock URL
  \url{https://proceedings.neurips.cc/paper/2020/file/83004190b1793d7aa15f8d0d49a13eba-Paper.pdf}.

\bibitem[Scellier and Bengio(2016)]{scellier2016towards}
Benjamin Scellier and Yoshua Bengio.
\newblock Towards a biologically plausible backprop.
\newblock \emph{arXiv preprint arXiv:1602.05179}, 914, 2016.

\bibitem[Schumacher(2021)]{schumacher2021livewired}
Thomas Schumacher.
\newblock Livewired neural networks: Making neurons that fire together wire
  together.
\newblock \emph{arXiv preprint arXiv:2105.08111}, 2021.

\bibitem[Sidorov et~al.(2014)Sidorov, Gelbukh, G{\'o}mez-Adorno, and
  Pinto]{sidorov2014soft}
Grigori Sidorov, Alexander Gelbukh, Helena G{\'o}mez-Adorno, and David Pinto.
\newblock Soft similarity and soft cosine measure: Similarity of features in
  vector space model.
\newblock \emph{Computaci{\'o}n y Sistemas}, 18\penalty0 (3):\penalty0
  491--504, 2014.

\bibitem[Sun et~al.(2016)Sun, Wang, and Tang]{sun2016sparsifying}
Yi~Sun, Xiaogang Wang, and Xiaoou Tang.
\newblock Sparsifying neural network connections for face recognition.
\newblock In \emph{Proceedings of the IEEE Conference on Computer Vision and
  Pattern Recognition}, pages 4856--4864, 2016.

\bibitem[Tanaka et~al.(2020)Tanaka, Kunin, Yamins, and
  Ganguli]{tanaka2020pruning}
Hidenori Tanaka, Daniel Kunin, Daniel~L Yamins, and Surya Ganguli.
\newblock Pruning neural networks without any data by iteratively conserving
  synaptic flow.
\newblock \emph{Advances in Neural Information Processing Systems}, 33, 2020.

\bibitem[Tolstikhin et~al.(2021)Tolstikhin, Houlsby, Kolesnikov, Beyer, Zhai,
  Unterthiner, Yung, Keysers, Uszkoreit, Lucic, et~al.]{tolstikhin2021mlp}
Ilya Tolstikhin, Neil Houlsby, Alexander Kolesnikov, Lucas Beyer, Xiaohua Zhai,
  Thomas Unterthiner, Jessica Yung, Daniel Keysers, Jakob Uszkoreit, Mario
  Lucic, et~al.
\newblock Mlp-mixer: An all-mlp architecture for vision.
\newblock \emph{arXiv preprint arXiv:2105.01601}, 2021.

\bibitem[Wang et~al.(2019{\natexlab{a}})Wang, Grosse, Fidler, and
  Zhang]{wang2019eigendamage}
Chaoqi Wang, Roger Grosse, Sanja Fidler, and Guodong Zhang.
\newblock Eigendamage: Structured pruning in the kronecker-factored eigenbasis.
\newblock In \emph{International Conference on Machine Learning}, pages
  6566--6575. PMLR, 2019{\natexlab{a}}.

\bibitem[Wang et~al.(2019{\natexlab{b}})Wang, Zhang, and
  Grosse]{wang2019picking}
Chaoqi Wang, Guodong Zhang, and Roger Grosse.
\newblock Picking winning tickets before training by preserving gradient flow.
\newblock In \emph{International Conference on Learning Representations},
  2019{\natexlab{b}}.

\bibitem[Wen et~al.(2016)Wen, Wu, Wang, Chen, and Li]{10.5555/3157096.3157329}
Wei Wen, Chunpeng Wu, Yandan Wang, Yiran Chen, and Hai Li.
\newblock Learning structured sparsity in deep neural networks.
\newblock In \emph{Proceedings of the 30th International Conference on Neural
  Information Processing Systems}, NIPS'16, page 2082–2090, Red Hook, NY,
  USA, 2016. Curran Associates Inc.
\newblock ISBN 9781510838819.

\bibitem[Xia et~al.(2015)Xia, Zhang, and Li]{XIA201539}
Peipei Xia, Li~Zhang, and Fanzhang Li.
\newblock Learning similarity with cosine similarity ensemble.
\newblock \emph{Information Sciences}, 307:\penalty0 39--52, 2015.
\newblock ISSN 0020-0255.
\newblock \doi{https://doi.org/10.1016/j.ins.2015.02.024}.
\newblock URL
  \url{https://www.sciencedirect.com/science/article/pii/S0020025515001243}.

\bibitem[Xiao et~al.(2017)Xiao, Rasul, and Vollgraf]{xiao2017}
Han Xiao, Kashif Rasul, and Roland Vollgraf.
\newblock Fashion-mnist: a novel image dataset for benchmarking machine
  learning algorithms, 2017.

\bibitem[Yang et~al.(2018)Yang, Xiao, Jiang, Hossain, Muhammad, and
  Amin]{yang2018ai}
Jun Yang, Wenjing Xiao, Chun Jiang, M~Shamim Hossain, Ghulam Muhammad, and
  Syed~Umar Amin.
\newblock Ai-powered green cloud and data center.
\newblock \emph{IEEE Access}, 7:\penalty0 4195--4203, 2018.

\bibitem[Zhang et~al.(2020)Zhang, Zhang, Lane, Shu, Zeng, Fang, Yan, and
  Xu]{zhang2020deep}
Mi~Zhang, Faen Zhang, Nicholas~D Lane, Yuanchao Shu, Xiao Zeng, Biyi Fang, Shen
  Yan, and Hui Xu.
\newblock Deep learning in the era of edge computing: Challenges and
  opportunities.
\newblock \emph{Fog Computing: Theory and Practice}, 2020.

\bibitem[Zhu and Gupta(2017)]{zhu2017prune}
Michael Zhu and Suyog Gupta.
\newblock To prune, or not to prune: exploring the efficacy of pruning for
  model compression.
\newblock \emph{arXiv preprint arXiv:1710.01878}, 2017.

\end{thebibliography}
